\documentclass[twoside]{article}

\usepackage[margin=2cm]{geometry}
\usepackage{amsmath}
\usepackage{amssymb}
\usepackage[ansinew]{inputenc}
\usepackage{bm}
\usepackage{float}
\usepackage{url}
\usepackage{algorithm}
\usepackage{algorithmic}
\usepackage{graphicx}
\usepackage{subfigure}
\usepackage[small]{caption}
\usepackage{url}
\usepackage{verbatim}
\usepackage{natbib}

\global\long\def\lengthScale{\ell}

\global\long\def\weightScalar{w}
\global\long\def\weightVector{{\bf \weightScalar}}
\global\long\def\meanVector{\boldsymbol{\mu}}
\global\long\def\zerosVector{{\bf 0}}

\global\long\def\numData{n}

\global\long\def\dataDim{p}
\global\long\def\dataScalar{y}
\global\long\def\dataMatrix{{\bf \uppercase{\dataScalar}}}
\global\long\def\dataVector{{\bf \dataScalar}}

\global\long\def\twoPi{\tau}

\global\long\def\parameterScalar{\theta}
\global\long\def\parameterVector{\boldsymbol{\parameterScalar}}
\global\long\def\kernelMatrix{\mathbf{K}}

\global\long\def\covarianceMatrix{\mathbf{C}}
\global\long\def\det#1{\left|#1\right|}

\global\long\def\refeq#1{(\ref{#1})}
\global\long\def\refsec#1{Section \ref{#1}}

\global\long\def\reffig#1{Figure \ref{#1}}

\global\long\def\eye{\mathbf{I}}

\global\long\def\trace#1{\text{tr}\left(#1\right)}
\global\long\def\diff#1#2{\frac{\text{d}#1}{\text{d}#2}}

\global\long\def\cut#1{}

\global\long\def\reffig#1{figure~\ref{#1}}

\global\long\def\reffigrange#1#2{figure~\ref{#1}--\ref{#2}}

\global\long\def\detail#1{}

\usepackage{color}
\usepackage{verbatim}

\newenvironment{octave}{\comment}{\endcomment}

\global\long\def\dataDim{p}
\global\long\def\latentDim{q}

\global\long\def\numData{n}

\global\long\def\numNeighbors{K}

\global\long\def\lengthScale{\ell}

\global\long\def\errorFunction{E}

\global\long\def\parameterScalar{\theta}
\global\long\def\parameterVector{\boldsymbol{\parameterScalar}}

\global\long\def\dataScalar{y}
\global\long\def\dataVector{\mathbf{\dataScalar}}
\global\long\def\dataMatrix{\mathbf{\MakeUppercase{\dataScalar}}}

\global\long\def\cdataMatrix{\hat{\dataMatrix}}
\global\long\def\cdataVector{\hat{\dataVector}}

\global\long\def\latentScalar{x}
\global\long\def\latentMatrix{\mathbf{\MakeUppercase{\latentScalar}}}
\global\long\def\latentVector{\mathbf{\latentScalar}}

\global\long\def\kernelScalar{k}
\global\long\def\kernelVector{\mathbf{\kernelScalar}}
\global\long\def\kernelMatrix{\mathbf{\MakeUppercase{\kernelScalar}}}

\global\long\def\centeredKernelScalar{b}

\global\long\def\centeredKernelMatrix{\mathbf{\MakeUppercase{\centeredKernelScalar}}}

\global\long\def\covarianceScalar{c}
\global\long\def\covarianceMatrix{\mathbf{\MakeUppercase{\covarianceScalar}}}

\global\long\def\precisionScalar{j}
\global\long\def\precisionMatrix{\mathbf{\MakeUppercase{\precisionScalar}}}

\global\long\def\meanScalar{\mu}

\global\long\def\meanVector{\boldsymbol{\meanScalar}}

\global\long\def\eigenvectorScalar{u}
\global\long\def\eigenvector{\mathbf{\eigenvectorScalar}}
\global\long\def\eigenvectorMatrix{\mathbf{\MakeUppercase{\eigenvectorScalar}}}

\global\long\def\eigenvalue{\lambda}

\global\long\def\eigenvalueMatrix{\boldsymbol{\Lambda}}

\global\long\def\eigenvectwo{\mathbf{v}}

\global\long\def\laplacianScalar{\ell}

\global\long\def\laplacianMatrix{\mathbf{L}}

\global\long\def\normalizedLaplacianMatrix{\hat{\mathbf{L}}}

\global\long\def\weightedAdjacencyScalar{a}

\global\long\def\weightedAdjacencyMatrix{\mathbf{\MakeUppercase{\weightedAdjacencyScalar}}}

\global\long\def\degreeScalar{d}

\global\long\def\degreeMatrix{\mathbf{\MakeUppercase{\degreeScalar}}}

\global\long\def\lagrangeMultiplier{\lambda}
\global\long\def\lagrangeMultiplierMatrix{\boldsymbol{\Lambda}}

\global\long\def\laplacianFactor{\mathbf{\MakeUppercase{\laplacianFactorScalar}}}
\global\long\def\laplacianFactorScalar{m}
\global\long\def\laplacianFactorVector{\mathbf{\laplacianFactorScalar}}

\global\long\def\latentDistanceScalar{\delta}

\global\long\def\latentDistanceMatrix{\boldsymbol{\Delta}}

\global\long\def\springScalar{\kappa}

\global\long\def\springMatrix{\boldsymbol{\mathcal{K}}}

\global\long\def\distanceScalar{d}

\global\long\def\distanceMatrix{\mathbf{\MakeUppercase{\distanceScalar}}}

\global\long\def\weightScalar{w}
\global\long\def\weightVector{\mathbf{\weightScalar}}
\global\long\def\weightMatrix{\mathbf{\MakeUppercase{\weightScalar}}}

\global\long\def\aMatrix{\mathbf{A}}
\global\long\def\aVector{\mathbf{a}}

\global\long\def\centeringMatrix{\mathbf{H}}

\global\long\def\eye{\mathbf{I}}
\global\long\def\onesVector{\mathbf{1}}
\global\long\def\zerosVector{\mathbf{0}}

\global\long\def\sampleCovScalar{s}

\global\long\def\sampleCovMatrix{\mathbf{\MakeUppercase{\sampleCovScalar}}}

\global\long\def\rotationScalar{r}

\global\long\def\rotationMatrix{\mathbf{\MakeUppercase{\rotationScalar}}}

\global\long\def\diff#1#2{\frac{\text{d}#1}{\text{d}#2}}

\global\long\def\gaussianSamp#1#2{\mathcal{N}\left(#1,#2\right)}
\global\long\def\gaussianDist#1#2#3{\mathcal{N}\left(#1|#2,#3\right)}

\global\long\def\expDist#1#2{\left<#1\right>_{#2}}

\global\long\def\diagonalMatrix{\mathbf{D}}

\global\long\def\expectation#1{\left\langle #1 \right\rangle }
\global\long\def\expectationDist#1#2{\left\langle #1 \right\rangle _{#2}}

\global\long\def\KL#1#2{\text{KL}\left( #1\,\|\,#2 \right)}

\global\long\def\det#1{\left|#1\right|}

\global\long\def\tr#1{\text{tr}\left(#1\right)}
\global\long\def\diag#1{\text{diag}\left(#1\right)}

\global\long\def\neighborhood#1{\mathcal{N}\left( #1 \right)}
\global\long\def\ltwoNorm#1{\left\Vert #1 \right\Vert_2}
\global\long\def\loneNorm#1{\left\Vert #1 \right\Vert_1}

\title{A Unifying Probabilistic Perspective for Spectral Dimensionality Reduction: Insights and New Models}

\author{
Neil D. Lawrence\thanks{Work also carried out at the School of Computer Science, University of Manchester.} \\
Sheffield Institute for Translational Neuroscience\\
and\\ Department of Computer Science\\
University of Sheffield\\
\texttt{N.Lawrence@dcs.shef.ac.uk} 
}

\global\long\def\lengthScale{\sigma}
\global\long\def\degreeScalar{v} 
\global\long\def\precisionScalar{p} 

\begin{document}

\maketitle


\begin{abstract}
  We introduce a new perspective on spectral dimensionality reduction
  which views these methods as Gaussian Markov random fields
  (GRFs). Our unifying perspective is based on the maximum entropy
  principle which is in turn inspired by maximum variance
  unfolding. The resulting model, which we call maximum entropy
  unfolding (MEU) is a nonlinear generalization of principal component
  analysis. We relate the model to Laplacian eigenmaps and isomap. We
  show that parameter fitting in the locally linear embedding (LLE) is
  approximate maximum likelihood MEU. We introduce a variant of LLE
  that performs maximum likelihood exactly: Acyclic LLE (ALLE).  We
  show that MEU and ALLE are competitive with the leading spectral
  approaches on a robot navigation visualization and a human motion
  capture data set. Finally the maximum likelihood perspective allows
  us to introduce a new approach to dimensionality reduction based on
  L1 regularization of the Gaussian random field via the graphical
  lasso.
\end{abstract}

\section{Introduction}
\begin{octave}

  clear all
  close all

  global printDiagram
  printDiagram = 0; 

  rand('seed', 1e5);
  randn('seed', 1e5);

  importTool('meu')
  meuToolboxes;
  textWidth = 17; 

  fontName = 'times';
  fontSize = 26;

  a = ver('octave');
  if length(a) == 0
    a = ver('matlab');
  end
  fid = fopen('vers.tex', 'w');
  fprintf(fid, [a.Name ' version ' a.Version]);
  fclose(fid);

  fid = fopen('computer.tex', 'w');
  fprintf(fid, ['\\verb+' computer '+']);
  fclose(fid);


  
\end{octave}

A representation of an object for processing by computer typically
requires that object to be summarized by a series of features,
represented by numbers. As the representation becomes more complex,
the number of features required typically increases. Examples include:
the characteristics of a customer in a database; the pixel intensities
in an image; a time series of angles associated with data captured
from human motion for animation; the energy at different frequencies
(or across the cepstrum) as a time series for interpreting speech; the
frequencies of given words as they appear in a set of documents; the
level of expression of thousands of genes, across a time series, or
for different diseases.

With the increasing complexity of the representation, the number of
features that are stored also increases. Data of this type is known
as high dimensional data.

\begin{figure}[ht]
  \subfigure[]{\includegraphics[width=0.27\textwidth]{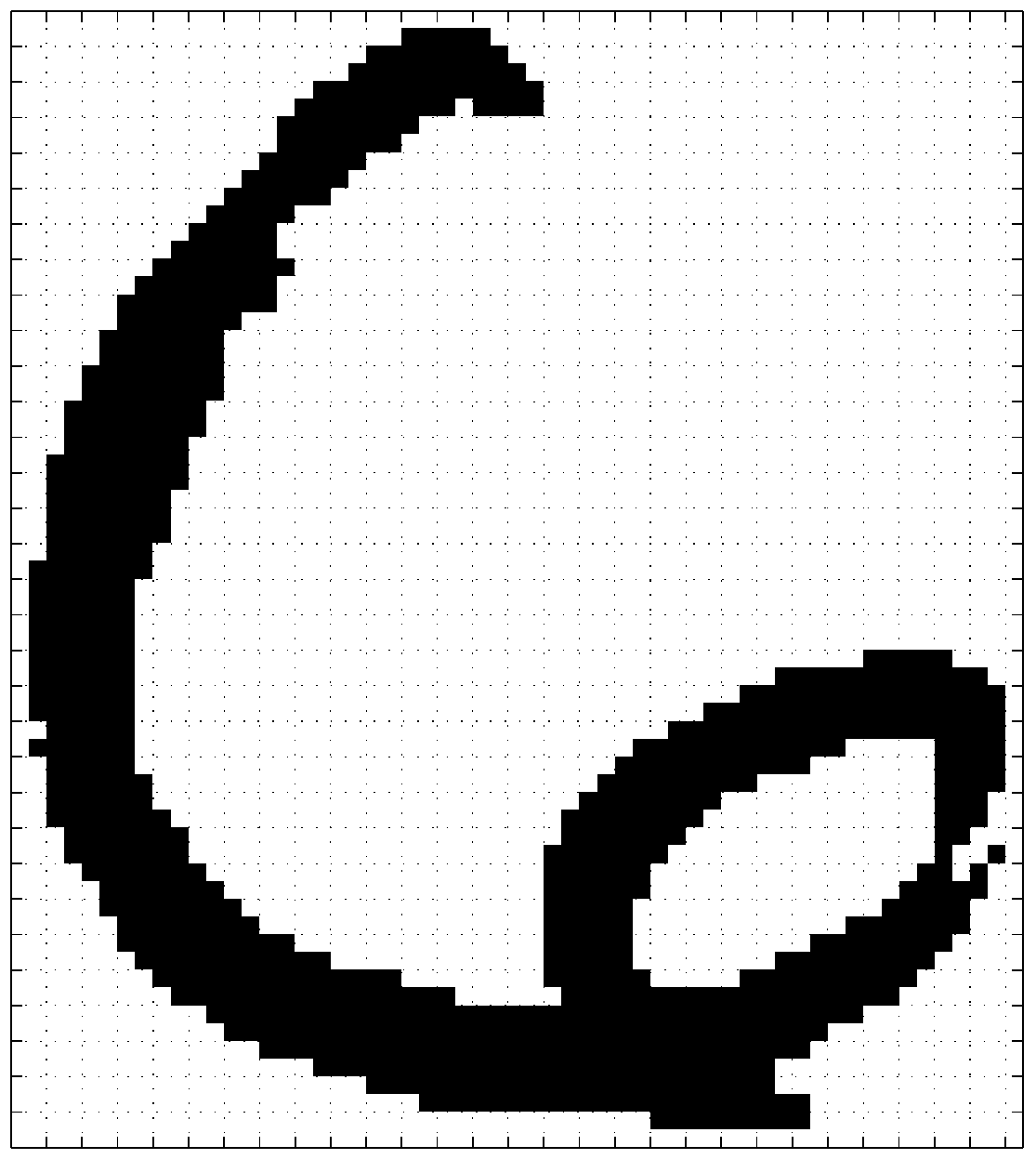}}\hfill
  \subfigure[]{\includegraphics[width=0.27\textwidth]{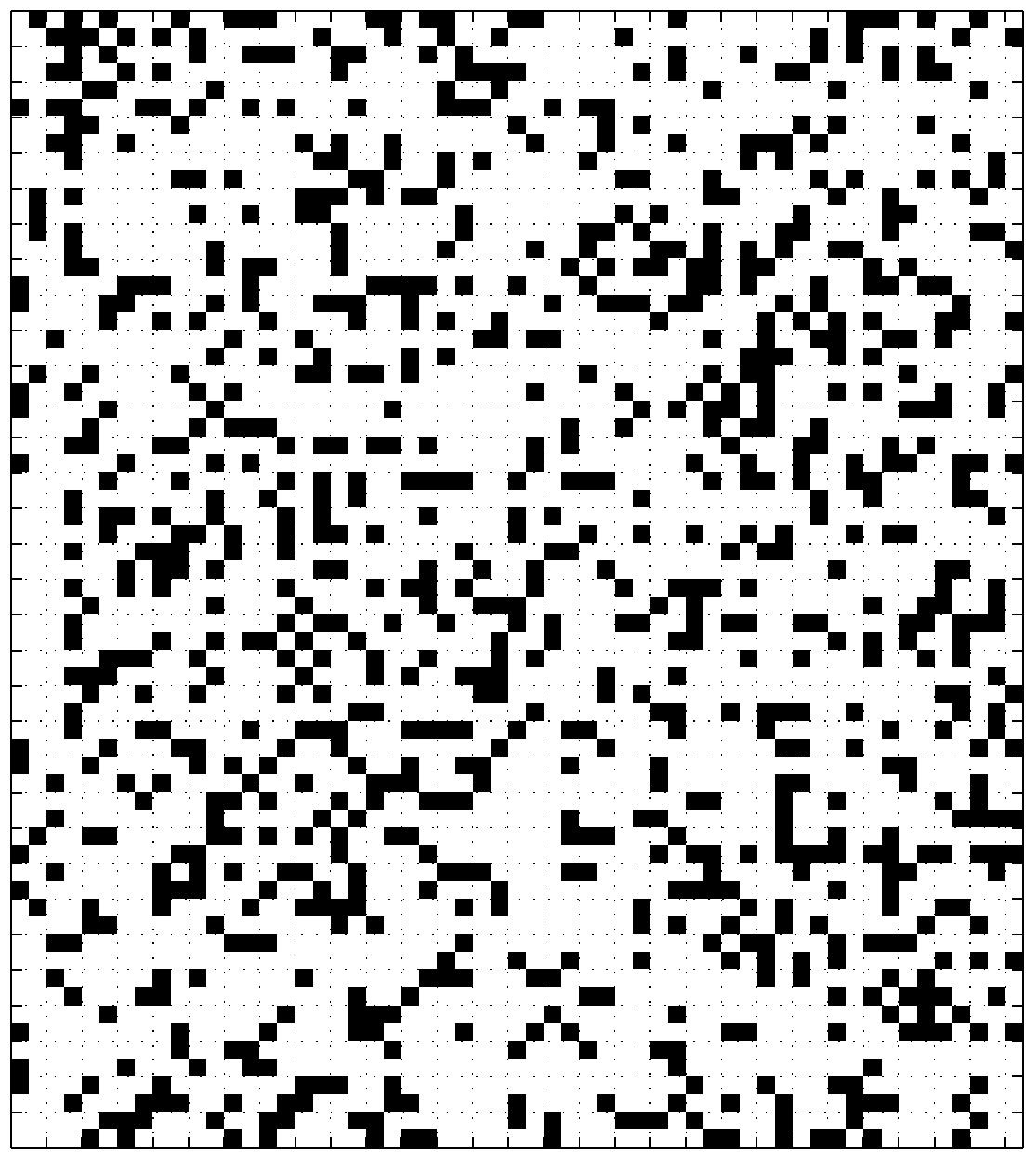}}\hfill
  \subfigure[]{\includegraphics[width=0.45\textwidth]{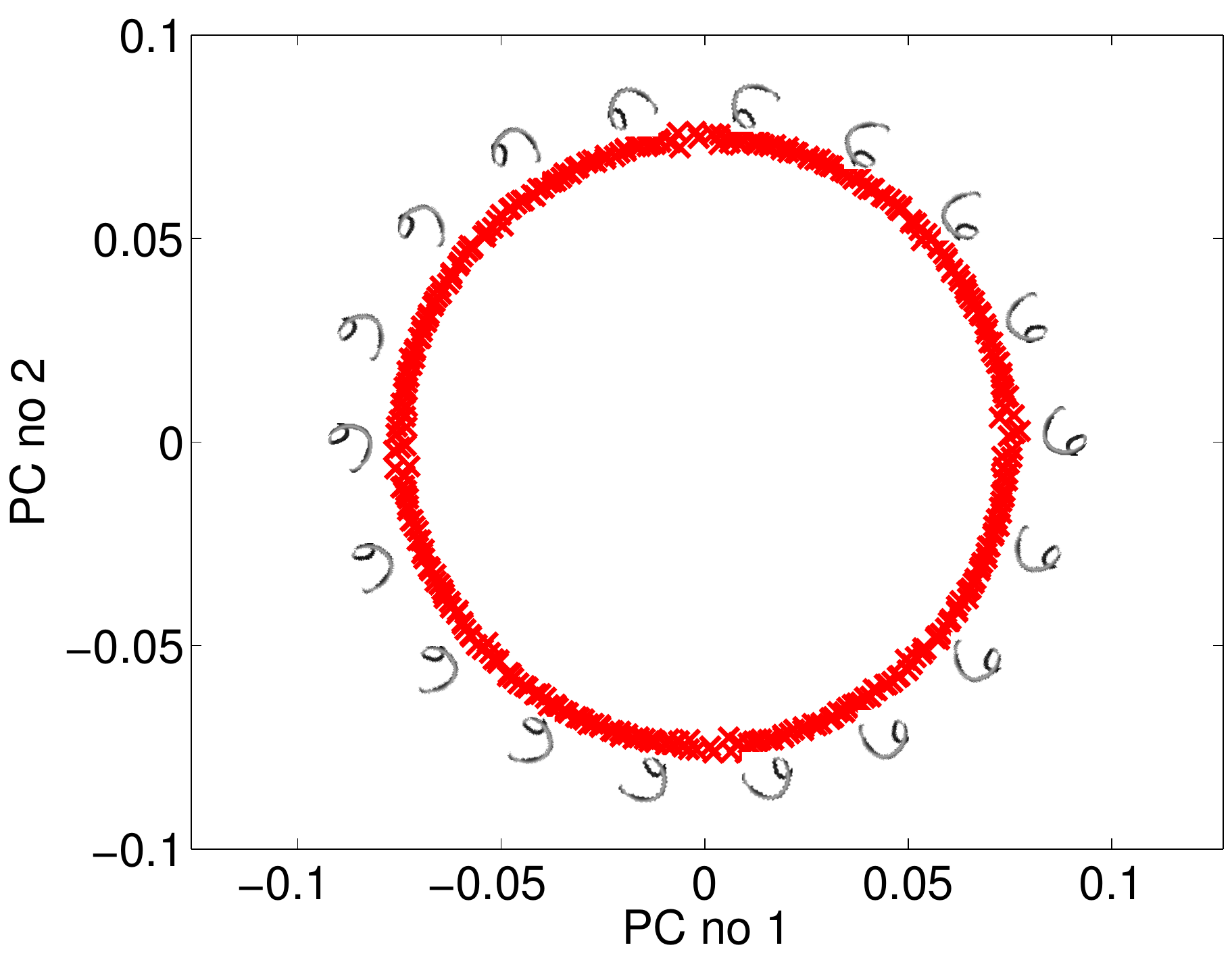}}
  
  \caption{ \small The storage capacity of high dimensional
    spaces. (a) A six from the USPS digit data set. (b) A sample from
    a simple independent pixel model of the six. There are $2^{3,648}$
    possible images. Even with an enormous number of samples from such
    a model we would never see the original six. (c) A data set
    generated by rotating the original six from (a) 360 times. The
    data is projected onto its first two principal components. These
    two principal components show us that the data lives on a circle
    in this high dimensional space. There is a small amount of noise
    due to interpolation used in the image rotation. Alongside the
    projected points we show some examples of the rotated
    sixes.}\label{fig:six}
\end{figure}

Consider the simple example of a handwritten six. The six in
\reffig{fig:six} is represented on a grid of pixels which is $64$ rows
by $57$ columns, giving a datum with 3,648 dimensions. The space in
which this digit sits contains far more than the digit. Imagine a
simple probabilistic model of the digit which assumes that each pixel
in the image is independent and is on with a given probability. We can
sample from such a model (\reffig{fig:six}(b)).

Even if we were to sample every nanosecond from now until the end of
the universe we would be highly unlikely to see the original six. The
space covered by this model is very large but fundamentally the data
lies on low dimensional embedded space. This is illustrated in
\reffig{fig:six}(c). Here a data set has been constructed by rotating
the digit 360 times in one degree intervals. The data is then
projected onto its first two principal components. The spherical
structure of the rotation is clearly visible in the projected
data. Despite the data being high dimensional, the underlying
structure is low dimensional. The objective of dimensionality
reduction is to recover this underlying structure.

Given a data set with $\numData$ data points and $\dataDim$
features associated with each data point, dimensionality reduction
involves representing the data set using $\numData$ points each with a
reduced number, $\latentDim$, of features, with
$\latentDim<\dataDim$. Dimensionality reduction is a popular approach
to dealing with high dimensional data: the hope is that while many
data sets seem high dimensional, it may be that their intrinsic
dimensionality is low like the rotated six above.

\subsection{Spectral Dimensionality Reduction}

Spectral approaches to dimensionality reduction involve taking a data
set containing $\numData$ points and forming a matrix of size
$\numData\times\numData$ from which eigenvectors are extracted to give
a representation of the data in a low dimensional space. Several
spectral methods have become popular in the machine learning community
including isomap \citep{Tenenbaum:isomap00}, locally linear embeddings
\citep[LLE,][]{Roweis:lle00}, Laplacian eigenmaps
\citep[][]{Belkin:laplacian03} and maximum variance unfolding
\citep[MVU,][]{Weinberger:learning04}. These approaches (and kernel
principal component analysis \citep[kernel
PCA,][]{Scholkopf:nonlinear98}) are closely related to classical
multidimensional scaling \citep[CMDS,][]{Mardia:multivariate79}. For a
kernel perspective on the relationships see
\cite{Ham:kernelDimred04,Bengio:outofsample03,Bengio:eigenfunctions04}.

In classical multidimensional scaling an $\numData\times\numData$
symmetric distance matrix, whose elements contain the distance between
two data points, is converted to a similarity matrix and visualized
through its principal eigenvectors. Viewed from the perspective of
CMDS the main difference between the spectral approaches developed in
the machine learning community is in the distance matrices they
(perhaps implicitly) proscribe.

In this paper we introduce a probabilistic approach to constructing
the distance matrix: maximum entropy unfolding (MEU). We describe how
isomap, LLE, Laplacian eigenmaps and MVU are related to MEU using the
unifying perspective of Gaussian random fields and CMDS.

The parameters of the model are fitted through maximum likelihood in a
Gaussian Markov random field (GRF). The random field specifies
dependencies between \emph{data points} rather than the more typical
approach which specifies dependencies between \emph{data features}. We
show that the locally linear embedding algorithm is an approximation
to maximum entropy unfolding where pseudolikelihood is maximized as an
approximation to the model likelihood. Our probabilistic perspective
inspires new dimensionality reduction algorithms. We introduce an
exact version of locally linear embedding based on an acyclic graph
structure that maximizes the true model likelihood (acyclic locally
linear embedding, ALLE). We also consider approaches to learning the
structure of the GRF through graphical regression,
\citep{Friedman:sparse08}. By L1 regularization of the dependencies we
explore whether learning the graph structure (rather than
prespecifying by nearest neighbour) improves performance. We call the
algorithm Dimensionality reduction
through Regularization of the Inverse covariance in the Log Likelihood (DRILL).

Our methods are based on maximum likelihood. Normally maximum
likelihood algorithms specify a distribution which factorizes over the
data points (each data point is independent given the model
parameters). In our models the likelihood factorizes over the features
(each feature from the data set is independent given the model
parameters). This means that maximum likelihood in our model is
consistent as the number of features increases,
$\dataDim\rightarrow\infty$ rather than the number of data
points. Alternatively, the parameters of our models become better
determined as the number of features increase, rather than the number
of data. This leads to a \emph{blessing} of dimensionality where the
parameters are better determined as the number of features
increases. This has significant implications for learning in high
dimensional data (known as the large $p$ small $n$ regime) which run
counter to received wisdom.

In \refsec{sec:meu} we derive our model through using standard
assumptions from the field of dimensionality reduction and the maximum
entropy principle \citep{Jaynes:bayes86}. We then relate the model to
other popular spectral approaches for dimensionality reduction and
show how the parameters of the model can be fitted through maximum
likelihood. This allows us to regularize the system with sparse priors
and seek MAP solutions that restrict the inter point dependencies.
Finally, we demonstrate the model (with comparisons) on two real world
data sets. First though, we will review classical multidimensional
scaling which provides the general framework through which these
approaches can be related \citep[see
also][]{Ham:kernelDimred04,Bengio:outofsample03,Bengio:eigenfunctions04}.

\subsection{Classical Multidimensional Scaling}

Given an $\numData\times\numData$ matrix of similarities,
$\kernelMatrix$, or dissimilarities, $\distanceMatrix$, between a set
of data points, multidimensional scaling considers the problem of how
to represent these data in a low dimensional space. One way of doing
this is to associate a $\latentDim$ dimensional latent vector with
each data point, $\dataVector_{i, :}$, and define a set of
dissimilarities between each latent point,
$\latentDistanceScalar_{i,j} =
\ltwoNorm{\latentVector_{i,:}-\latentVector_{j, :}}^2$ (where
$\ltwoNorm{\cdot}$ represents the L2-norm) to give a matrix
$\latentDistanceMatrix$. Here we have specified the squared distance
between each point as the dissimilarity.\footnote{It is more usual to
  specify the distance directly as the dissimilarity, however, for our
  purposes it will be more convenient to work with squared distances}

If the error for the latent representation is then taken to be the sum
of absolute values between the dissimilarity matrix entries,
\begin{equation}
  \errorFunction(\latentMatrix) = \sum_{i=1}^\numData\sum_{j=1}^{i-1}\loneNorm{\distanceScalar_{i,j} - \latentDistanceScalar_{i,j}}, \label{eq:mdsError}
\end{equation}
and we assume that the data dissimilarities also represent a squared
Euclidean distance matrix (perhaps computed in some high, maybe
infinite, dimensional space) then the optimal \emph{linear} dimensionality
reduction is given by the following procedure \citep[][pg
400]{Mardia:multivariate79},
\begin{enumerate}
\item Convert the matrix of dissimilarities to a matrix of
  similarities by taking
  $\centeredKernelMatrix=-\frac{1}{2}\centeringMatrix\distanceMatrix\centeringMatrix$
  where $\centeringMatrix = \eye -
  \numData^{-1}\onesVector\onesVector^\top$ is a centering matrix.
\item Extract the first $\latentDim$ principal eigenvectors of
  $\centeredKernelMatrix$.
\item Setting $\latentMatrix$ to these principal eigenvectors
  (appropriately scaled) gives a global minimum for the error function
  \refeq{eq:mdsError}.
\end{enumerate}

The centering matrix $\centeringMatrix$ is so called because when
applied to data in the form of a design matrix, $\dataMatrix$,
i.e. one where each row is a data point and each column is a data set
feature, the centred data matrix is recovered,
\begin{align*}
\cdataMatrix = & \dataMatrix \centeringMatrix \\
= & \dataMatrix - \numData^{-1}\dataMatrix\onesVector \onesVector^\top,\\
= & \dataMatrix - \meanVector \onesVector^\top
\end{align*}
where $\meanVector = \numData^{-1}\dataMatrix \onesVector$ is the empirical mean of the data set. 

\section{Maximum Entropy Unfolding}\label{sec:meu}

Classical multidimensional scaling provides the optimal \emph{linear}
transformation of the space in which the squared distances are
expressed. The key contribution of recently developed spectral
approaches in machine learning is to compute these distances in a
space which is nonlinearly related to the data thereby ensuring a
\emph{nonlinear} dimensionality reduction algorithm. From a machine
learning perspective this is perhaps clearest for kernel PCA
\citep{Scholkopf:nonlinear98}. In kernel PCA the squared distances are
computed between points in a Hilbert space and related to the original data
through a kernel function,
\begin{equation}
  \distanceScalar_{i,j}
  = \kernelScalar(\dataVector_{i, :}, \dataVector_{i,:}) -
  2\kernelScalar(\dataVector_{i, :}, \dataVector_{j, :}) +
  \kernelScalar(\dataVector_{j,:}, \dataVector_{j, :}).
  \label{eq:standardTransformation}
\end{equation}
For the linear kernel function, $\kernelScalar(\dataVector_{i, :},
\dataVector_{j,:})= \dataVector_{i, :}^\top\dataVector_{j, :}$ this
reduces to the squared Euclidean distance, but for nonlinear kernel
functions such as $\kernelScalar(\dataVector_{i, :}, \dataVector_{j,
  :}) = \exp(-\gamma\ltwoNorm{\dataVector_{i, :} - \dataVector_{j,
    :}}^2)$ the distances are nonlinearly related to the data
space. They are recognized as squared distances which are computed in
a ``feature space'' \citep[see
e.g.][]{Ham:kernelDimred04,Bengio:outofsample03,Bengio:eigenfunctions04}. If
we equate the kernel matrix, $\kernelMatrix$, to the similarity matrix
in CMDS then this equation is also known as the \emph{standard
  transformation} between a similarity and distance
\citep{Mardia:multivariate79}. 

Kernel PCA (KPCA) recovers an $\latentVector_{i, :}$ for each data
point and a mapping from the data space to the $\latentMatrix$
space. Under the CMDS procedure we outlined above the eigenvalue
problem is performed on the centered kernel matrix,
\[
\centeredKernelMatrix=\centeringMatrix\kernelMatrix\centeringMatrix,
\]
where $\kernelMatrix = \left[\kernelScalar(\dataVector_{i, :},
  \dataVector_{j, :})\right]_{i,j}$. This matches the procedure for
the KPCA algorithm \citep{Scholkopf:nonlinear98}\footnote{For
  stationary kernels, kernel PCA also has an interpretation as a
  particular form of \emph{metric} multidimensional scaling, see
  \cite{Williams:connection01} for details.}. However, for the
commonly used exponentiated quadratic kernel,
\[
\kernelScalar(\dataScalar_{i, :}, \dataScalar_{j, :}) = \exp(-\gamma
\ltwoNorm{\dataVector_{i, :} - \dataVector_{j, :}}^2),
\] 
KPCA actually \emph{expands} the feature space rather than reducing
the dimension \citep[see][for some examples of
this]{Weinberger:learning04}. Unless data points are repeated the
exponentiated quadratic kernel always leads to a full rank matrix,
$\kernelMatrix$, and correspondingly a rank $\numData-1$ centred
kernel matrix, $\centeredKernelMatrix$. To exactly reconstruct the
squared distances computed in feature space all but one of the
eigenvectors of $\centeredKernelMatrix$ need to be retained for our
latent representation, $\latentMatrix$. If the dimensionality of the
data, $\dataDim$, is smaller than the number of data points,
$\numData$, then we have a latent representation for our data which
has higher dimensionality than the original data.

The observation that KPCA doesn't reduce the data dimensionality
motivated the maximum variance unfolding algorithm
\citep[MVU,][]{Weinberger:learning04}. The idea in MVU is to learn a
kernel matrix that will allow for dimensionality reduction. This is
achieved by only considering \emph{local relationships} in the data. A
set of neighbors is defined (e.g. by $k$-nearest neighbors) and only
distances between neighboring data points are respected. These
distances are specified as constraints, and the other elements of the
kernel matrix are filled in by maximizing its trace,
$\tr{\kernelMatrix}$, i.e. the \emph{total variance} of the data in
feature space, while respecting the distance constraints and keeping
the resulting matrix centered. Maximizing $\tr{\kernelMatrix}$
maximizes the interpoint squared distances for all points that are
unconnected in the neighborhood graph, thereby unravelling the
manifold.

In this paper we consider an alternative maximum entropy formalism of
this problem. Since entropy is related to variance, we might expect a
similar result in the quality of the resulting algorithm, but since
maximum entropy also provides a probability distribution we should
also obtain a probabilistic model with all the associated advantages
(dealing with missing data, extensions to mixture models, fitting
parameters by Bayesian methods, combining with other probabilistic
models). Importantly, our interpretation will also enable us to relate
our algorithm to other well known spectral techniques as they each
turn out to approximate maximum entropy unfolding in some way.

\subsection{Constraints from $\distanceMatrix$ Lead to a Density on $\dataMatrix$}
\label{sec:constraintDensity}
The maximum entropy formalism \citep[see e.g.][]{Jaynes:bayes86}
allows us to derive a probability density given only a set of
constraints on expectations under that density. These constraints may be
derived from observation. In our case the observations will be squared
distances between data points, but we will derive a density over
$\dataMatrix$ directly (not over the squared distances). We will do
this by looking to constrain the expected squared inter-point
distances, $\distanceScalar_{i, j}$, of any two samples,
$\dataVector_{i, :}$ and $\dataVector_{j, :}$, from the density. This
means that while our observations may be only of the squared
distances, $\distanceScalar_{i,j}$, the corresponding density will be
over the data space that gives rise to those distances,
$p(\dataMatrix)$. Of course, once we have found the form of
probability density we are free to directly model in the space
$\dataMatrix$ or make use only of the squared distance
constraints. Direct modeling in $\dataMatrix$ turns out to be
equivalent to maximum likelihood. However, since we do not construct a
density over the squared distance matrix, modeling based on that
information alone should be thought of as maximum entropy under
distance constraints rather than maximum likelihood.

Maximum entropy is a free form optimization over all possible forms
for the density given the moment constraints we impose. In the maximum
entropy formalism, we specify the density by a free form maximization of the entropy subject to the imposed expectation constraints. The constraints
we use will correspond to the constraints applied to maximum variance
unfolding: the expectations of the squared distances between two
neighboring data points sampled from the model. 

\subsection{Maximum Entropy in Continuous Systems}
The entropy of a continuous density is normally defined as the limit
of a discrete system. The continuous distribution is discretized and
we consider the limit as the discrete bin widths approach
zero. However, as that limit is taken the a term dependent on the bin
width approaches $\infty$. Normally this is dealt with by ignoring
that term and referring to the remaining term as differential
entropy. However, the maximum entropy solution for this differential
entropy turns out to be undefined. Jaynes proposes an alternative
\emph{invariant measure} to the entropy.  For maximum entropy in
continuous systems we maximize the negative Kullback Leibler
divergence \citep[KL divergence,][]{Kullback:info51} between a base
density, $m(\dataMatrix)$, and the density of interest,
$p(\dataMatrix)$,
\[
H = -\int p(\dataMatrix) \log \frac{p(\dataMatrix)}{m(\dataMatrix)}\text{d}\dataMatrix.
\]
Maximizing this measure is equivalent to minimizing the KL divergence
between $p(\dataMatrix)$ and the base density, $m(\dataMatrix)$. Any
choice of base density can be made, but the solution will be pulled
towards the base density (through the minimization of the KL
divergence). We choose a base density to be a very broad,
spherical, Gaussian density with covariance $\gamma^{-1}\eye$. This
adds a new parameter, $\gamma$, to the system, but it will turn out
that this parameter has little affect on our analysis. Typically it
can be taken to zero or assumed small. The density that minimizes the
KL divergence under the constraints on the expectations is then 
\[
p(\dataMatrix) \propto
\exp\left(-\frac{1}{2}\tr{\gamma\dataMatrix\dataMatrix^\top}\right)\exp\left(-\frac{1}{2}\sum_{i}\sum_{j\in\neighborhood{i}}\lagrangeMultiplier_{i,j}
  \distanceScalar_{i,j}\right),
\] 
where $\neighborhood{i}$ represents the set of neighbors of data point
$i$, and $\dataMatrix=[\dataVector_{1, :}, \dots,
\dataVector_{\numData, :}]^\top\in\Re^{\numData\times\dataDim}$ is a
\emph{design matrix} containing our data. Note that we have introduced
a factor of $-1/2$ in front of our Lagrange multipliers,
$\{\lagrangeMultiplier_{i,j}\}$, for later notational convenience. We
now define the matrix $\lagrangeMultiplierMatrix$ to contain
$\lagrangeMultiplier_{i,j}$ if $i$ is a neighbor of $j$ and zero
otherwise. This allows us to write the distribution\footnote{In our
  matrix notation the Lagrange multipliers and distances are appearing
  twice inside the trace, in matrices that are constrained symmetric,
  $\lagrangeMultiplierMatrix$ and $\distanceMatrix$. The factor of
  $\frac{1}{4}$ replaces the factor of $\frac{1}{2}$ in the previous
  equation to account for this ``double counting''.} as
\[
p(\dataMatrix) \propto
\exp\left(-\frac{1}{2}\tr{\gamma\dataMatrix\dataMatrix^\top}-\frac{1}{4}\tr{\lagrangeMultiplierMatrix
    \distanceMatrix}\right).
\] 
We now introduce a matrix $\laplacianMatrix$, which has the form of a
graph Laplacian. It is symmetric and constrained to have a null space
in the constant vector, $\laplacianMatrix \onesVector =
\zerosVector$. Its off diagonal elements are given by
$-\lagrangeMultiplierMatrix$ and its diagonal elements are given by
\[
\laplacianScalar_{i,i}=\sum_{j\in\neighborhood{i}}
\lagrangeMultiplier_{i, j}
\]
to enforce the null space constraint. The null space constraint
enables us to write
\begin{equation}
  p(\dataMatrix) = \frac{\det{\laplacianMatrix + \gamma\eye}^{\frac{1}{2}}}{\twoPi^{\frac{\numData\dataDim}{2}}} \exp\left(-\frac{1}{2}\tr{(\laplacianMatrix + \gamma\eye)\dataMatrix\dataMatrix^\top}\right), \label{eq:randomField}
\end{equation}
where for convenience we have defined $\twoPi = 2\pi$. We arrive here
because the distance matrix is zero along the diagonal. This allows us
to set the diagonal elements of $\laplacianMatrix$ as we please
without changing the value of
$\tr{\laplacianMatrix\distanceMatrix}$. Our choice to set them as the
sum of the off diagonals gives the matrix a null space in the constant
vector enabling us to use the fact that
\[
\distanceMatrix = \onesVector\diag{\dataMatrix\dataMatrix^\top}^\top -
2\dataMatrix\dataMatrix^\top +
\diag{\dataMatrix\dataMatrix^\top}\onesVector^\top
\] 
(where the operator $\diag{\mathbf{A}}$ forms a vector from the
diagonal of $\mathbf{A}$) to write
\[
-\tr{\lagrangeMultiplierMatrix\distanceMatrix}=\tr{\laplacianMatrix\distanceMatrix}
=
\tr{\laplacianMatrix\onesVector\diag{\dataMatrix\dataMatrix^\top}^\top
  - 2\laplacianMatrix\dataMatrix\dataMatrix^\top +
  \diag{\dataMatrix\dataMatrix^\top}\onesVector^\top\laplacianMatrix}
= -2\tr{\laplacianMatrix\dataMatrix\dataMatrix^\top},
\] 
which in turn allows us to recover \refeq{eq:randomField}.  This
probability distribution is a \emph{Gaussian random field}. It can also be
written as
\[ 
p(\dataMatrix) = \prod_{j=1}^\dataDim\frac{\det{\laplacianMatrix +
    \gamma\eye}^{\frac{1}{2}}}{\twoPi^{\frac{\numData}{2}}}
\exp\left(-\frac{1}{2}\dataVector_{:,j}^\top(\laplacianMatrix +
  \gamma\eye)\dataVector_{:, j}\right),
\]
which emphasizes the independence of the density across data features.

\subsection{Gaussian Markov Random Fields}

Multivariate Gaussian densities are specified by their mean,
$\meanVector$, and a covariance matrix, $\covarianceMatrix$. A
standard modeling assumption is that data is draw independently from
identical Gaussian densities. For this case the likelihood of the
data, $p(\dataMatrix)$, will be factorized across the individual data
points,
\[
p(\dataMatrix) = \prod_{i=1}^\numData p(\dataVector_{i, :}) = \prod_{i=1}^\numData \gaussianDist{\dataVector_{i, :}}{\meanVector}{\covarianceMatrix}
\]
and the mean and covariance of the Gaussian are estimated by
maximizing the log likelihood of the data. The covariance matrix is
symmetric and positive definite. It contains
$\frac{\dataDim(\dataDim+1)}{2}$ parameters. However, if the number of
data points, $\numData$, is small relative to the number of features
$\dataDim$, then the parameters may not be well determined. For this
reason we might seek a representation of the covariance matrix which
has fewer parameters. One option is a low rank representation,
\[
\covarianceMatrix = \weightMatrix\weightMatrix^\top + \mathbf{D},
\]
where $\mathbf{D}$ is a diagonal matrix and $\weightMatrix \in
\Re^{\dataDim\times\latentDim}$. This is the representation underlying
factor analysis, and if $\mathbf{D}=\sigma^2\eye$, probabilistic
principal component analysis \citep[PPCA,][]{Tipping:probpca99}. For
PPCA there are $\dataDim\latentDim + 1$ parameters in the covariance
representation. 

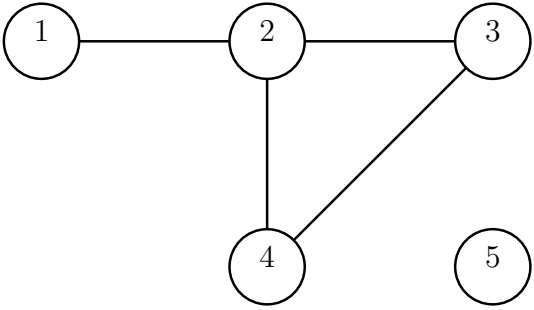
\begin{figure}
\begin{picture}(0,0)%
\includegraphics{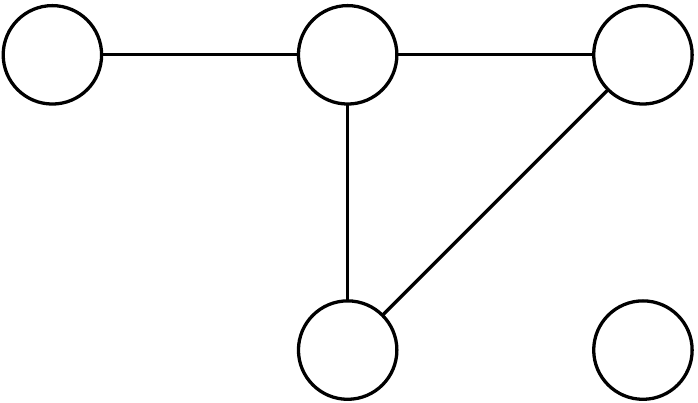}%
\end{picture}%
\setlength{\unitlength}{4144sp}%
\begingroup\makeatletter\ifx\SetFigFont\undefined%
\gdef\SetFigFont#1#2#3#4#5{%
  \reset@font\fontsize{#1}{#2pt}%
  \fontfamily{#3}\fontseries{#4}\fontshape{#5}%
  \selectfont}%
\fi\endgroup%
\begin{picture}(3180,1830)(661,-1651)
\put(901,-61){\makebox(0,0)[b]{\smash{{\SetFigFont{12}{14.4}{\rmdefault}{\mddefault}{\updefault}{\color[rgb]{0,0,0}$1$}%
}}}}
\put(2251,-1411){\makebox(0,0)[b]{\smash{{\SetFigFont{12}{14.4}{\rmdefault}{\mddefault}{\updefault}{\color[rgb]{0,0,0}$4$}%
}}}}
\put(3601,-1411){\makebox(0,0)[b]{\smash{{\SetFigFont{12}{14.4}{\rmdefault}{\mddefault}{\updefault}{\color[rgb]{0,0,0}$5$}%
}}}}
\put(2251,-61){\makebox(0,0)[b]{\smash{{\SetFigFont{12}{14.4}{\rmdefault}{\mddefault}{\updefault}{\color[rgb]{0,0,0}$2$}%
}}}}
\put(3601,-61){\makebox(0,0)[b]{\smash{{\SetFigFont{12}{14.4}{\rmdefault}{\mddefault}{\updefault}{\color[rgb]{0,0,0}$3$}%
}}}}
\end{picture}%
\caption{Graph representing conditional relationships between $\dataDim=5$ features from a Gaussian Markov random field. Here the 5th feature is independent of the others. Feature 1 is conditionally dependent on 2, feature 2 is conditional dependent on 1, 3 and 4. Feature 3 is conditionally dependent on 2 and 4, and feature 4 is conditionally dependent on 2 and 3. }\label{fig:conditonalDependencies}
\end{figure}

An alternative approach, and one that is particularly
popular in spatial systems, is to assume a sparse \emph{inverse}
covariance matrix, known as the precision matrix, or information
matrix. In this representation we consider each feature to be a vertex
in a graph. If two vertices are unconnected they are conditionally
independent in the graph. In \reffig{fig:conditonalDependencies} we show a simple example graph where the precision matrix is
\begin{align*}
\kernelMatrix^{-1} = \precisionMatrix = \left[\begin{matrix} \precisionScalar_{1,1} & \precisionScalar_{1, 2} & 0 & 0 & 0\\
\precisionScalar_{2, 1} & \precisionScalar_{2, 2} & \precisionScalar_{2, 3} & \precisionScalar_{2, 4} & 0 \\
0 & \precisionScalar_{3, 2} & \precisionScalar_{3, 3} & \precisionScalar_{3, 4} & 0 \\
0 & \precisionScalar_{4, 2} & \precisionScalar_{4, 3} & \precisionScalar_{4, 4} & 0 \\
0 & 0 & 0 & 0 & \precisionScalar_{5, 5}
\end{matrix}\right].
\end{align*}
Zeros correspond to locations where there are no edges between vertices in the graph. 

If each feature is constrained to only have $\numNeighbors$ neighbors
in the graph, then the inverse covariance (and correspondingly the
covariance) is only parameterized by $\numNeighbors\dataDim +
\dataDim$ parameters. So the GRF provides an alternative approach to
reducing the number of parameters in the covariance matrix.
\subsection{Independence Over Data Features}

The Gaussian Markov random field (GRF) for maximum entropy unfolding
is unusual in that the independence is being expressed over data
features (in the $\dataDim$-dimensional direction) instead of over
data points (in the $\numData$-dimensional direction). This means that
our model assumes that data \emph{features} are independently and
identically distributed (i.i.d.)  given the model parameters. The
standard assumption for Gaussian models is that data \emph{points} we
are expressing conditional probability densities between data points
are i.i.d. given the parameters. This specification cannot be thought
of as ``the wrong way around'' as it is merely a consequence of the
constraints we chose to impose on the maximum entropy solution. If
those constraints are credible, then this model is also credible. This
isn't the first model proposed for which the independence assumptions
are reversed. Such models have been proposed formerly in the context
of semi-supervised learning \citep{Zhu:graphsemi03}, probabilistic
nonlinear dimensionality reduction
\citep{Lawrence:gplvm03,Lawrence:pnpca05} and in models that aim to
discover structural form from data \citep{Kemp:form08}.

\subsection{Maximum Likelihood and Blessing of Dimensionality}

Once the form of a maximum entropy density is determined, finding the
Lagrange multipliers in the model is equivalent to maximizing the
likelihood of the model, where the Lagrange multipliers are now
considered to be parameters. The theory underpinning maximum
likelihood is broad and well understood, but much of it relies on
assuming independence across data points rather than data
features. For example, maximum likelihood with independence across
data points can be shown to be consistent by viewing the objective as
a sample based approximation to the Kullback-Leibler (KL) divergence
between the true data generating density, $\tilde{p}(\dataVector)$,
and our approximation $p(\dataVector|\parameterVector)$ which in turn
depends on parameters, $\parameterVector$. Taking the expectations
under the generating density this KL divergence is written as
\[
\KL{\tilde{p}(\dataVector)}{p(\dataVector)} = \int
\tilde{p}(\dataVector)\log \tilde{p}(\dataVector)
\text{d}\dataVector - \int \tilde{p}(\dataVector)\log
p(\dataVector) \text{d}\dataVector.
\]
Given $\numData$ sampled data points from $\tilde{p}(\dataVector)$,
$\left\{\dataVector_{i, :}\right\}$ we can write down a sample based
approximation to the KL divergence in the form
\[
\KL{\tilde{p}(\dataVector)}{p(\dataVector)} \approx -\frac{1}{\numData}\sum_{i=1}^\numData \log p(\dataVector_{i, :}|\parameterVector) + \text{const.},
\]
where the constant term derives from the entropy of the generating
density, which whilst unknown, does not depend on our model
parameters. Since the sample based approximation is known to become
exact in the large sample limit, and the KL divergence has a global
minimum of zero \emph{only} when the generating density and our approximation
are identical, we know that, \emph{if} the generating density falls within
our chosen class of densities, maximum likelihood will reveal it in
the large data limit. The global maximum of the likelihood will
correspond to a global minimum of the KL divergence. Further, we can
show that as we approach this limit, \emph{if} the total number of parameters
is fixed, our parameter values, $\parameterVector$, will become better
determined \citep[see e.g.][pg 126]{Wasserman:all03}. Since the number
of parameters is often related to data dimensionality, $\dataDim$,
this implies that for a given data dimensionality, $\dataDim$, we
require a large number of data points, $\numData$, to have confidence
we are approaching the large sample limit and our model's parameters
will be well determined. We refer to this model set up as the
\emph{sampled-points} formalism.

The scenario described above does not apply for the situation where we
have independence across data features. In this situation we
construct an alternative consistency argument, but it is based around a
density which describes correlation between data points instead of
data features. This model is independent across data features. Models
of this type can occur quite naturally. Consider the following
illustrative example from cognitive science \citep{Kemp:form08}. We
wish to understand the relationship between different animals as more
information about those animals' features is uncovered. There are 33
species in the group, and information is gained by unveiling features
of the animals. A model which assumes independence over animals would
struggle to incorporate additional feature information (such as
whether or not the animal has feet, or whether or not it lives in the
ocean). A model which assumes independence across features handles
this situation naturally. However, to show the consistency of the
model we must now think of our model as a generative model for data
features, $\tilde{p}(\dataVector^\prime)$, rather than data
points. Our approximation to the KL divergence still applies,
\[
\KL{\tilde{p}(\dataVector^\prime)}{p(\dataVector^\prime)} \approx -\frac{1}{\dataDim}\sum_{j=1}^\dataDim \log p(\dataVector_{:, i}|\parameterVector) + \text{const.},
\]
but now the sample based approximation is based on independent samples
of features (in the animal example, whether or not it has a beak, or
whether the animal can fly), instead of samples of data points. This
model will typically have a parameter vector that increases in size
with the data set size, $\numData$ (in that sense it is
non-parametric), rather than the data dimensionality, $\dataDim$. The
model is consistent as the number of features becomes large, rather
than data points. For our Gaussian random field, the number of
parameters increases linearly with the number of data points, but
doesn't increase with the number of data (each datum requires
$\mathcal{O}(\numNeighbors)$ parameters to connect with $\numNeighbors$
neighbors). However, as we increase features there is no corresponding
increase in parameters. In other words as the number of features
increases there is a clear \emph{blessing of dimensionality}. We refer
to this model set up as the \emph{sampled-features} formalism.

There is perhaps a deeper lesson here in terms of how we should
interpret such consistency results. In the sampled-points formalism,
as we increase the number of data points, the parameters become better
determined. In the sampled-features formalism, as we increase the
number of features, the parameters become better determined. However,
for consistency results to hold, the class of models we consider must
include the actual model that generated the data. If we believe that
``Essentially, all models are wrong, but some are useful'' \citep[][pg
424]{Box:empirical87} we may feel that encapsulating the right model
within our class is a practical impossibility. Given that, we might
pragmatically bias our choice somewhat to ensure utility of the
resulting model. From this perspective, in the large $p$ small $n$
domain, the sampled-features formalism is attractive. A practical
issue can arise though. If we wish to compute the likelihood of an out
of sample data-point, we must first estimate the parameters associated
with that new data point. This can be problematic. Of course, for the
sampled-points formalism the same problem exists when you wish to
include an out of sample data-feature in your model (such as in the
animals example in \cite{Kemp:form08}). Unsurprisingly, addressing
this issue for spectral methods is nontrivial
\citep{Bengio:outofsample03}.

\subsubsection{Parameter Gradients}

We can find the parameters, $\lagrangeMultiplierMatrix$, through
maximum likelihood on the Gaussian Markov random field given in
\refeq{eq:randomField}. Some algebra shows that the gradient of each
Lagrange multiplier is given by,
\[
\diff{\log p(\dataMatrix)}{\lagrangeMultiplier_{i,j}} = \frac{1}{2}\expectationDist{\distanceScalar_{i,j}}{p(\dataMatrix)} - \frac{1}{2}\distanceScalar_{i,j},
\]
where $\expectationDist{}{p(\cdot)}$ represents an expectation under
the distribution $p(\cdot)$. This result is a consequence of the
maximum entropy formulation: the Lagrange multipliers have a gradient
of zero when the constraints are satisfied.  To compute gradients we
need the expectation of the squared distance given by
\[
\expectation{\distanceScalar_{i,j}} =
\expectation{\dataScalar_{i, :}^\top\dataScalar_{i, :}} -
2\expectation{\dataScalar_{i, :}^\top\dataScalar_{j, :}} +
\expectation{\dataScalar_{j, :}^\top\dataScalar_{j, :}},
\]
which we can compute directly from the covariance matrix of the GRF,
$\kernelMatrix = \left(\laplacianMatrix + \gamma\eye\right)^{-1}$,
\[
\expectation{\distanceScalar_{i,j}} = \frac{\dataDim}{2}\left(\kernelScalar_{i,i} - 2\kernelScalar_{i,j} + \kernelScalar_{j,j}\right).
\]
This is immediately recognized as a scaled version of the standard
transformation between distances and similarities (see
\refeq{eq:standardTransformation}). This relationship arises naturally
in the probablistic model. Every GRF has an associated interpoint
distance matrix. It is this matrix that is being used in CMDS. The
machine learning community might interpret this as the relationship
between distances in ``feature space'' and the kernel function. Note
though that here (and also in MVU) each individual element of the
kernel matrix \emph{cannot} be represented only as a function of the
corresponding two data points (i.e. we can't represent them as
$\kernelScalar_{i,j}=\kernelScalar(\dataVector_{i,:},
\dataVector_{j,:})$, where each $k_{i,j}$ is a function \emph{only} of
the $i$ and $j$th data points). Given this we feel it is more correct
to think of this matrix as a covariance matrix induced by our
specification of the random field rather than a true Mercer kernel. We
use the notation $\kernelScalar_{i,j}$ to denote an element of such a
covariance (or similarity matrix) and only use $\kernelScalar(\cdot,
\cdot)$ notation when the value of the similarity matrix can be
explicitly represented as a Mercer kernel.

\paragraph{The Base Density Parameter} One role of the base density
parameter, $\gamma$, is to ensure that the precision matrix is
positive definite. Recall that the Laplacian has a null space in the
constant vector, implying that $\kernelMatrix\onesVector =
\gamma^{-1}$, which becomes infinite as $\gamma\rightarrow 0$. This
reflects an insensitivity of the covariance matrix to the data mean,
and this in turn arises because that information is lost when we
specify the expectation constraints only through interpoint
distances. In practise though, $\kernelMatrix$ is always centred
before its eigenvectors are extracted, $\centeredKernelMatrix =
\centeringMatrix \kernelMatrix \centeringMatrix$, resulting in
$\centeredKernelMatrix\onesVector = \zerosVector$ so $\gamma$ has no
effect on the final visualization. In some cases, it may be necessary
to set $\gamma$ to a small non-zero value to ensure stability of the
inverse $\laplacianMatrix + \gamma\eye$. In these cases we set it to
$\gamma=1\times10^{-4}$ but in many of the comparisons we make to
other spectral algorithms below we take it to be zero.

\paragraph{Number of Model Parameters} If $\numNeighbors$ neighbors
are used for each data point there are $O(\numNeighbors\numData)$
parameters in the model, so the model is nonparametric in the sense
that the number of parameters increases with the number of data. For
the parameters to be well determined we require a large number of
features, $\dataDim$, for each data point, otherwise we would need to
regularize the model (see \refsec{sec:drill}). This implies that the
model is well primed for the so-called ``large $p$ small $n$ domain''.

Once the maximum likelihood solution is recovered the data can be
visualized, as for MVU and kernel PCA, by looking at the eigenvectors
of the centered covariance matrix
$\centeringMatrix\kernelMatrix\centeringMatrix$. We call this
algorithm maximum entropy unfolding (MEU).

\paragraph{Positive Definite Constraints} The maximum variance unfolding (MVU) algorithm maximizes the trace of the covariance matrix
\[
\tr{\kernelMatrix}=\sum_{i=1}^\numData \eigenvalue_i,
\]
subject to constraints on the elements of $\kernelMatrix$ arising from
the squared distances. These constraints are linear in the elements of
$\kernelMatrix$. There is a further constraint on $\kernelMatrix$, that it should be positive semi-definite. This means MVU can be optimized through a  a semi-definite program. In contrast MEU cannot be optimized through a semi-definite program because the objective
linear in $\kernelMatrix$. This implies we need to find other
approaches to maintaining the positive-definite constraint on
$\kernelMatrix$.  Possibilities include exploiting the fact that if
the Lagrange multipliers are constrained to be positive the system is
``attractive'' and this guarantees a valid covariance \citep[see
e.g.][pg 255]{Koller:book09}. Although now (as in a suggested variant
of the MVU) the distance constraints would be inequalities. Another
alternative would be to constrain $\laplacianMatrix$ to be diagonally
dominant through adjusting $\gamma$. We will also consider two further
approaches in \refsec{sec:lle} and
\refsec{sec:drill}.\label{sec:positivedefinite}

\paragraph{Non-linear Generalizations of PCA}

Kernel PCA provides a non-linear generalization of PCA. This is
achieved by 'kernelizing' the principal coordinate analysis algorithm:
replacing data point inner products with a kernel function. Maximum variance
unfolding and maximum entropy unfolding also provide non linear
generalizations of PCA. For these algorithms, if we increase the
neighborhood size to $\numNeighbors = \numData-1$, then all squared
distances implied by the GRF model are constrained to match the
observed inter data point squared distances and $\laplacianMatrix$
becomes non-sparse. Classical multidimensional scaling on the
resulting squared distance matrix is known as principal coordinate
analysis and is equivalent to principal component analysis
\citep[see][]{Mardia:multivariate79}\footnote{In this case CMDS proceeds by computing the eigendecomposition of the centred negative squared distance matrix, which is the eigendecomposition of the centred inner product matrix as is performed for principal coordinate analysis.}.

\subsection{Relation to Laplacian Eigenmaps}

Laplacian eigenmaps is a spectral algorithm introduced by
\cite{Belkin:laplacian03}. In the Laplacian eigenmap procedure a
neighborhood is first defined in the data space. Typically this is
done through nearest neighbor algorithms or defining all points within
distance $\epsilon$ of each point to be neighbors. In Laplacian
eigenmaps a symmetric sparse (possibly weighted) adjacency matrix,
$\weightedAdjacencyMatrix\in\Re^{\numData \times \numData}$, is
defined whose $i, j$th element, $\weightedAdjacencyScalar_{i, j}$ is
non-zero if the $i$th and $j$th data points are
neighbors. \citeauthor{Belkin:laplacian03} argue that a good \emph{one
  dimensional embedding} is one where the latent points,
$\latentMatrix$ minimize
\[
\errorFunction(\latentMatrix) =
\frac{1}{4}\sum_{i=1}^\numData\sum_{j=1}^\numData
\weightedAdjacencyScalar_{i,j}(\latentScalar_{i} -
\latentScalar_{j})^2,
\]
For a multidimensional embedding we can rewrite this objective in
terms of the squared distance between two latent points,
$\latentDistanceScalar_{i,j} = \ltwoNorm{\latentVector_{i, :} -
  \latentVector_{j, :}}^2$, as
\[
\errorFunction(\latentMatrix) =
\frac{1}{4}\sum_{i=1}^\numData\sum_{j=1}^\numData
\weightedAdjacencyScalar_{i, j} \latentDistanceScalar_{i, j}.
\]
The motivation behind this objective function is that neighboring
points have non-zero entries in the adjacency matrix, therefore their
inter point squared distances in latent space need to be minimized. In
other words points which are neighbors in data space will be kept
close together in the latent space. The objective function can be
rewritten in matrix form as
\[
\errorFunction(\latentMatrix) =
\frac{1}{4}\trace{\weightedAdjacencyMatrix\latentDistanceMatrix}.
\]
Squared Euclidean distance matrices of this type can be rewritten in
terms of the original vector space by introducing the Laplacian
matrix. Introducing the degree matrix, $\degreeMatrix$, which is
diagonal with entries, $\degreeScalar_{i, i} =
\sum_{j}\weightedAdjacencyMatrix_{i, j}$ the Laplacian associated with
the neighborhood graph can be written
\[
\laplacianMatrix = \degreeMatrix - \weightedAdjacencyMatrix
\]
and the error function can now be written directly in terms of the
latent coordinates,
\[
\errorFunction(\latentMatrix) =
\frac{1}{2}\trace{\laplacianMatrix\latentMatrix \latentMatrix^\top}
\]
by exploiting the null space of the Laplacian
($\laplacianMatrix\onesVector = \zerosVector$) as we saw in
\refsec{sec:constraintDensity}.

Let's consider the properties of this objective. Since the error
function is in terms of interpoint distances, it is insensitive to
translations of the embeddings. The mean of the latent projections is
therefore undefined. Further, there is a trivial solution for this
objective. If the latent points are all placed on top of one another
the interpoint distance matrix will be all zeros. To prevent this
collapse \citeauthor{Belkin:laplacian03} suggest that each dimension
of the latent representation is constrained,
\[
\latentVector_{:, i}^\top \degreeMatrix \latentVector_{:, i} = 1.
\]
Here the degree matrix, $\degreeMatrix$, acts to scale each data point
so that points associated with a larger neighborhood are pulled
towards the origin.

Given this constraint the objective function is minimized for a
$\latentDim$ dimensional space by the generalized eigenvalue problem,
\[
\laplacianMatrix \eigenvector_i = \eigenvalue_i\degreeMatrix \eigenvector_i,
\]
where $\eigenvalue$ is an eigenvalue and $\eigenvector$ is its
associated eigenvector. The smallest eigenvalue is zero and is
associated with the constant eigenvector. This eigenvector is
discarded, whereas the eigenvectors associated with the next
$\latentDim$ smallest eigenvalues are retained for the embedding. So
we have,
\[
\latentVector_{:, i} = \eigenvector_{i+1} \quad \text{for} \quad i=1 .. \latentDim
\]
if we assume that eigenvalues are ordered according to magnitude with
the smallest first.

Note that the generalized eigenvalue problem underlying Laplacian
eigenmaps can be readily converted to the related, symmetric, eigenvalue
problem.
\begin{equation}
\normalizedLaplacianMatrix \eigenvectwo_i = \eigenvalue_i \eigenvectwo_i
\label{eq:normalizedEigenvalueProblem}
\end{equation}
where $\normalizedLaplacianMatrix$ is the \emph{normalized Laplacian
  matrix}, 
\[
\normalizedLaplacianMatrix = \degreeMatrix^{-\frac{1}{2}}
\laplacianMatrix \degreeMatrix^{-\frac{1}{2}} = \eye -
\degreeMatrix^{-\frac{1}{2}}\weightedAdjacencyMatrix\degreeMatrix^{-\frac{1}{2}}
\]
and the relationship between the eigenvectors is through scaling by
the degree matrix, $\eigenvectwo_i =
\degreeMatrix^{\frac{1}{2}}\eigenvector_i$ (implying $\eigenvectwo_i^\top\eigenvectwo_i = 1$). The eigenvalues remain unchanged in each case.

\subsubsection{Parameterization in Laplacian Eigenmaps}

In Laplacian eigenmaps the adjacency matrix can either be unweighted
(\citeauthor{Belkin:laplacian03} refer to this as the simple-minded
approach) or weighted according to the distance between two data
points,
\begin{equation}
\weightedAdjacencyScalar_{i,j} = \exp\left(-\frac{\ltwoNorm{\dataVector_{i, :} - \dataVector_{j, :}}^2}{2\lengthScale^2}\right),\label{eq:adjacencyParameterization}
\end{equation} 
which is justified by analogy between the discrete graph Laplacian and
its continuous equivalent, the Laplace Beltrami operator \citep{Belkin:laplacian03}.

\subsubsection{Relating Laplacian Eigenmaps to MEU}

The relationship of MEU to Laplacian eigenmaps is starting to become
clear. In Laplacian eigenmaps a graph Laplacian is specified across
the data points just as in maximum entropy unfolding. In classical
multidimensional scaling, as applied in MEU and MVU, the eigenvectors
associated with the largest eigenvalues of the centred covariance
matrix,
\begin{equation}
\centeredKernelMatrix  = \centeringMatrix \left(\laplacianMatrix + \gamma \eye\right)^{-1} \centeringMatrix \label{eq:kernelEigenvalueProblem}
\end{equation}
are used for visualization. In Laplacian eigenmaps the smallest eigenvectors of
$\laplacianMatrix$ are used, disregarding the eigenvector associated
with the null space. 

Note that if we define the eigendecomposition of the covariance in the GRF as
\[
\kernelMatrix =
\eigenvectorMatrix\eigenvalueMatrix\eigenvectorMatrix^\top
\]
it is easy to show that the eigendecomposition of the associated
Laplacian matrix is
\[
\laplacianMatrix = \eigenvectorMatrix \left(\eigenvalueMatrix^{-1}-\gamma\eye\right)\eigenvectorMatrix^\top.
\]
We know that the smallest eigenvalue of $\laplacianMatrix$ is zero
with a constant eigenvector. That implies that the largest eigenvalue
of $\kernelMatrix$ is $\gamma^{-1}$ and is associated with a constant
eigenvector. However, we don't use the eigenvectors of $\kernelMatrix$
directly. We first apply the centering operation in
\refeq{eq:kernelEigenvalueProblem}. This projects out the constant
eigenvector, but leaves the remaining eigenvectors and eigenvalues
intact. 

To make the analogy with Laplacian eigenmaps direct we consider the
formulation of its eigenvalue problem with the normalized graph
Laplacian as given in
\refeq{eq:normalizedEigenvalueProblem}. Substituting the normalized
graph Laplacian into our covariance matrix, $\kernelMatrix$, we see
that for Laplacian eigenmaps we are visualizing a Gaussian random
field with a covariance as follows,
\[
\kernelMatrix = (\normalizedLaplacianMatrix + \gamma
\eye)^{-1}.
\]
Naturally we could also consider a variant of the algorithm which used
the unnormalized Laplacian directly, $ \kernelMatrix =
(\laplacianMatrix + \gamma \eye)^{-1}$. \label{sec:LEunnormalized}
Under the Laplacian eigenmap formulation that would be equivalent to
preventing the collapse of the latent points by constraining
$\latentVector_{:, i}^\top\latentVector_{:, i} = 1$ instead of
$\latentVector_{:, i}^\top\degreeMatrix\latentVector_{:, i} = 1$.
 
This shows the relationship between the eigenvalue problems for
Laplacian eigenmaps and CMDS. The principal eigenvalues of
$\kernelMatrix$ will be the smallest eigenvalues of
$\laplacianMatrix$. The very smallest eigenvalue of $\laplacianMatrix$
is zero and associated with the constant eigenvector. However, in CMDS
this would be removed by the centering operation and in Laplacian
eigenmaps it is discarded. Once the parameters of the Laplacian have
been set CMDS is being performed to recover the latent variables in
Laplacian eigenmaps.

\subsubsection{Laplacian Eigenmaps Summary}

The Laplacian eigenmaps procedure doesn't fit parameters through maximum
likelihood. It uses analogies with the continuous Laplace Beltrami
operator to set them via the Gaussian-like relationship in
\refeq{eq:adjacencyParameterization}. This means that the local
distance constraints are not a feature of Laplacian eigenmaps. The
implied squared distance matrix used for CMDS will not preserve the
interneighbor distances as it will for MVU and MEU. In fact since the
covariance matrix is never explicitly computed it is not possible to
make specific statements about what these distances will be in
general. However, Laplacian eigenmaps gains significant computational
advantage by not representing the covariance matrix explicitly. No
matrix inverses are required in the algorithm and the resulting
eigenvalue problem is sparse. This means that Laplacian eigenmaps can
be applied to much larger data sets than would be possible for MEU or
MVU.

\subsection{Relation of MEU to Locally Linear Embedding}\label{sec:lle}

The locally linear embedding \citep[LLE][]{Roweis:lle00} is a
dimensionality reduction that was originally motivated by the idea
that a non-linear manifold could be approximated by small linear
patches. If the distance between data points is small relative to the
curvature of the manifold at a particular point, then the manifold
encircling a data point and its nearest neighbors may be approximated
locally by a linear patch.  This idea gave rise to the locally linear
embedding algorithm. First define a local neighborhood for each data
point and find a set of linear regression weights that allows each
data point to be reconstructed by its neighbors. Considering the $i$th
data point, $\dataVector_{i, :}$ and a vector of reconstruction
weights, $\weightVector_{:, i}$, associated with that data point a
standard least squares regression objective takes the form,
\begin{equation}
\errorFunction(\weightVector_{:, i}) = \frac{1}{2}\ltwoNorm{\dataVector_{i, :} - \sum_{j\in\neighborhood{i}}\dataVector_{j, :}\weightScalar_{j, i}}^2, \label{eq:lleSinglePointObjective}
\end{equation}
for each data point. Here the sum over the reconstruction weights,
$\weightVector_{:, j}$ is restricted to data points,
$\left\{\dataVector_{j, :}\right\}_{j\in \neighborhood(i)}$, which are
in the neighborhood of the data point of interest, $\dataVector_{i,
  :}$. \citeauthor{Roweis:lle00} point out that the objective function
in \refeq{eq:lleSinglePointObjective} is invariant to rotation and
rescaling of the data. If we rotate each data
vector in \refeq{eq:lleSinglePointObjective} the objective does
not change. If data are rescaled, e.g. multiplied by a factor
$\alpha$, then the objective is simply rescaled by a factor
$\alpha^2$. However, the objective is not invariant to
translation. For example if we were to translate the data,
$\cdataVector_{i, :} = \dataVector_{i, :} - \meanVector$, where
$\meanVector$ could be the sample mean of our data set (or any other
translation), we obtain the following modified objective,
\[
\errorFunction(\weightVector_{:, i}) =
\frac{1}{2}\ltwoNorm{\cdataVector_{i, :} +\meanVector -
  \sum_{j\in\neighborhood{i}}\cdataVector_{j,
    :}\weightScalar_{j, i}
  -\meanVector\sum_{j\in\neighborhood{i}}\weightScalar_{j, i}}^2,
\]
which retains a dependence on $\meanVector$. \citeauthor{Roweis:lle00} point out that if we constrain
$\sum_{j\in\neighborhood{i}}\weightScalar_{j, i} = 1$ the terms
involving $\meanVector$ cancel and we recover the original
objective. Imposing this constraint on the regression weights (which
can also be written $\weightVector_{:, i}^\top\onesVector = 1$),
ensures the objective is translation invariant. 

To facilitate the comparison with the maximum entropy unfolding
algorithm we now introduce an alternative approach to enforcing
translation invariance. Our approach generalizes the LLE
algorithm. First of all we introduce a new matrix
$\laplacianFactor$. We define off diagonal elements of this matrix to
be given by $\weightMatrix$ so we have $\laplacianFactorScalar_{j, i}
= \weightScalar_{j, i}$ for $i\neq j$. We set the diagonal elements of $\laplacianFactor$ to be the negative sum of the off diagonal columns, so we have
$\laplacianFactorScalar_{i, i} = -
\sum_{j\in\neighborhood{i}}\weightScalar_{j, i}$. We can then rewrite
the objective in \refeq{eq:lleSinglePointObjective} as,
\[
\errorFunction(\weightVector_{:, i}) =
\frac{1}{2}\ltwoNorm{\dataMatrix^\top \laplacianFactorVector_{:, i}}^2 = \laplacianFactorVector_{:, i}^\top \dataMatrix\dataMatrix^\top\laplacianFactorVector_{:, i},
\]
which is identical to \refeq{eq:lleSinglePointObjective} if
$\laplacianFactorScalar_{i, i}$ is further constrained to 1. However, even if this
constraint isn't imposed, the translational invariance is
retained. This is clear if we rewrite the objective in terms of the non-zero elements of $\laplacianFactorVector_{:, i}$,
\begin{align*}
  \errorFunction(\weightVector_{:, i}) &=
  \frac{\laplacianFactorScalar_{i, i}^2}{2}\ltwoNorm{\dataVector_{i,
      :} + \sum_{j\in\neighborhood{i}}\dataVector_{j,
      :}\frac{\laplacianFactorScalar_{j,
        i}}{\laplacianFactorScalar_{i, i}}}^2 \\
& =
  \frac{\laplacianFactorScalar_{i, i}^2}{2}\ltwoNorm{\dataVector_{i,
      :} - \sum_{j\in\neighborhood{i}}\dataVector_{j,
      :}\weightScalar_{j, i}}^2
\end{align*}
where 
\[
\weightScalar_{j, i} =
-\frac{\laplacianFactorScalar_{j,
    i}}{\laplacianFactorScalar_{i,i}}
\]
 and by definition of $\laplacianFactorScalar_{i, i}$ we have
$\sum_{j\in\neighborhood{i}}\weightScalar_{j, i}=1$. 
We now see that up to a scalar factor, $\laplacianFactorScalar_{i, i}^2$,
this equation is identical to \refeq{eq:lleSinglePointObjective}. 

This form of the objective also shows us that
$\laplacianFactorScalar_{i,i}$ has the role of scaling each data
point's contribution to the overall objective function (rather like
the degree, $\degreeScalar_{i, i}$ would do in the unnormalized
variant of Laplacian eigenmaps we discussed in
\refsec{sec:LEunnormalized}).

The objective function is a least squares formulation with particular
constraints on the regression weights, $\laplacianFactorVector_{:,
  i}$. As with all least squares regressions, there is an underlying
probabilistic interpretation of the regression which suggests Gaussian
noise. In our objective function the variance of the Gaussian noise
for the $i$th data point is given by $\laplacianFactorScalar_{i,
  i}^{-2}$. We can be a little more explicit about this by writing
down the error as the negative log likelihood of the equivalent
Gaussian model. This then includes a normalization term, $\log \laplacianFactorScalar^2_{i, i}$, which is zero in standard LLE where $\laplacianFactorScalar^2_{i, i} = 1$,
\begin{align}
\errorFunction(\weightVector_{:, i}) & = -\log\gaussianDist{\dataVector_{i, :}}{\sum_{j\in\neighborhood{i}}\dataVector_{j, :}\hat{\laplacianFactorScalar}_{j, i}}{\laplacianFactorScalar^{-2}_{i, i}} \nonumber \\
& = \frac{\laplacianFactorScalar_{i,
  i}^2}{2}\ltwoNorm{\dataVector_{i, :} -
  \sum_{j\in\neighborhood{i}}\dataVector_{j,
    :}\hat{\laplacianFactorScalar}_{j, i}}^2 - \frac{1}{2}\log
\laplacianFactorScalar_{i, i}^2 +\text{const} \nonumber \\
& = \frac{1}{2}\laplacianFactorVector_{:, i}^\top\dataMatrix\dataMatrix^\top\laplacianFactorVector_{:, i} - \frac{1}{2}\log
\laplacianFactorScalar_{i, i}^2 +\text{const}.\label{eq:lleModifiedSingleObjective}
\end{align}
The overall objective is the sum of the objectives for each column of
$\weightMatrix$. Under the probabilistic interpretation this is
equivalent to assuming independence between the individual
regressions. The objective can be written in matrix form as
\begin{equation}
\errorFunction(\weightMatrix) = \frac{1}{2}\sum_{i=1}^\numData
\laplacianFactorVector_{:, i}^\top\dataMatrix \dataMatrix^\top
\laplacianFactorVector_{:, i} -\frac{1}{2}\sum_{i=1}^\numData\log
\laplacianFactorScalar_{i, i}^2 +\text{const}\label{eq:lleObjective}.
\end{equation}

Recalling that our definition of $\laplacianFactor$ was in terms of
$\weightMatrix$, we now make that dependence explicit by
parameterizing the objective function only in terms of the non-zero
elements of $\weightMatrix$. To do this we introduce a `croupier matrix'
$\mathbf{S}_i\in \Re^{\numData \times k_i}$, where $k_i$ is the size
of the $i$ data point's neighborhood. This matrix will distribute the
non-zero elements of $\weightMatrix$ appropriately into
$\laplacianFactor$. It is defined in such a way that for the $i$th
data point we have $\laplacianFactorVector_{:, i} = \mathbf{S}_i
\weightVector_{i}$, where we use the shorthand $\weightVector_{i}=
\weightVector_{\neighborhood{i}, i}$. In other words
$\weightVector_{i}$ is the vector of regression weights being used to
reconstruct the $i$th data point. It contains the non-zero elements
from the $i$th column of $\weightMatrix$. The matrix $\mathbf{S}_i$ is
constructed by setting all elements in its $i$th row to $-1$ (causing
$\laplacianFactorScalar_{i,i}$ to be the negative sum of the elements
of $\weightVector_{i}$ as defined). Then we set $s_{\ell,j}$ to 1 if
$\ell$ is the $j$th neighbor of the data point $i$ and zero
otherwise. We can then the rewrite the objective function for the data
set as
\begin{equation}
\errorFunction(\weightMatrix) = \frac{1}{2}\sum_{i=1}^\numData
\weightVector_{i}^\top\mathbf{S}_i^\top\dataMatrix \dataMatrix^\top
\mathbf{S}_i\weightVector_{i} - \frac{1}{2}\sum_{i=1}^\numData \log
\weightVector_{i}^\top\onesVector\onesVector^\top \weightVector_{i}
+\text{const},
\end{equation}
A fixed point can be found by taking gradients with respect to $\weightVector_{i}$,
\[
\diff{\errorFunction(\weightMatrix)}{\weightVector_{i}} =
\mathbf{S}_i^\top\dataMatrix\dataMatrix^\top\mathbf{S}_i
\weightVector_{i} - \frac{1}{\weightVector_{i}^\top\onesVector}\onesVector
\]
which implies that the direction of $\weightVector_{i}$ is given by 
\[
\weightVector_{i}\propto \covarianceMatrix_i^{-1}\onesVector
\]
where $\covarianceMatrix_i=\mathbf{S}_i^\top\dataMatrix \dataMatrix^\top\mathbf{S}_i$ has been called the ``local covariance matrix''  by \cite{Roweis:lle00}, removing the croupier matrix we can express the local covariance matrix in the same form given by \cite{Roweis:lle00},
\[
\covarianceMatrix_i =  \sum_{j\in\neighborhood{i}} (\dataVector_{j, :} - \dataVector_{i, :})(\dataVector_{j, :} - \dataVector_{i, :})^\top.
\]
For standard LLE the magnitude of the vector $\weightVector_{i}$ is
set by the fact that $\onesVector^\top\weightVector_{i} = 1$. In our
alternative formulation we can find the magnitude of the vector
through differentiation of \refeq{eq:lleModifiedSingleObjective} with
respect to $\laplacianFactorScalar_{i, i}^2$ leading to the following
fixed point
\[
\laplacianFactorScalar_{i,i}^{-2} = \ltwoNorm{\dataVector_{i, :} -
  \sum_{j\in\neighborhood{i}}\dataVector_{j,
    :}\hat{\laplacianFactorScalar}_{j, i}}^{2}, 
\]
where $\hat{\laplacianFactorScalar}_{j, i}=
-\laplacianFactorScalar_{j,i}/\laplacianFactorScalar_{i, i}$. This
update shows explicitly that $\laplacianFactorScalar_{i, i}$ estimates
the precision with which each individual regression problem is solved.

\subsubsection{Determining the Embedding in LLE}

If the data is truly low dimensional, then we might expect that the
local linear relationships between neighbors continue to hold for a
data set $\latentMatrix$, of lower dimensionality,
$\latentDim<\dataDim$, than the original data $\dataMatrix$. The next
step in the LLE procedure is to find this data set. We do this by
minimizing the objective function in \refeq{eq:lleObjective} with
respect to this new, low dimensional data set. Writing the objective
in terms of this reduced dimensional data set, $\latentMatrix$, we
have
\begin{align*}
\errorFunction(\latentMatrix) &= \frac{1}{2}\sum_{i=1}^\numData
\laplacianFactorVector_{:, i}^\top\latentMatrix \latentMatrix^\top
\laplacianFactorVector_{:, i} +\text{const} \\ 
&=
\frac{1}{2}\trace{\laplacianFactor\laplacianFactor^\top
  \latentMatrix\latentMatrix} + \text{const} \\
& = \frac{1}{2}\sum_{i=1}^\numData \latentVector_{i,
  :}^\top\laplacianFactor\laplacianFactor^\top\latentVector_{i, :} +
\text{const}.
\end{align*}
Clearly the objective function is trivially minimized by setting
$\latentMatrix = \zerosVector$, so to avoid this solution a constraint
is imposed that $\latentMatrix^\top \latentMatrix = \eye$. This leads
to an eigenvalue problem of the form
\[
\laplacianFactor\laplacianFactor^\top \eigenvector_{i} = \eigenvalue_i\eigenvector_{i}.
\]
Here the smallest $\latentDim+1$ eigenvalues are extracted. The
smallest eigenvector is the constant eigenvector and is associated
with an eigenvalue of zero. This is because, by construction, we have
set $\laplacianFactor\laplacianFactor^\top
\onesVector=\zerosVector$. The next $\latentDim$ eigenvectors are
retained to make up the low dimensional representation so we have
\[
\latentVector_{:, i} = \eigenvector_{i+1} \quad \text{for} \quad i = 1 .. \latentDim.
\]

Extracting the latent coordinates in LLE is extremely similar to the
process suggested in Laplacian eigenmaps, despite different
motivations. Though in the LLE case the constraint on the latent
embeddings is not scaled by the degree matrix. The procedure is also
identical to that used in classical multidimensional scaling, and
therefore matches that used in MVU and MEU, although again the
motivation is different. Rather than distance matching, as suggested
for CMDS, in LLE we are looking for a `representative', low dimensional,
data set.

\subsubsection{Relating LLE to MEU}

We can see the similarity now between LLE and the Laplacian
eigenmaps. If we interpret $\laplacianFactor\laplacianFactor^\top$ as
a Laplacian we notice that the eigenvalue problem being solved for LLE
to recover the embedding is similar to that being solved in Laplacian
eigenmaps. The key difference between LLE and Laplacian eigenmaps is
the manner in which the Laplacian is parameterized. 

When introducing MEU we discussed how it is necessary to constrain the
Laplacian matrix to be positive definite (see
\refsec{sec:positivedefinite}). One way of doing this is to assume the
Laplacian factorizes as
\[
\laplacianMatrix =
\laplacianFactor\laplacianFactor^\top
\]
where $\laplacianFactor$ is non-symmetric.  If $\laplacianFactor$ is
constrained so that $\laplacianFactor^\top\onesVector = \zerosVector$
then we will also have $\laplacianMatrix\onesVector=\zerosVector$. As
we saw in the last section this constraint is easily achieved by
setting the diagonal elements $\laplacianFactorScalar_{i,i} =
-\sum_{j\in\neighborhood{i}} \laplacianFactorScalar_{j, i}$. Then if
we force $\laplacianFactorScalar_{j,i}=0$ if $j\notin\neighborhood{i}$
we will have a Laplacian matrix which is positive semidefinite without
need for any further constraint on $\laplacianFactor$. The sparsity
pattern of $\laplacianMatrix$ will, however, be different from the
pattern of $\laplacianFactor$. The entry for $\laplacianScalar_{i,j}$
will only be zero if there are no shared neighbors between $i$ and
$j$.

We described above
how the parameters of LLE, $\weightMatrix$, are chosen to reflect
locally linear relationships between neighboring data points. Here we
show that this algorithm is actually approximate maximum likelihood in
the MEU model. Indeed LLE turns out to be the specific case of maximum
entropy unfolding where:
\begin{enumerate}
  \item The diagonal sums, $\laplacianFactorScalar_{i,i}$, are further constrained to unity.
  \item The parameters of the model are optimized by maximizing the \emph{pseudolikelihood} of the resulting GRF.
\end{enumerate}
As we described in our introduction to LLE, traditionally the
reconstruction weights, $\weightVector_{i}$, are constrained to sum to
1. If this is the case then by our definition of $\laplacianFactor$ we
can write $\laplacianFactor = \eye - \weightMatrix$. The sparsity pattern of
$\weightMatrix$ matches $\laplacianFactor$, apart from the diagonal of
$\weightMatrix$ which is set to zero. These constraints mean that
$(\eye - \weightMatrix)^\top \onesVector = \zerosVector$. The LLE
algorithm \citep{Roweis:lle00} proscribes that the smallest
eigenvectors of $(\eye - \weightMatrix)(\eye - \weightMatrix)^\top =
\laplacianFactor\laplacianFactor^\top = \laplacianMatrix$ are used
with the constant eigenvector associated with the eigenvalue of 0
being discarded. This matches the CMDS procedure as applied to the MEU model, where the eigenvectors of $\laplacianMatrix$ are computed with the smallest eigenvector discarded through the centering operation.

\subsubsection{Pseudolikelihood Approximation}

To see how pseudolikelihood in the MEU model results in the LLE
procedure we firstly review the pseudolikelihood approximation
\citep[][]{Besag:pseudolikelihood75}. 

The Hammersley-Clifford theorem \citep{Hammersley:markov71} states that
for a Markov random field (of which our Gaussian random field is one
example) the joint probability density can be represented as a
factorization over the cliques of the graph. In the Gaussian random
field underlying maximum entropy unfolding the cliques are defined by
the neighbors of each data point and the relevant factorization is
\begin{equation}
p(\dataMatrix)\propto \prod_{i=1}^\numData
p(\dataVector_{i,:}|\dataMatrix_{\backslash i}), \label{eq:grfFactorization}
\end{equation}
where $\dataMatrix_{\backslash i}$ represents all data other than
the $i$th point and in practice each conditional distribution is
typically only dependent on a sub-set of $\dataMatrix_{\backslash i}$
(as defined by the neighborhood). As we will see, these conditional
distributions are straightforward to write out for maximum entropy
unfolding, particularly in the case where we have assumed the
factorization of the Laplacian,
$\laplacianMatrix=\laplacianFactor\laplacianFactor^\top$. 

The pseudolikelihood assumes that the proportionality in
\refeq{eq:grfFactorization} can be ignored and that the approximation
\[
p(\dataMatrix)\approx \prod_{i=1}^\numData
p(\dataVector_{i,:}|\dataMatrix_{\backslash i})
\]
is valid.

To see how the decomposition into cliques applies in the factorizable
MEU model first recall that
\[
\tr{\dataMatrix\dataMatrix^\top
  \laplacianFactor\laplacianFactor^\top}=\sum_{i=1}^\numData
\laplacianFactorVector_{:, i}^\top \dataMatrix\dataMatrix^\top
\laplacianFactorVector_{:, i}
\] 
so for the MEU model we have\footnote{Here we have ignored the term arising from the base density, $\trace{\gamma\dataMatrix\dataMatrix^\top}$. It also factorizes, but it doesn't affect the dependence of the pseudolikelihood on $\weightMatrix$.} 
\[
p(\dataMatrix)\propto \exp\left(-\frac{1}{2}\tr{\dataMatrix\dataMatrix^\top
  \laplacianFactor\laplacianFactor^\top}\right) = \prod_{i=1}^\numData
\exp\left(-\frac{1}{2} \laplacianFactorVector_{:, i}^\top \dataMatrix
\dataMatrix^\top \laplacianFactorVector_{:, i}\right).
\] 
This provides the necessary factorization for each conditional density
which can now be rewritten as
\[
p(\dataVector_{i, :}|\dataMatrix_{\backslash i}) = \left(\frac{\laplacianFactorScalar_{i,i}^2}{\twoPi }\right)^\frac{\dataDim}{2}\exp\left(-\frac{\laplacianFactorScalar_{i,i}^2}{2}\ltwoNorm{\dataVector_{i, :} + \sum_{j\in\neighborhood{i}}\frac{\laplacianFactorScalar_{j, i}}{\laplacianFactorScalar_{i,i}}\dataVector_{j, :} }^2\right).
\]
Optimizing the pseudolikelihood is equivalent to optimizing the
conditional density for each of these cliques independently,
\[
\log
p(\dataMatrix) \approx \sum_{i=1}^\numData \log p(\dataVector_{i,
  :}|\dataMatrix_{\backslash i}),
\] 
which is equivalent to solving $\numData$ independent regression
problems with a constraint on the regression weights that they sum to
one. This is exactly the optimization suggested in
\refeq{eq:lleObjective}. In maximum entropy unfolding the constraint
arises because the regression weights are constrained to be
$\weightScalar_{j,i}/\laplacianFactorScalar_{i,i}$ and
$\laplacianFactorScalar_{i,i} = \sum_{j\in \neighborhood{i}}
\weightScalar_{j, i}$. In standard LLE a further constraint is placed
that $\laplacianFactorScalar_{i,i}=1$ which implies none of these
regression problems should be solved to a greater precision than
another. However, as we derived above, LLE is also applicable even if
this further constraint isn't imposed.

Locally linear embeddings make use of the pseudolikelihood
approximation to parameter determination Gaussian random
field. Underpinning this is a neat way of constraining the Laplacian
to be positive semidefinite by assuming a factorized form. The
pseudolikelihood also allows for relatively quick parameter estimation
by ignoring the partition function from the actual likelihood. This
again removes the need to invert to recover the covariance matrix and
means that LLE can be applied to larger data sets than MEU or
MVU. However, the sparsity pattern in the Laplacian for LLE will not
match that used in the Laplacian for the other algorithms due to the
factorized representation.

\subsubsection{When is the Pseudolikelihood Valid in LLE?}

The pseudolikelihood was motivated by \cite{Besag:pseudolikelihood75}
for computational reasons. However, it obtains speed ups whilst
sacrificing accuracy: it doesn't make use of the correct form of the
normalization of the Gaussian random field. For a Gaussian model the
normalization is the determinant of the covariance matrix,
\[
\det{\kernelMatrix} = \det{\laplacianMatrix + \gamma
\eye}^{-1}.
\]
However, under particular circumstances the approximation is
exact. Here we quickly review an occasion when this occurs.

Imagine if we force $\laplacianFactor$ to be lower triangular, i.e. we
have a Cholesky form for our factorization of $\laplacianMatrix =
\laplacianFactor\laplacianFactor^\top$. The interpretation here is now
that $\laplacianFactor$ is a weighted adjacency matrix from a
\emph{directed acyclic graph}. When constructing the LLE neighborhood
the triangular form for this matrix can be achieved by first
imposing an ordering on the data points. Then, when seeking the
nearest $\numNeighbors$ neighbors for $i$, we only consider a candidate
data point $j$ if $j>i$. In the resulting directed acyclic graph the
neighbors of each data point are its \emph{parents}\footnote{Note that
  parents having a \emph{lower} index than children is the reverse of
  the standard convention. However, here it is necessary to maintain
  the structure of the Cholesky decomposition.}. The weighting of the
edge between node $j$ and its parent, $i$, is given by the $(i, j)$th
element of $\laplacianFactor$. To enforce the constraint that
$\laplacianFactor^\top\onesVector = \zerosVector$ the diagonal
elements of $\laplacianFactor$ are given by the negative sum of the
off diagonal elements from each column (i.e. the sum of their
parents). Note that the last data point (index $\numData$) has no
parents and so the $(\numData, \numData)$th element of
$\laplacianFactor$ is zero.
Now we use the fact that the log determinant of $\laplacianMatrix$ is
given by $\log \det{\laplacianFactor\laplacianFactor^\top} =
\sum_{i}\log \laplacianFactorScalar_{i, i}^2$ if $\laplacianFactor$ is lower
triangular. This means that for the particular structure we have imposed on the covariance the true log likelihood \emph{does}
factorize into $\numData$ independent regression problems,
\begin{align*}
  \log p(\dataMatrix) &=
  \frac{\det{\laplacianFactor\laplacianFactor^\top}^{\frac{1}{2}}}{\twoPi^{\frac{\numData}{2}}}
  \exp\left(-\frac{1}{2}\trace{\dataMatrix\dataMatrix
      \laplacianFactor\laplacianFactor^\top}\right) \\
  & = \prod_i^{\numData}
  \frac{\laplacianFactorScalar_{i,i}^2}{\twoPi^{\frac{1}{2}}}
  \exp\left(-\frac{\laplacianFactorScalar_{i,i}^2}{2}
    \ltwoNorm{\dataVector_{i, :} +
      \sum_{j\in\neighborhood{i}}\frac{\laplacianFactorScalar_{j,
          i}}{\laplacianFactorScalar_{i, i}}\dataVector_{j,
        :}}^2\right).
\end{align*}
The representation corresponds to a Gaussian random field which is
constructed from specifying the directed relationship between the
nodes in the graph. We can derive the Gaussian random field by
considering a series of conditional relationships,
\[
p(\dataVector_{i, :}| \dataMatrix_{\backslash i}) = p(\dataVector_{i, :} | \dataMatrix_{j>i, :})
\]
where our notation here is designed to indicate that the model is
constrained so that the density associated with each data point,
$\dataVector_{i, :}$, is only dependent on data points with an index
greater than $i$, a matrix we denote with $\dataMatrix_{j>i, :}$. This
constraint is enforced by our demand that the only potential neighbors
(parents in the directed graph) are those data points with an index
greater than $i$. The undirected system can now be produced by
taking the conditional densities of each data point,
\[
p(\dataVector_{i, :} | \dataMatrix_{j>i, :}) =
\gaussianDist{\dataVector_{i, :} }{\dataMatrix_{j>i, :}^\top \laplacianFactorVector_{j>i, i}}{\laplacianFactorScalar_{i, i}^{-2}\eye},
\]
and multiplying them together,
\[
p(\dataMatrix) = \prod_{i=1}^\numData p(\dataVector_{i, :} | \dataMatrix_{j>i, :}),
\]
to form the joint density. Note that the $\numData$th data point has
no parents so we can write $p(\dataVector_{\numData, :}|
\dataMatrix_{j>\numData, :}) = p(\dataVector_{\numData, :})$. However,
since we defined $\laplacianFactorScalar_{j, j} =
-\sum_{i>j}\laplacianFactorScalar_{i, j}$ the model as it currently
stands associates an infinite variance with this marginal density
($\laplacianFactorScalar_{\numData, \numData} =0$). This is a
consequence of the constraint $\laplacianFactor^\top \onesVector =
\zerosVector$. The problem manifests itself when computing the log
determinant of $\laplacianMatrix$, $\log
\det{\laplacianFactor\laplacianFactor^\top} = \sum_{i}\log
\laplacianFactorScalar_{i, i}^2$ to develop the log likelihood. The last
term in this sum is now $\log \laplacianFactorScalar_{\numData, \numData}^2
= \log 0$. As for the standard model this is resolved if we include
the $\gamma\eye$ term from the base density when computing the
determinant, but this destroys the separability of the determinant
computation. If the likelihood is required the value
$\laplacianFactorScalar_{\numData, \numData}$ could be set to a small
value, or optimized, relaxing the constraint on $\laplacianFactor$.

We call the algorithm based on the above decomposition acyclic locally
linear embedding (ALLE). A weakness for the ALLE is the need to
specify an ordering for the data. The ordering specifies which points
can be neighbors and different orderings will lead to different
results. Ideally one might want to specify the sparsity pattern in
$\laplacianMatrix$ and derive the appropriate sparsity structure for
$\laplacianFactor$. However, given a general undirected graph it is
not possible, in general, to find an equivalent directed acyclic
graph. This is because co-parents in the directed graph gain an edge
in the undirected graph, but the weight associated with this edge
cannot be set independently of the weights associated with the edges
between those co-parents and their children.
\subsubsection{LLE and PCA}

LLE is motivated by considering local linear embeddings of the data,
although interestingly, as we increase the neighborhood size to
$\numNeighbors = \numData-1$ we do not recover PCA, which is known to
be the optimal linear embedding of the data under linear Gaussian
constraints. The fact that LLE is optimizing the pseudolikelihood
makes it clear why this is the case. In contrast the MEU algorithm,
which LLE approximates, does recover PCA when $\numNeighbors =
\numData-1$. The ALLE algorithm also recovers PCA.

\subsection{Relation to Isomap}

The isomap algorithm \citep{Tenenbaum:isomap00} more directly follows
the CMDS framework. In isomap \citep{Tenenbaum:isomap00} a sparse
graph of distances is created between all points considered to be
neighbors. This graph is then filled in for all non-neighboring points
by finding the shortest distance between any two neighboring points in
the graph (along the edges specified by the neighbors). The resulting
matrix is then element-wise squared to give a matrix of square
distances which is then processed in the usual manner (centering and
multiplying by -0.5) to provide a similarity matrix for
multidimensional scaling. Compare this to the situation for MVU and
MEU. Both MVU and MEU can be thought of as starting with a sparse
graph of (squared) distances. The other distances are then filled in
by either maximizing the trace of the associated covariance or
maximizing the entropy. Importantly, though, the interneighbor
distances in this graph are preserved (through constraints imposed by
Lagrange multipliers) just like in isomap. For both MVU and MEU the
covariance matrix, $\kernelMatrix$, is guaranteed positive
semidefinite because the distances are implied by an underlying
covariance matrix that is constrained positive definite. For isomap
the shortest path algorithm is effectively approximating the distances
between non-neighboring points. This can lead to an implied covariance
matrix which has negative eigenvalues
\citep[see][]{Weinberger:learning04}.  The algorithm is still slower
than LLE and Laplacian eigenmaps because it requires a dense
eigenvalue problem and the application of a shortest path algorithm to
the graph provided by the neighbors.
\section{Estimating Graph Structure}\label{sec:drill}

The relationship between spectral dimensionality reduction algorithms
and Gaussian random fields now leads us to consider a novel approach
to dimensionality reduction. Recently it's been shown that the
structure of a Gaussian random field can be estimated through using L1
shrinkage on the parameters of the inverse covariance
\citep[see][Chapter 17]{Hastie:elements09}. These sparse graph
estimators are attractive as the regularization allows some structure
determination. In other words, rather than relying entirely on the
structure provided by the $\numNeighbors$ nearest neighbors in data
space, we can estimate this structure from the data.  We call the resulting class of approaches Dimensionality reduction
through Regularization of the Inverse covariance in the Log Likelihood
(DRILL). 

Before introducing the method, we need to first re-derive the maximum
entropy approach by constraining the second moment of neighboring data
points to equal the empirical observation instead of the expected inter data
point squared distances. We first define the empirically observed
second moment observation to be
\[
\sampleCovMatrix = \dataMatrix\dataMatrix^\top
\]
so if two points, $i$ and $j$ are neighbors then we constrain 
\[
\sampleCovScalar_{i, j} = \expectation{\dataVector_{i,:}^\top\dataVector_{j, :}},
\] 
where $\sampleCovScalar_{i, j}$ is the $i, j$th element of $\sampleCovMatrix$.  If we then further constrain the diagonal moments,
\begin{equation}
\expectation{\dataVector_{i, :}^\top\dataVector_{i, :}}=\sampleCovScalar_{i, i}\label{eq:diagonalConstraint}
\end{equation}
then the  expected squared
distance between two data points, will be given by
\[
\expectation{\distanceScalar_{i,j}} = \expectation{\dataVector_{i, :}^\top\dataVector_{i, :}}-2\expectation{\dataVector_{i, :}^\top\dataVector_{j, :}}+\expectation{\dataVector_{j, :}^\top\dataVector_{j, :}} = \sampleCovScalar_{i, i}-2\sampleCovScalar_{i, j} + \sampleCovScalar_{j, j}.
\]
So the expected interpoint squared distance will match the empirically
observed interpoint squared distance from the data. In other words,
whilst we've formulated the constraints slightly differently, the
final model will respect the same interpoint squared distance
constraints as our original formulation of maximum entropy unfolding.

The maximum entropy solution for the distribution has the form
\[
p(\dataMatrix) \propto \exp\left(-\frac{1}{2}\tr{\dataMatrix\dataMatrix^\top (\lagrangeMultiplierMatrix + \gamma\eye)}\right),
\]
where now the matrix of Lagrange multipliers matches the sparsity structure of the underlying neighborhood graph but also contains diagonal elements to enforce the constraint from \refeq{eq:diagonalConstraint}. Writing the full log likelihood in terms of the matrix $\sampleCovMatrix$ we have
\[
\log p(\dataMatrix) = -\frac{\dataDim\numData}{2}\log \twoPi + \frac{\dataDim}{2}\log \det{\lagrangeMultiplierMatrix+\gamma\eye} -\frac{1}{2}\tr{\sampleCovMatrix (\lagrangeMultiplierMatrix + \gamma\eye)},
\]
Once again, maximum likelihood in this system is equivalent to finding
the Lagrange multipliers so, given the structure from the neighborhood
relationships, we simply need to maximize the likelihood to solve the
system. That will lead to an implied covariance matrix,
\[
\kernelMatrix = (\lagrangeMultiplierMatrix + \gamma\eye)^{-1},
\]
which once again should be centred, $\centeredKernelMatrix =
\centeringMatrix \kernelMatrix \centeringMatrix$, and the principal
eigenvectors extracted to visualize the embedding. Here, though, we
are proposing some additional structure learning. If elements of the
inverse covariance are regularized appropriately the model can perform
some additional structure learning. In particular recent work on
application of L1 priors on the elements of the inverse covariance
\citep[see e.g.][]{Banerjee:sparse07,Friedman:sparse08} allows us to
apply a L1 regularizer to the inverse covariance and learn the
elements of $\lagrangeMultiplierMatrix$ efficiently. The objective
function for this system is now
\[
\errorFunction(\lagrangeMultiplierMatrix) = -\log p(\dataMatrix) + \sum_{i<j}\loneNorm{\lagrangeMultiplier_{i,j}}
\]
There has been a great deal of recent work on maximizing objectives of
this form. In our experiments we used the graphical lasso
algorithm \citep{Friedman:sparse08} which converts the optimization
into a series of iteratively applied lasso regressions.

\section{Experiments}
The models we have introduced are illustrative and draw on the
connections between existing methods. The advantages of our approaches
are in the unifying perspective they give and their potential to
exploit the characteristics of the probabilistic formulation to
explore extensions based on missing data, Bayesian formulations
etc.. However, for illustrative purposes we conclude with a short
experimental section.

For our experiments we consider two real world data sets. Code to
recreate all our experiments is available online. We applied each of
the spectral methods we have reviewed along with MEU using positive
constraints on the Lagrange multipliers (denoted MEU) and the DRILL
described in \refsec{sec:drill}. To evaluate the quality of our
embeddings we follow the suggestion of Harmeling
\citep{Harmeling:exploring07} and use the GP-LVM likelihood
\citep{Lawrence:pnpca05}. The higher the likelihood the better the
embedding. Harmeling conducted exhaustive tests over different
manifold types (with known ground truth) and found the GP-LVM
likelihood was the best indicator of the manifold quality amoungst all
the measures he tried. Our first data set consists of human motion
capture data.

\subsection{Motion Capture Data}

The data consists of a 3-dimensional point cloud of the location of 34
points from a subject performing a run. This leads to a 102
dimensional data set containing 55 frames of motion capture. The
subject begins the motion from stationary and takes approximately
three strides of run. We hope to see this structure in the
visualization: a starting position followed by a series of loops.  The
data was made available by Ohio State University. The data is
characterized by a cyclic pattern during the strides of run. However,
the angle of inclination during the run changes so there are slight
differences for each cycle. The data is very low noise, as the motion
capture rig is designed to extract the point locations of the subject
to a high precision.

\begin{figure}
\begingroup
  \makeatletter
  \providecommand\color[2][]{%
    \GenericError{(gnuplot) \space\space\space\@spaces}{%
      Package color not loaded in conjunction with
      terminal option `colourtext'%
    }{See the gnuplot documentation for explanation.%
    }{Either use 'blacktext' in gnuplot or load the package
      color.sty in LaTeX.}%
    \renewcommand\color[2][]{}%
  }%
  \providecommand\includegraphics[2][]{%
    \GenericError{(gnuplot) \space\space\space\@spaces}{%
      Package graphicx or graphics not loaded%
    }{See the gnuplot documentation for explanation.%
    }{The gnuplot epslatex terminal needs graphicx.sty or graphics.sty.}%
    \renewcommand\includegraphics[2][]{}%
  }%
  \providecommand\rotatebox[2]{#2}%
  \@ifundefined{ifGPcolor}{%
    \newif\ifGPcolor
    \GPcolortrue
  }{}%
  \@ifundefined{ifGPblacktext}{%
    \newif\ifGPblacktext
    \GPblacktexttrue
  }{}%
  \let\gplgaddtomacro\g@addto@macro
  \gdef\gplbacktext{}%
  \gdef\gplfronttext{}%
  \makeatother
  \ifGPblacktext
    \def\colorrgb#1{}%
    \def\colorgray#1{}%
  \else
    \ifGPcolor
      \def\colorrgb#1{\color[rgb]{#1}}%
      \def\colorgray#1{\color[gray]{#1}}%
      \expandafter\def\csname LTw\endcsname{\color{white}}%
      \expandafter\def\csname LTb\endcsname{\color{black}}%
      \expandafter\def\csname LTa\endcsname{\color{black}}%
      \expandafter\def\csname LT0\endcsname{\color[rgb]{1,0,0}}%
      \expandafter\def\csname LT1\endcsname{\color[rgb]{0,1,0}}%
      \expandafter\def\csname LT2\endcsname{\color[rgb]{0,0,1}}%
      \expandafter\def\csname LT3\endcsname{\color[rgb]{1,0,1}}%
      \expandafter\def\csname LT4\endcsname{\color[rgb]{0,1,1}}%
      \expandafter\def\csname LT5\endcsname{\color[rgb]{1,1,0}}%
      \expandafter\def\csname LT6\endcsname{\color[rgb]{0,0,0}}%
      \expandafter\def\csname LT7\endcsname{\color[rgb]{1,0.3,0}}%
      \expandafter\def\csname LT8\endcsname{\color[rgb]{0.5,0.5,0.5}}%
    \else
      \def\colorrgb#1{\color{black}}%
      \def\colorgray#1{\color[gray]{#1}}%
      \expandafter\def\csname LTw\endcsname{\color{white}}%
      \expandafter\def\csname LTb\endcsname{\color{black}}%
      \expandafter\def\csname LTa\endcsname{\color{black}}%
      \expandafter\def\csname LT0\endcsname{\color{black}}%
      \expandafter\def\csname LT1\endcsname{\color{black}}%
      \expandafter\def\csname LT2\endcsname{\color{black}}%
      \expandafter\def\csname LT3\endcsname{\color{black}}%
      \expandafter\def\csname LT4\endcsname{\color{black}}%
      \expandafter\def\csname LT5\endcsname{\color{black}}%
      \expandafter\def\csname LT6\endcsname{\color{black}}%
      \expandafter\def\csname LT7\endcsname{\color{black}}%
      \expandafter\def\csname LT8\endcsname{\color{black}}%
    \fi
  \fi
  \setlength{\unitlength}{0.0500bp}%
  \begin{picture}(4336.00,3252.00)%
    \gplgaddtomacro\gplbacktext{%
      \colorrgb{0.00,0.00,0.00}%
      \put(431,358){\makebox(0,0)[r]{\strut{}\small -2}}%
      \colorrgb{0.00,0.00,0.00}%
      \put(431,888){\makebox(0,0)[r]{\strut{}\small -1}}%
      \colorrgb{0.00,0.00,0.00}%
      \put(431,1418){\makebox(0,0)[r]{\strut{}\small 0}}%
      \colorrgb{0.00,0.00,0.00}%
      \put(431,1947){\makebox(0,0)[r]{\strut{}\small 1}}%
      \colorrgb{0.00,0.00,0.00}%
      \put(431,2477){\makebox(0,0)[r]{\strut{}\small 2}}%
      \colorrgb{0.00,0.00,0.00}%
      \put(431,3007){\makebox(0,0)[r]{\strut{}\small 3}}%
      \colorrgb{0.00,0.00,0.00}%
      \put(1043,138){\makebox(0,0){\strut{}\small -1}}%
      \colorrgb{0.00,0.00,0.00}%
      \put(2003,138){\makebox(0,0){\strut{}\small 0}}%
      \colorrgb{0.00,0.00,0.00}%
      \put(2964,138){\makebox(0,0){\strut{}\small 1}}%
      \colorrgb{0.00,0.00,0.00}%
      \put(3924,138){\makebox(0,0){\strut{}\small 2}}%
    }%
    \gplgaddtomacro\gplfronttext{%
    }%
    \gplbacktext
    \put(0,0){\includegraphics{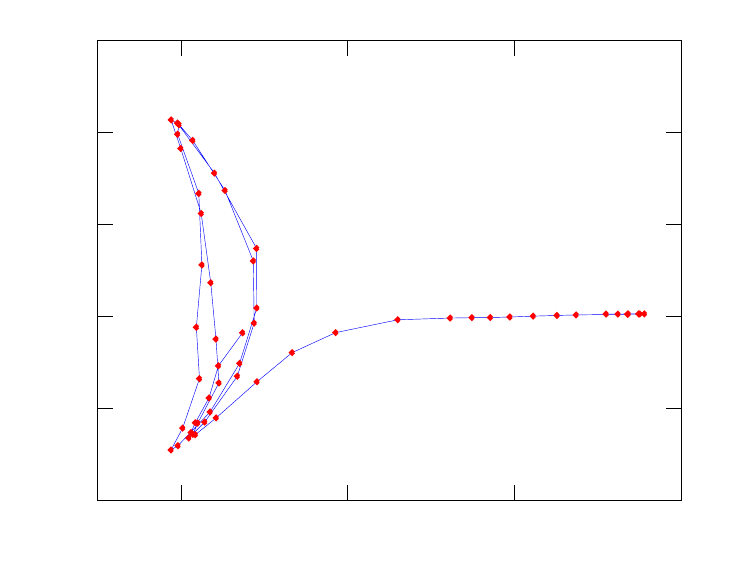}}%
    \gplfronttext
  \end{picture}%
\endgroup
  \hfill
\begingroup
  \makeatletter
  \providecommand\color[2][]{%
    \GenericError{(gnuplot) \space\space\space\@spaces}{%
      Package color not loaded in conjunction with
      terminal option `colourtext'%
    }{See the gnuplot documentation for explanation.%
    }{Either use 'blacktext' in gnuplot or load the package
      color.sty in LaTeX.}%
    \renewcommand\color[2][]{}%
  }%
  \providecommand\includegraphics[2][]{%
    \GenericError{(gnuplot) \space\space\space\@spaces}{%
      Package graphicx or graphics not loaded%
    }{See the gnuplot documentation for explanation.%
    }{The gnuplot epslatex terminal needs graphicx.sty or graphics.sty.}%
    \renewcommand\includegraphics[2][]{}%
  }%
  \providecommand\rotatebox[2]{#2}%
  \@ifundefined{ifGPcolor}{%
    \newif\ifGPcolor
    \GPcolortrue
  }{}%
  \@ifundefined{ifGPblacktext}{%
    \newif\ifGPblacktext
    \GPblacktexttrue
  }{}%
  \let\gplgaddtomacro\g@addto@macro
  \gdef\gplbacktext{}%
  \gdef\gplfronttext{}%
  \makeatother
  \ifGPblacktext
    \def\colorrgb#1{}%
    \def\colorgray#1{}%
  \else
    \ifGPcolor
      \def\colorrgb#1{\color[rgb]{#1}}%
      \def\colorgray#1{\color[gray]{#1}}%
      \expandafter\def\csname LTw\endcsname{\color{white}}%
      \expandafter\def\csname LTb\endcsname{\color{black}}%
      \expandafter\def\csname LTa\endcsname{\color{black}}%
      \expandafter\def\csname LT0\endcsname{\color[rgb]{1,0,0}}%
      \expandafter\def\csname LT1\endcsname{\color[rgb]{0,1,0}}%
      \expandafter\def\csname LT2\endcsname{\color[rgb]{0,0,1}}%
      \expandafter\def\csname LT3\endcsname{\color[rgb]{1,0,1}}%
      \expandafter\def\csname LT4\endcsname{\color[rgb]{0,1,1}}%
      \expandafter\def\csname LT5\endcsname{\color[rgb]{1,1,0}}%
      \expandafter\def\csname LT6\endcsname{\color[rgb]{0,0,0}}%
      \expandafter\def\csname LT7\endcsname{\color[rgb]{1,0.3,0}}%
      \expandafter\def\csname LT8\endcsname{\color[rgb]{0.5,0.5,0.5}}%
    \else
      \def\colorrgb#1{\color{black}}%
      \def\colorgray#1{\color[gray]{#1}}%
      \expandafter\def\csname LTw\endcsname{\color{white}}%
      \expandafter\def\csname LTb\endcsname{\color{black}}%
      \expandafter\def\csname LTa\endcsname{\color{black}}%
      \expandafter\def\csname LT0\endcsname{\color{black}}%
      \expandafter\def\csname LT1\endcsname{\color{black}}%
      \expandafter\def\csname LT2\endcsname{\color{black}}%
      \expandafter\def\csname LT3\endcsname{\color{black}}%
      \expandafter\def\csname LT4\endcsname{\color{black}}%
      \expandafter\def\csname LT5\endcsname{\color{black}}%
      \expandafter\def\csname LT6\endcsname{\color{black}}%
      \expandafter\def\csname LT7\endcsname{\color{black}}%
      \expandafter\def\csname LT8\endcsname{\color{black}}%
    \fi
  \fi
  \setlength{\unitlength}{0.0500bp}%
  \begin{picture}(4336.00,3252.00)%
    \gplgaddtomacro\gplbacktext{%
      \colorrgb{0.00,0.00,0.00}%
      \put(431,736){\makebox(0,0)[r]{\strut{}\small -2}}%
      \colorrgb{0.00,0.00,0.00}%
      \put(431,1493){\makebox(0,0)[r]{\strut{}\small -1}}%
      \colorrgb{0.00,0.00,0.00}%
      \put(431,2250){\makebox(0,0)[r]{\strut{}\small 0}}%
      \colorrgb{0.00,0.00,0.00}%
      \put(431,3007){\makebox(0,0)[r]{\strut{}\small 1}}%
      \colorrgb{0.00,0.00,0.00}%
      \put(563,138){\makebox(0,0){\strut{}\small -2}}%
      \colorrgb{0.00,0.00,0.00}%
      \put(1235,138){\makebox(0,0){\strut{}\small -1}}%
      \colorrgb{0.00,0.00,0.00}%
      \put(1907,138){\makebox(0,0){\strut{}\small 0}}%
      \colorrgb{0.00,0.00,0.00}%
      \put(2580,138){\makebox(0,0){\strut{}\small 1}}%
      \colorrgb{0.00,0.00,0.00}%
      \put(3252,138){\makebox(0,0){\strut{}\small 2}}%
      \colorrgb{0.00,0.00,0.00}%
      \put(3924,138){\makebox(0,0){\strut{}\small 3}}%
    }%
    \gplgaddtomacro\gplfronttext{%
    }%
    \gplbacktext
    \put(0,0){\includegraphics{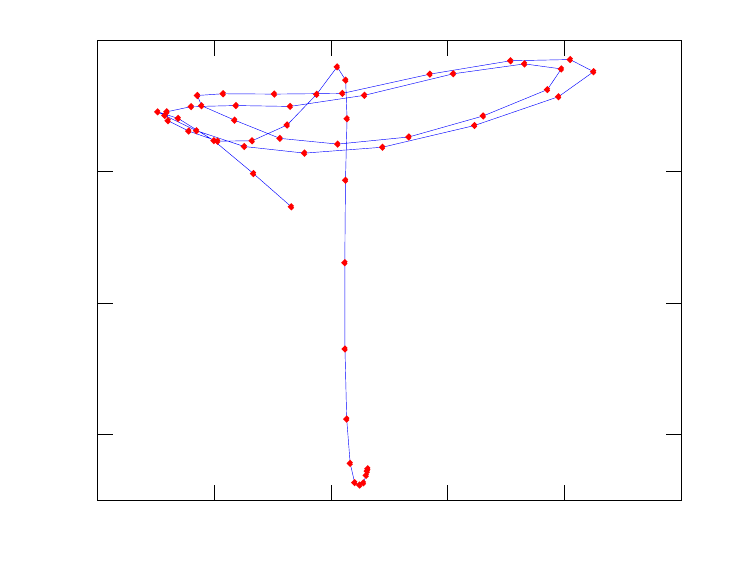}}%
    \gplfronttext
  \end{picture}%
\endgroup
  \hfill
\begingroup
  \makeatletter
  \providecommand\color[2][]{%
    \GenericError{(gnuplot) \space\space\space\@spaces}{%
      Package color not loaded in conjunction with
      terminal option `colourtext'%
    }{See the gnuplot documentation for explanation.%
    }{Either use 'blacktext' in gnuplot or load the package
      color.sty in LaTeX.}%
    \renewcommand\color[2][]{}%
  }%
  \providecommand\includegraphics[2][]{%
    \GenericError{(gnuplot) \space\space\space\@spaces}{%
      Package graphicx or graphics not loaded%
    }{See the gnuplot documentation for explanation.%
    }{The gnuplot epslatex terminal needs graphicx.sty or graphics.sty.}%
    \renewcommand\includegraphics[2][]{}%
  }%
  \providecommand\rotatebox[2]{#2}%
  \@ifundefined{ifGPcolor}{%
    \newif\ifGPcolor
    \GPcolortrue
  }{}%
  \@ifundefined{ifGPblacktext}{%
    \newif\ifGPblacktext
    \GPblacktexttrue
  }{}%
  \let\gplgaddtomacro\g@addto@macro
  \gdef\gplbacktext{}%
  \gdef\gplfronttext{}%
  \makeatother
  \ifGPblacktext
    \def\colorrgb#1{}%
    \def\colorgray#1{}%
  \else
    \ifGPcolor
      \def\colorrgb#1{\color[rgb]{#1}}%
      \def\colorgray#1{\color[gray]{#1}}%
      \expandafter\def\csname LTw\endcsname{\color{white}}%
      \expandafter\def\csname LTb\endcsname{\color{black}}%
      \expandafter\def\csname LTa\endcsname{\color{black}}%
      \expandafter\def\csname LT0\endcsname{\color[rgb]{1,0,0}}%
      \expandafter\def\csname LT1\endcsname{\color[rgb]{0,1,0}}%
      \expandafter\def\csname LT2\endcsname{\color[rgb]{0,0,1}}%
      \expandafter\def\csname LT3\endcsname{\color[rgb]{1,0,1}}%
      \expandafter\def\csname LT4\endcsname{\color[rgb]{0,1,1}}%
      \expandafter\def\csname LT5\endcsname{\color[rgb]{1,1,0}}%
      \expandafter\def\csname LT6\endcsname{\color[rgb]{0,0,0}}%
      \expandafter\def\csname LT7\endcsname{\color[rgb]{1,0.3,0}}%
      \expandafter\def\csname LT8\endcsname{\color[rgb]{0.5,0.5,0.5}}%
    \else
      \def\colorrgb#1{\color{black}}%
      \def\colorgray#1{\color[gray]{#1}}%
      \expandafter\def\csname LTw\endcsname{\color{white}}%
      \expandafter\def\csname LTb\endcsname{\color{black}}%
      \expandafter\def\csname LTa\endcsname{\color{black}}%
      \expandafter\def\csname LT0\endcsname{\color{black}}%
      \expandafter\def\csname LT1\endcsname{\color{black}}%
      \expandafter\def\csname LT2\endcsname{\color{black}}%
      \expandafter\def\csname LT3\endcsname{\color{black}}%
      \expandafter\def\csname LT4\endcsname{\color{black}}%
      \expandafter\def\csname LT5\endcsname{\color{black}}%
      \expandafter\def\csname LT6\endcsname{\color{black}}%
      \expandafter\def\csname LT7\endcsname{\color{black}}%
      \expandafter\def\csname LT8\endcsname{\color{black}}%
    \fi
  \fi
  \setlength{\unitlength}{0.0500bp}%
  \begin{picture}(4336.00,3252.00)%
    \gplgaddtomacro\gplbacktext{%
      \colorrgb{0.00,0.00,0.00}%
      \put(431,736){\makebox(0,0)[r]{\strut{}\small -1}}%
      \colorrgb{0.00,0.00,0.00}%
      \put(431,1493){\makebox(0,0)[r]{\strut{}\small 0}}%
      \colorrgb{0.00,0.00,0.00}%
      \put(431,2250){\makebox(0,0)[r]{\strut{}\small 1}}%
      \colorrgb{0.00,0.00,0.00}%
      \put(431,3007){\makebox(0,0)[r]{\strut{}\small 2}}%
      \colorrgb{0.00,0.00,0.00}%
      \put(563,138){\makebox(0,0){\strut{}\small -2}}%
      \colorrgb{0.00,0.00,0.00}%
      \put(1403,138){\makebox(0,0){\strut{}\small -1}}%
      \colorrgb{0.00,0.00,0.00}%
      \put(2244,138){\makebox(0,0){\strut{}\small 0}}%
      \colorrgb{0.00,0.00,0.00}%
      \put(3084,138){\makebox(0,0){\strut{}\small 1}}%
      \colorrgb{0.00,0.00,0.00}%
      \put(3924,138){\makebox(0,0){\strut{}\small 2}}%
    }%
    \gplgaddtomacro\gplfronttext{%
    }%
    \gplbacktext
    \put(0,0){\includegraphics{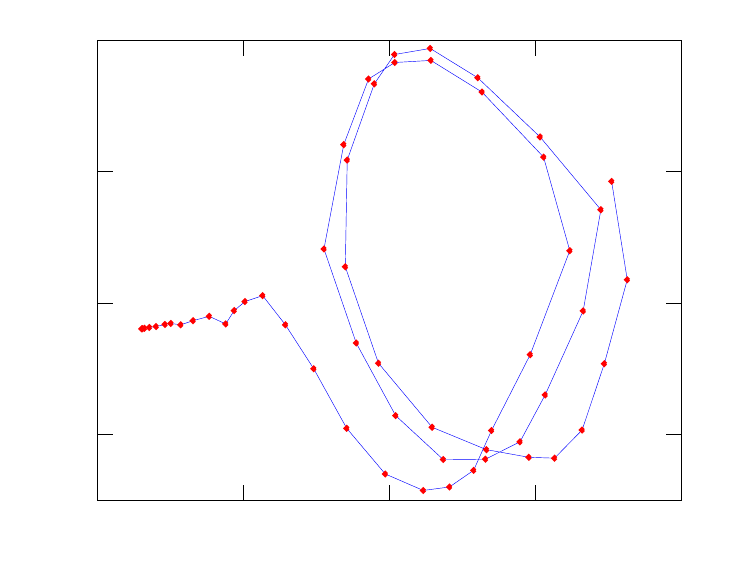}}%
    \gplfronttext
  \end{picture}%
\endgroup
  \hfill
\begingroup
  \makeatletter
  \providecommand\color[2][]{%
    \GenericError{(gnuplot) \space\space\space\@spaces}{%
      Package color not loaded in conjunction with
      terminal option `colourtext'%
    }{See the gnuplot documentation for explanation.%
    }{Either use 'blacktext' in gnuplot or load the package
      color.sty in LaTeX.}%
    \renewcommand\color[2][]{}%
  }%
  \providecommand\includegraphics[2][]{%
    \GenericError{(gnuplot) \space\space\space\@spaces}{%
      Package graphicx or graphics not loaded%
    }{See the gnuplot documentation for explanation.%
    }{The gnuplot epslatex terminal needs graphicx.sty or graphics.sty.}%
    \renewcommand\includegraphics[2][]{}%
  }%
  \providecommand\rotatebox[2]{#2}%
  \@ifundefined{ifGPcolor}{%
    \newif\ifGPcolor
    \GPcolortrue
  }{}%
  \@ifundefined{ifGPblacktext}{%
    \newif\ifGPblacktext
    \GPblacktexttrue
  }{}%
  \let\gplgaddtomacro\g@addto@macro
  \gdef\gplbacktext{}%
  \gdef\gplfronttext{}%
  \makeatother
  \ifGPblacktext
    \def\colorrgb#1{}%
    \def\colorgray#1{}%
  \else
    \ifGPcolor
      \def\colorrgb#1{\color[rgb]{#1}}%
      \def\colorgray#1{\color[gray]{#1}}%
      \expandafter\def\csname LTw\endcsname{\color{white}}%
      \expandafter\def\csname LTb\endcsname{\color{black}}%
      \expandafter\def\csname LTa\endcsname{\color{black}}%
      \expandafter\def\csname LT0\endcsname{\color[rgb]{1,0,0}}%
      \expandafter\def\csname LT1\endcsname{\color[rgb]{0,1,0}}%
      \expandafter\def\csname LT2\endcsname{\color[rgb]{0,0,1}}%
      \expandafter\def\csname LT3\endcsname{\color[rgb]{1,0,1}}%
      \expandafter\def\csname LT4\endcsname{\color[rgb]{0,1,1}}%
      \expandafter\def\csname LT5\endcsname{\color[rgb]{1,1,0}}%
      \expandafter\def\csname LT6\endcsname{\color[rgb]{0,0,0}}%
      \expandafter\def\csname LT7\endcsname{\color[rgb]{1,0.3,0}}%
      \expandafter\def\csname LT8\endcsname{\color[rgb]{0.5,0.5,0.5}}%
    \else
      \def\colorrgb#1{\color{black}}%
      \def\colorgray#1{\color[gray]{#1}}%
      \expandafter\def\csname LTw\endcsname{\color{white}}%
      \expandafter\def\csname LTb\endcsname{\color{black}}%
      \expandafter\def\csname LTa\endcsname{\color{black}}%
      \expandafter\def\csname LT0\endcsname{\color{black}}%
      \expandafter\def\csname LT1\endcsname{\color{black}}%
      \expandafter\def\csname LT2\endcsname{\color{black}}%
      \expandafter\def\csname LT3\endcsname{\color{black}}%
      \expandafter\def\csname LT4\endcsname{\color{black}}%
      \expandafter\def\csname LT5\endcsname{\color{black}}%
      \expandafter\def\csname LT6\endcsname{\color{black}}%
      \expandafter\def\csname LT7\endcsname{\color{black}}%
      \expandafter\def\csname LT8\endcsname{\color{black}}%
    \fi
  \fi
  \setlength{\unitlength}{0.0500bp}%
  \begin{picture}(4336.00,3252.00)%
    \gplgaddtomacro\gplbacktext{%
      \colorrgb{0.00,0.00,0.00}%
      \put(431,358){\makebox(0,0)[r]{\strut{}\small -2}}%
      \colorrgb{0.00,0.00,0.00}%
      \put(431,1020){\makebox(0,0)[r]{\strut{}\small -1}}%
      \colorrgb{0.00,0.00,0.00}%
      \put(431,1683){\makebox(0,0)[r]{\strut{}\small 0}}%
      \colorrgb{0.00,0.00,0.00}%
      \put(431,2345){\makebox(0,0)[r]{\strut{}\small 1}}%
      \colorrgb{0.00,0.00,0.00}%
      \put(431,3007){\makebox(0,0)[r]{\strut{}\small 2}}%
      \colorrgb{0.00,0.00,0.00}%
      \put(1571,138){\makebox(0,0){\strut{}\small -3}}%
      \colorrgb{0.00,0.00,0.00}%
      \put(1907,138){\makebox(0,0){\strut{}\small -2}}%
      \colorrgb{0.00,0.00,0.00}%
      \put(2244,138){\makebox(0,0){\strut{}\small -1}}%
      \colorrgb{0.00,0.00,0.00}%
      \put(2580,138){\makebox(0,0){\strut{}\small 0}}%
      \colorrgb{0.00,0.00,0.00}%
      \put(2916,138){\makebox(0,0){\strut{}\small 1}}%
      \colorrgb{0.00,0.00,0.00}%
      \put(3252,138){\makebox(0,0){\strut{}\small 2}}%
      \colorrgb{0.00,0.00,0.00}%
      \put(3588,138){\makebox(0,0){\strut{}\small 3}}%
    }%
    \gplgaddtomacro\gplfronttext{%
    }%
    \gplbacktext
    \put(0,0){\includegraphics{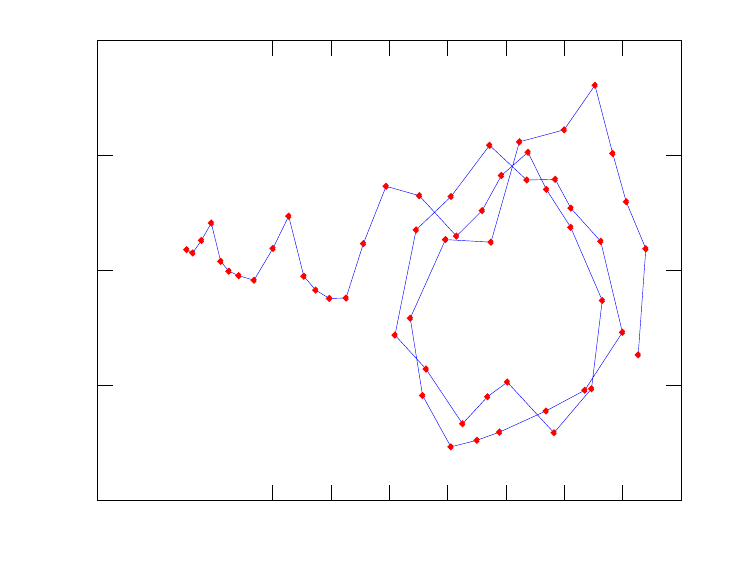}}%
    \gplfronttext
  \end{picture}%
\endgroup
  \caption{Motion capture data visualized in two dimensions for each
    algorithm we reviewed using 6 nearest neighbors. Models capture either the
    cyclic structure or the structure associated with the start of the
    run or both parts.}\label{fig:embedStick1}
\end{figure}

\begin{figure}
  \begin{center}
\begingroup
  \makeatletter
  \providecommand\color[2][]{%
    \GenericError{(gnuplot) \space\space\space\@spaces}{%
      Package color not loaded in conjunction with
      terminal option `colourtext'%
    }{See the gnuplot documentation for explanation.%
    }{Either use 'blacktext' in gnuplot or load the package
      color.sty in LaTeX.}%
    \renewcommand\color[2][]{}%
  }%
  \providecommand\includegraphics[2][]{%
    \GenericError{(gnuplot) \space\space\space\@spaces}{%
      Package graphicx or graphics not loaded%
    }{See the gnuplot documentation for explanation.%
    }{The gnuplot epslatex terminal needs graphicx.sty or graphics.sty.}%
    \renewcommand\includegraphics[2][]{}%
  }%
  \providecommand\rotatebox[2]{#2}%
  \@ifundefined{ifGPcolor}{%
    \newif\ifGPcolor
    \GPcolortrue
  }{}%
  \@ifundefined{ifGPblacktext}{%
    \newif\ifGPblacktext
    \GPblacktexttrue
  }{}%
  \let\gplgaddtomacro\g@addto@macro
  \gdef\gplbacktext{}%
  \gdef\gplfronttext{}%
  \makeatother
  \ifGPblacktext
    \def\colorrgb#1{}%
    \def\colorgray#1{}%
  \else
    \ifGPcolor
      \def\colorrgb#1{\color[rgb]{#1}}%
      \def\colorgray#1{\color[gray]{#1}}%
      \expandafter\def\csname LTw\endcsname{\color{white}}%
      \expandafter\def\csname LTb\endcsname{\color{black}}%
      \expandafter\def\csname LTa\endcsname{\color{black}}%
      \expandafter\def\csname LT0\endcsname{\color[rgb]{1,0,0}}%
      \expandafter\def\csname LT1\endcsname{\color[rgb]{0,1,0}}%
      \expandafter\def\csname LT2\endcsname{\color[rgb]{0,0,1}}%
      \expandafter\def\csname LT3\endcsname{\color[rgb]{1,0,1}}%
      \expandafter\def\csname LT4\endcsname{\color[rgb]{0,1,1}}%
      \expandafter\def\csname LT5\endcsname{\color[rgb]{1,1,0}}%
      \expandafter\def\csname LT6\endcsname{\color[rgb]{0,0,0}}%
      \expandafter\def\csname LT7\endcsname{\color[rgb]{1,0.3,0}}%
      \expandafter\def\csname LT8\endcsname{\color[rgb]{0.5,0.5,0.5}}%
    \else
      \def\colorrgb#1{\color{black}}%
      \def\colorgray#1{\color[gray]{#1}}%
      \expandafter\def\csname LTw\endcsname{\color{white}}%
      \expandafter\def\csname LTb\endcsname{\color{black}}%
      \expandafter\def\csname LTa\endcsname{\color{black}}%
      \expandafter\def\csname LT0\endcsname{\color{black}}%
      \expandafter\def\csname LT1\endcsname{\color{black}}%
      \expandafter\def\csname LT2\endcsname{\color{black}}%
      \expandafter\def\csname LT3\endcsname{\color{black}}%
      \expandafter\def\csname LT4\endcsname{\color{black}}%
      \expandafter\def\csname LT5\endcsname{\color{black}}%
      \expandafter\def\csname LT6\endcsname{\color{black}}%
      \expandafter\def\csname LT7\endcsname{\color{black}}%
      \expandafter\def\csname LT8\endcsname{\color{black}}%
    \fi
  \fi
  \setlength{\unitlength}{0.0500bp}%
  \begin{picture}(4336.00,3252.00)%
    \gplgaddtomacro\gplbacktext{%
      \colorrgb{0.00,0.00,0.00}%
      \put(431,736){\makebox(0,0)[r]{\strut{}\small -1}}%
      \colorrgb{0.00,0.00,0.00}%
      \put(431,1493){\makebox(0,0)[r]{\strut{}\small 0}}%
      \colorrgb{0.00,0.00,0.00}%
      \put(431,2250){\makebox(0,0)[r]{\strut{}\small 1}}%
      \colorrgb{0.00,0.00,0.00}%
      \put(431,3007){\makebox(0,0)[r]{\strut{}\small 2}}%
      \colorrgb{0.00,0.00,0.00}%
      \put(563,138){\makebox(0,0){\strut{}\small -2}}%
      \colorrgb{0.00,0.00,0.00}%
      \put(1683,138){\makebox(0,0){\strut{}\small -1}}%
      \colorrgb{0.00,0.00,0.00}%
      \put(2804,138){\makebox(0,0){\strut{}\small 0}}%
      \colorrgb{0.00,0.00,0.00}%
      \put(3924,138){\makebox(0,0){\strut{}\small 1}}%
    }%
    \gplgaddtomacro\gplfronttext{%
    }%
    \gplbacktext
    \put(0,0){\includegraphics{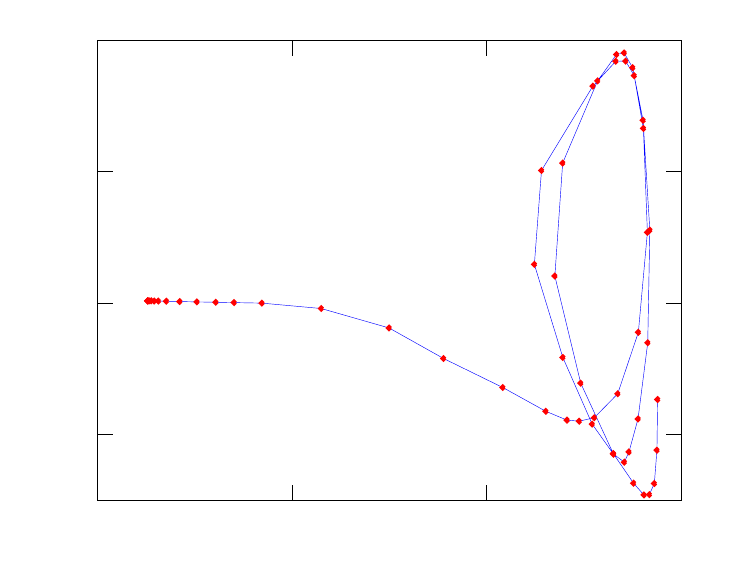}}%
    \gplfronttext
  \end{picture}%
\endgroup
\begingroup
  \makeatletter
  \providecommand\color[2][]{%
    \GenericError{(gnuplot) \space\space\space\@spaces}{%
      Package color not loaded in conjunction with
      terminal option `colourtext'%
    }{See the gnuplot documentation for explanation.%
    }{Either use 'blacktext' in gnuplot or load the package
      color.sty in LaTeX.}%
    \renewcommand\color[2][]{}%
  }%
  \providecommand\includegraphics[2][]{%
    \GenericError{(gnuplot) \space\space\space\@spaces}{%
      Package graphicx or graphics not loaded%
    }{See the gnuplot documentation for explanation.%
    }{The gnuplot epslatex terminal needs graphicx.sty or graphics.sty.}%
    \renewcommand\includegraphics[2][]{}%
  }%
  \providecommand\rotatebox[2]{#2}%
  \@ifundefined{ifGPcolor}{%
    \newif\ifGPcolor
    \GPcolortrue
  }{}%
  \@ifundefined{ifGPblacktext}{%
    \newif\ifGPblacktext
    \GPblacktexttrue
  }{}%
  \let\gplgaddtomacro\g@addto@macro
  \gdef\gplbacktext{}%
  \gdef\gplfronttext{}%
  \makeatother
  \ifGPblacktext
    \def\colorrgb#1{}%
    \def\colorgray#1{}%
  \else
    \ifGPcolor
      \def\colorrgb#1{\color[rgb]{#1}}%
      \def\colorgray#1{\color[gray]{#1}}%
      \expandafter\def\csname LTw\endcsname{\color{white}}%
      \expandafter\def\csname LTb\endcsname{\color{black}}%
      \expandafter\def\csname LTa\endcsname{\color{black}}%
      \expandafter\def\csname LT0\endcsname{\color[rgb]{1,0,0}}%
      \expandafter\def\csname LT1\endcsname{\color[rgb]{0,1,0}}%
      \expandafter\def\csname LT2\endcsname{\color[rgb]{0,0,1}}%
      \expandafter\def\csname LT3\endcsname{\color[rgb]{1,0,1}}%
      \expandafter\def\csname LT4\endcsname{\color[rgb]{0,1,1}}%
      \expandafter\def\csname LT5\endcsname{\color[rgb]{1,1,0}}%
      \expandafter\def\csname LT6\endcsname{\color[rgb]{0,0,0}}%
      \expandafter\def\csname LT7\endcsname{\color[rgb]{1,0.3,0}}%
      \expandafter\def\csname LT8\endcsname{\color[rgb]{0.5,0.5,0.5}}%
    \else
      \def\colorrgb#1{\color{black}}%
      \def\colorgray#1{\color[gray]{#1}}%
      \expandafter\def\csname LTw\endcsname{\color{white}}%
      \expandafter\def\csname LTb\endcsname{\color{black}}%
      \expandafter\def\csname LTa\endcsname{\color{black}}%
      \expandafter\def\csname LT0\endcsname{\color{black}}%
      \expandafter\def\csname LT1\endcsname{\color{black}}%
      \expandafter\def\csname LT2\endcsname{\color{black}}%
      \expandafter\def\csname LT3\endcsname{\color{black}}%
      \expandafter\def\csname LT4\endcsname{\color{black}}%
      \expandafter\def\csname LT5\endcsname{\color{black}}%
      \expandafter\def\csname LT6\endcsname{\color{black}}%
      \expandafter\def\csname LT7\endcsname{\color{black}}%
      \expandafter\def\csname LT8\endcsname{\color{black}}%
    \fi
  \fi
  \setlength{\unitlength}{0.0500bp}%
  \begin{picture}(4336.00,3252.00)%
    \gplgaddtomacro\gplbacktext{%
      \colorrgb{0.00,0.00,0.00}%
      \put(431,736){\makebox(0,0)[r]{\strut{}\small -1}}%
      \colorrgb{0.00,0.00,0.00}%
      \put(431,1493){\makebox(0,0)[r]{\strut{}\small 0}}%
      \colorrgb{0.00,0.00,0.00}%
      \put(431,2250){\makebox(0,0)[r]{\strut{}\small 1}}%
      \colorrgb{0.00,0.00,0.00}%
      \put(431,3007){\makebox(0,0)[r]{\strut{}\small 2}}%
      \colorrgb{0.00,0.00,0.00}%
      \put(563,138){\makebox(0,0){\strut{}\small -2}}%
      \colorrgb{0.00,0.00,0.00}%
      \put(1523,138){\makebox(0,0){\strut{}\small -1}}%
      \colorrgb{0.00,0.00,0.00}%
      \put(2484,138){\makebox(0,0){\strut{}\small 0}}%
      \colorrgb{0.00,0.00,0.00}%
      \put(3444,138){\makebox(0,0){\strut{}\small 1}}%
    }%
    \gplgaddtomacro\gplfronttext{%
    }%
    \gplbacktext
    \put(0,0){\includegraphics{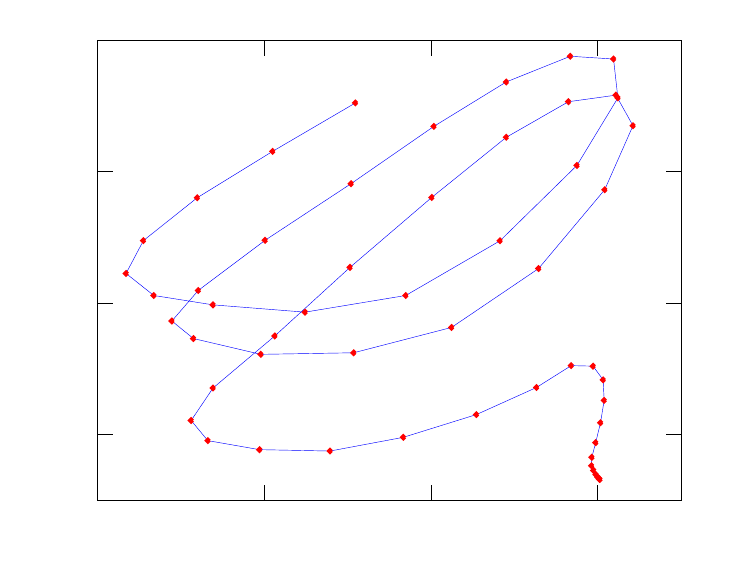}}%
    \gplfronttext
  \end{picture}%
\endgroup
\begingroup
  \makeatletter
  \providecommand\color[2][]{%
    \GenericError{(gnuplot) \space\space\space\@spaces}{%
      Package color not loaded in conjunction with
      terminal option `colourtext'%
    }{See the gnuplot documentation for explanation.%
    }{Either use 'blacktext' in gnuplot or load the package
      color.sty in LaTeX.}%
    \renewcommand\color[2][]{}%
  }%
  \providecommand\includegraphics[2][]{%
    \GenericError{(gnuplot) \space\space\space\@spaces}{%
      Package graphicx or graphics not loaded%
    }{See the gnuplot documentation for explanation.%
    }{The gnuplot epslatex terminal needs graphicx.sty or graphics.sty.}%
    \renewcommand\includegraphics[2][]{}%
  }%
  \providecommand\rotatebox[2]{#2}%
  \@ifundefined{ifGPcolor}{%
    \newif\ifGPcolor
    \GPcolortrue
  }{}%
  \@ifundefined{ifGPblacktext}{%
    \newif\ifGPblacktext
    \GPblacktexttrue
  }{}%
  \let\gplgaddtomacro\g@addto@macro
  \gdef\gplbacktext{}%
  \gdef\gplfronttext{}%
  \makeatother
  \ifGPblacktext
    \def\colorrgb#1{}%
    \def\colorgray#1{}%
  \else
    \ifGPcolor
      \def\colorrgb#1{\color[rgb]{#1}}%
      \def\colorgray#1{\color[gray]{#1}}%
      \expandafter\def\csname LTw\endcsname{\color{white}}%
      \expandafter\def\csname LTb\endcsname{\color{black}}%
      \expandafter\def\csname LTa\endcsname{\color{black}}%
      \expandafter\def\csname LT0\endcsname{\color[rgb]{1,0,0}}%
      \expandafter\def\csname LT1\endcsname{\color[rgb]{0,1,0}}%
      \expandafter\def\csname LT2\endcsname{\color[rgb]{0,0,1}}%
      \expandafter\def\csname LT3\endcsname{\color[rgb]{1,0,1}}%
      \expandafter\def\csname LT4\endcsname{\color[rgb]{0,1,1}}%
      \expandafter\def\csname LT5\endcsname{\color[rgb]{1,1,0}}%
      \expandafter\def\csname LT6\endcsname{\color[rgb]{0,0,0}}%
      \expandafter\def\csname LT7\endcsname{\color[rgb]{1,0.3,0}}%
      \expandafter\def\csname LT8\endcsname{\color[rgb]{0.5,0.5,0.5}}%
    \else
      \def\colorrgb#1{\color{black}}%
      \def\colorgray#1{\color[gray]{#1}}%
      \expandafter\def\csname LTw\endcsname{\color{white}}%
      \expandafter\def\csname LTb\endcsname{\color{black}}%
      \expandafter\def\csname LTa\endcsname{\color{black}}%
      \expandafter\def\csname LT0\endcsname{\color{black}}%
      \expandafter\def\csname LT1\endcsname{\color{black}}%
      \expandafter\def\csname LT2\endcsname{\color{black}}%
      \expandafter\def\csname LT3\endcsname{\color{black}}%
      \expandafter\def\csname LT4\endcsname{\color{black}}%
      \expandafter\def\csname LT5\endcsname{\color{black}}%
      \expandafter\def\csname LT6\endcsname{\color{black}}%
      \expandafter\def\csname LT7\endcsname{\color{black}}%
      \expandafter\def\csname LT8\endcsname{\color{black}}%
    \fi
  \fi
  \setlength{\unitlength}{0.0500bp}%
  \begin{picture}(4336.00,3252.00)%
    \gplgaddtomacro\gplbacktext{%
      \colorrgb{0.00,0.00,0.00}%
      \put(431,736){\makebox(0,0)[r]{\strut{}\small -1}}%
      \colorrgb{0.00,0.00,0.00}%
      \put(431,1493){\makebox(0,0)[r]{\strut{}\small 0}}%
      \colorrgb{0.00,0.00,0.00}%
      \put(431,2250){\makebox(0,0)[r]{\strut{}\small 1}}%
      \colorrgb{0.00,0.00,0.00}%
      \put(431,3007){\makebox(0,0)[r]{\strut{}\small 2}}%
      \colorrgb{0.00,0.00,0.00}%
      \put(563,138){\makebox(0,0){\strut{}\small -2}}%
      \colorrgb{0.00,0.00,0.00}%
      \put(1523,138){\makebox(0,0){\strut{}\small -1}}%
      \colorrgb{0.00,0.00,0.00}%
      \put(2484,138){\makebox(0,0){\strut{}\small 0}}%
      \colorrgb{0.00,0.00,0.00}%
      \put(3444,138){\makebox(0,0){\strut{}\small 1}}%
    }%
    \gplgaddtomacro\gplfronttext{%
    }%
    \gplbacktext
    \put(0,0){\includegraphics{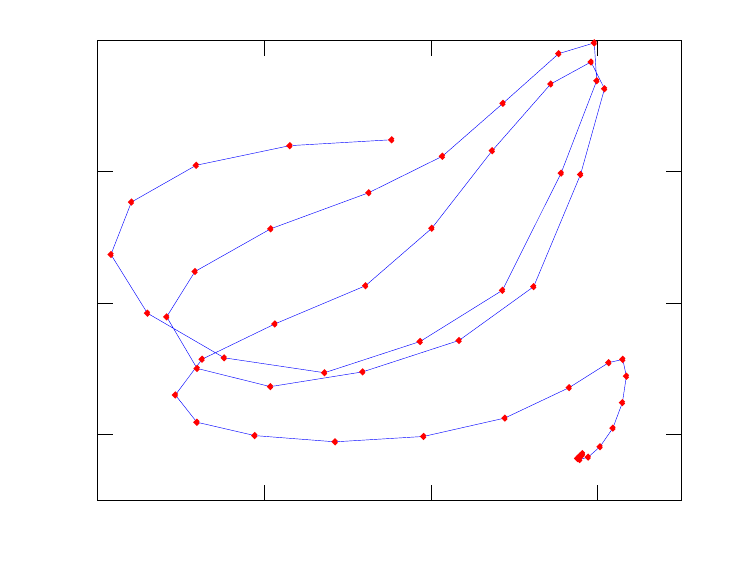}}%
    \gplfronttext
  \end{picture}%
\endgroup
  \end{center}
  \caption{Motion capture data visualized in two dimensions for models
    derived from the maximum entropy perspective. Again for each
    algorithm we used 6 nearest neighbors. }\label{fig:embedStick2}

\end{figure}

The two dominant eigenvectors are visualized in
\reffigrange{fig:embedStick1}{fig:embedStick2} and the quality of the
visualizations under the GP-LVM likelihood is given in
\reffig{fig:histograms}(a).
\begin{octave}
  textSize = '\\small';
  clear score;
  models = {'le1', 'lle1', 'isomap1', 'mvu1', 'meu1', 'lle2', 'drill1'};
  names = {[textSize ' Laplacian eigenmaps'],  [textSize ' LLE'],  [textSize ' isomap'],  [textSize ' MVU'],  [textSize ' MEU'],  [textSize ' Acyclic LLE'], [textSize ' DRILL']};
  for i = 1:length(models);
    loadName = models{i};
    loadName(1) = upper(loadName(1));
    load(['demStick' loadName '.mat']);
    eval(['score(i) = ' models{i}(1:end-1) 'Info.score;']);
    eval(['X = ' models{i}(1:end-1) 'Info.X;']);
    figure(1)
    clf
    plot(X(:, 1), X(:, 2), 'b-');
    hold on
    b = plot(X(:, 1), X(:, 2), 'ro');
    set(b, 'markersize', 2, 'linewidth', 3);
    set(gca, 'xtick', [-3 -2 -1 0 1 2 3]);
    set(gca, 'xticklabel', [-3 -2 -1 0 1 2 3]);
    set(gca, 'ytick', [-3 -2 -1 0 1 2 3]);
    set(gca, 'yticklabel', [-3 -2 -1 0 1 2 3]);
    set(gca, 'xticklabel', {[textSize ' -3'], [textSize ' -2'], [textSize ' -1'], [textSize ' 0'], [textSize ' 1'], [textSize ' 2'], [textSize ' 3']});
    set(gca, 'ytick', [-3 -2 -1 0 1 2 3]);
    set(gca, 'yticklabel', {[textSize ' -3'], [textSize ' -2'], [textSize ' -1'], [textSize ' 0'], [textSize ' 1'], [textSize ' 2'], [textSize ' 3']});
    printLatexPlot(['demStick' loadName], '../../../meu/tex/diagrams/', 0.45*textWidth)
  end
  clf
  barh(score);
  set(gca, 'ytick', [1:length(names)])
  set(gca, 'yticklabel', names)
  set(gca, 'xtick', [0 2000 4000]);
  
  printLatexPlot('demStickBar1', '../../../meu/tex/diagrams/', 0.65*textWidth)
\end{octave}

There is a clear difference in quality between the methods that
constrain local distances (ALLE, MVU, isomap, MEU and DRILL) which are
much better under the score than those that don't (Laplacian eigenmaps
and LLE). Amongst the distance preserving methods isomap is the best
performer under the GPLVM score, followed by ALLE, MVU, DRILL and
MEU. The MEU model here preserves the positive definiteness of the
covariance by constraining the Lagrange multipliers to be positive (an
`attractive' network as discussed in
\refsec{sec:positivedefinite}). It may be that this departure from the
true maximum entropy framework explains its relatively poorer
performance..

\subsection{Robot Navigation Example}

The second data set we use is a series of recordings from a robot as
it traces a square path in a building. The robot records the strength
of WiFi signals in an attempt to localize its position \citep[see][for
an application]{Ferris:wifi07}. Since the robot moves only in two
dimensions, the inherent dimensionality of the data should be two: the
reduced dimensional space should reflect the robot's movement. The WiFi
signals are noisier than the motion capture data, so it makes an
interesting contrast. The robot completes a single circuit after
entering from a separate corridor, so it is expected to exhibit ``loop
closure'' in the resulting map. The data consists of 215 frames of
measurement, each frame consists of the WiFi signal strength of 30
access points.

\begin{figure}
\begingroup
  \makeatletter
  \providecommand\color[2][]{%
    \GenericError{(gnuplot) \space\space\space\@spaces}{%
      Package color not loaded in conjunction with
      terminal option `colourtext'%
    }{See the gnuplot documentation for explanation.%
    }{Either use 'blacktext' in gnuplot or load the package
      color.sty in LaTeX.}%
    \renewcommand\color[2][]{}%
  }%
  \providecommand\includegraphics[2][]{%
    \GenericError{(gnuplot) \space\space\space\@spaces}{%
      Package graphicx or graphics not loaded%
    }{See the gnuplot documentation for explanation.%
    }{The gnuplot epslatex terminal needs graphicx.sty or graphics.sty.}%
    \renewcommand\includegraphics[2][]{}%
  }%
  \providecommand\rotatebox[2]{#2}%
  \@ifundefined{ifGPcolor}{%
    \newif\ifGPcolor
    \GPcolortrue
  }{}%
  \@ifundefined{ifGPblacktext}{%
    \newif\ifGPblacktext
    \GPblacktexttrue
  }{}%
  \let\gplgaddtomacro\g@addto@macro
  \gdef\gplbacktext{}%
  \gdef\gplfronttext{}%
  \makeatother
  \ifGPblacktext
    \def\colorrgb#1{}%
    \def\colorgray#1{}%
  \else
    \ifGPcolor
      \def\colorrgb#1{\color[rgb]{#1}}%
      \def\colorgray#1{\color[gray]{#1}}%
      \expandafter\def\csname LTw\endcsname{\color{white}}%
      \expandafter\def\csname LTb\endcsname{\color{black}}%
      \expandafter\def\csname LTa\endcsname{\color{black}}%
      \expandafter\def\csname LT0\endcsname{\color[rgb]{1,0,0}}%
      \expandafter\def\csname LT1\endcsname{\color[rgb]{0,1,0}}%
      \expandafter\def\csname LT2\endcsname{\color[rgb]{0,0,1}}%
      \expandafter\def\csname LT3\endcsname{\color[rgb]{1,0,1}}%
      \expandafter\def\csname LT4\endcsname{\color[rgb]{0,1,1}}%
      \expandafter\def\csname LT5\endcsname{\color[rgb]{1,1,0}}%
      \expandafter\def\csname LT6\endcsname{\color[rgb]{0,0,0}}%
      \expandafter\def\csname LT7\endcsname{\color[rgb]{1,0.3,0}}%
      \expandafter\def\csname LT8\endcsname{\color[rgb]{0.5,0.5,0.5}}%
    \else
      \def\colorrgb#1{\color{black}}%
      \def\colorgray#1{\color[gray]{#1}}%
      \expandafter\def\csname LTw\endcsname{\color{white}}%
      \expandafter\def\csname LTb\endcsname{\color{black}}%
      \expandafter\def\csname LTa\endcsname{\color{black}}%
      \expandafter\def\csname LT0\endcsname{\color{black}}%
      \expandafter\def\csname LT1\endcsname{\color{black}}%
      \expandafter\def\csname LT2\endcsname{\color{black}}%
      \expandafter\def\csname LT3\endcsname{\color{black}}%
      \expandafter\def\csname LT4\endcsname{\color{black}}%
      \expandafter\def\csname LT5\endcsname{\color{black}}%
      \expandafter\def\csname LT6\endcsname{\color{black}}%
      \expandafter\def\csname LT7\endcsname{\color{black}}%
      \expandafter\def\csname LT8\endcsname{\color{black}}%
    \fi
  \fi
  \setlength{\unitlength}{0.0500bp}%
  \begin{picture}(4336.00,3252.00)%
    \gplgaddtomacro\gplbacktext{%
      \colorrgb{0.00,0.00,0.00}%
      \put(431,358){\makebox(0,0)[r]{\strut{}\small -3}}%
      \colorrgb{0.00,0.00,0.00}%
      \put(431,888){\makebox(0,0)[r]{\strut{}\small -2}}%
      \colorrgb{0.00,0.00,0.00}%
      \put(431,1418){\makebox(0,0)[r]{\strut{}\small -1}}%
      \colorrgb{0.00,0.00,0.00}%
      \put(431,1947){\makebox(0,0)[r]{\strut{}\small 0}}%
      \colorrgb{0.00,0.00,0.00}%
      \put(431,2477){\makebox(0,0)[r]{\strut{}\small 1}}%
      \colorrgb{0.00,0.00,0.00}%
      \put(431,3007){\makebox(0,0)[r]{\strut{}\small 2}}%
      \colorrgb{0.00,0.00,0.00}%
      \put(563,138){\makebox(0,0){\strut{}\small -3}}%
      \colorrgb{0.00,0.00,0.00}%
      \put(1235,138){\makebox(0,0){\strut{}\small -2}}%
      \colorrgb{0.00,0.00,0.00}%
      \put(1907,138){\makebox(0,0){\strut{}\small -1}}%
      \colorrgb{0.00,0.00,0.00}%
      \put(2580,138){\makebox(0,0){\strut{}\small 0}}%
      \colorrgb{0.00,0.00,0.00}%
      \put(3252,138){\makebox(0,0){\strut{}\small 1}}%
      \colorrgb{0.00,0.00,0.00}%
      \put(3924,138){\makebox(0,0){\strut{}\small 2}}%
    }%
    \gplgaddtomacro\gplfronttext{%
    }%
    \gplbacktext
    \put(0,0){\includegraphics{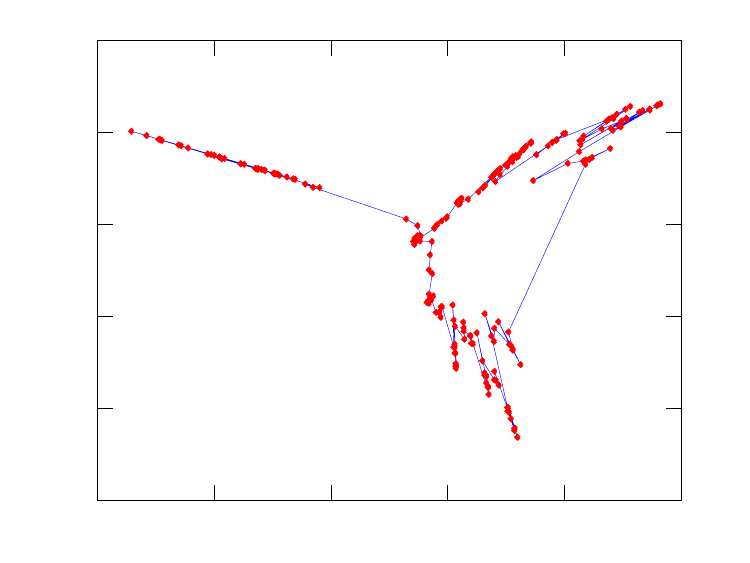}}%
    \gplfronttext
  \end{picture}%
\endgroup
  \hfill
\begingroup
  \makeatletter
  \providecommand\color[2][]{%
    \GenericError{(gnuplot) \space\space\space\@spaces}{%
      Package color not loaded in conjunction with
      terminal option `colourtext'%
    }{See the gnuplot documentation for explanation.%
    }{Either use 'blacktext' in gnuplot or load the package
      color.sty in LaTeX.}%
    \renewcommand\color[2][]{}%
  }%
  \providecommand\includegraphics[2][]{%
    \GenericError{(gnuplot) \space\space\space\@spaces}{%
      Package graphicx or graphics not loaded%
    }{See the gnuplot documentation for explanation.%
    }{The gnuplot epslatex terminal needs graphicx.sty or graphics.sty.}%
    \renewcommand\includegraphics[2][]{}%
  }%
  \providecommand\rotatebox[2]{#2}%
  \@ifundefined{ifGPcolor}{%
    \newif\ifGPcolor
    \GPcolortrue
  }{}%
  \@ifundefined{ifGPblacktext}{%
    \newif\ifGPblacktext
    \GPblacktexttrue
  }{}%
  \let\gplgaddtomacro\g@addto@macro
  \gdef\gplbacktext{}%
  \gdef\gplfronttext{}%
  \makeatother
  \ifGPblacktext
    \def\colorrgb#1{}%
    \def\colorgray#1{}%
  \else
    \ifGPcolor
      \def\colorrgb#1{\color[rgb]{#1}}%
      \def\colorgray#1{\color[gray]{#1}}%
      \expandafter\def\csname LTw\endcsname{\color{white}}%
      \expandafter\def\csname LTb\endcsname{\color{black}}%
      \expandafter\def\csname LTa\endcsname{\color{black}}%
      \expandafter\def\csname LT0\endcsname{\color[rgb]{1,0,0}}%
      \expandafter\def\csname LT1\endcsname{\color[rgb]{0,1,0}}%
      \expandafter\def\csname LT2\endcsname{\color[rgb]{0,0,1}}%
      \expandafter\def\csname LT3\endcsname{\color[rgb]{1,0,1}}%
      \expandafter\def\csname LT4\endcsname{\color[rgb]{0,1,1}}%
      \expandafter\def\csname LT5\endcsname{\color[rgb]{1,1,0}}%
      \expandafter\def\csname LT6\endcsname{\color[rgb]{0,0,0}}%
      \expandafter\def\csname LT7\endcsname{\color[rgb]{1,0.3,0}}%
      \expandafter\def\csname LT8\endcsname{\color[rgb]{0.5,0.5,0.5}}%
    \else
      \def\colorrgb#1{\color{black}}%
      \def\colorgray#1{\color[gray]{#1}}%
      \expandafter\def\csname LTw\endcsname{\color{white}}%
      \expandafter\def\csname LTb\endcsname{\color{black}}%
      \expandafter\def\csname LTa\endcsname{\color{black}}%
      \expandafter\def\csname LT0\endcsname{\color{black}}%
      \expandafter\def\csname LT1\endcsname{\color{black}}%
      \expandafter\def\csname LT2\endcsname{\color{black}}%
      \expandafter\def\csname LT3\endcsname{\color{black}}%
      \expandafter\def\csname LT4\endcsname{\color{black}}%
      \expandafter\def\csname LT5\endcsname{\color{black}}%
      \expandafter\def\csname LT6\endcsname{\color{black}}%
      \expandafter\def\csname LT7\endcsname{\color{black}}%
      \expandafter\def\csname LT8\endcsname{\color{black}}%
    \fi
  \fi
  \setlength{\unitlength}{0.0500bp}%
  \begin{picture}(4336.00,3252.00)%
    \gplgaddtomacro\gplbacktext{%
      \colorrgb{0.00,0.00,0.00}%
      \put(431,888){\makebox(0,0)[r]{\strut{}\small -3}}%
      \colorrgb{0.00,0.00,0.00}%
      \put(431,1418){\makebox(0,0)[r]{\strut{}\small -2}}%
      \colorrgb{0.00,0.00,0.00}%
      \put(431,1947){\makebox(0,0)[r]{\strut{}\small -1}}%
      \colorrgb{0.00,0.00,0.00}%
      \put(431,2477){\makebox(0,0)[r]{\strut{}\small 0}}%
      \colorrgb{0.00,0.00,0.00}%
      \put(431,3007){\makebox(0,0)[r]{\strut{}\small 1}}%
      \colorrgb{0.00,0.00,0.00}%
      \put(983,138){\makebox(0,0){\strut{}\small 0}}%
      \colorrgb{0.00,0.00,0.00}%
      \put(3084,138){\makebox(0,0){\strut{}\small 1}}%
    }%
    \gplgaddtomacro\gplfronttext{%
    }%
    \gplbacktext
    \put(0,0){\includegraphics{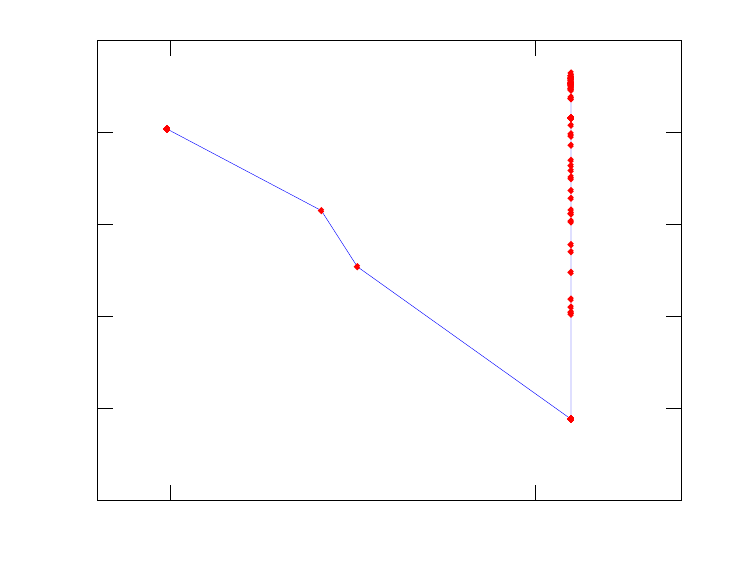}}%
    \gplfronttext
  \end{picture}%
\endgroup
  \hfill
\begingroup
  \makeatletter
  \providecommand\color[2][]{%
    \GenericError{(gnuplot) \space\space\space\@spaces}{%
      Package color not loaded in conjunction with
      terminal option `colourtext'%
    }{See the gnuplot documentation for explanation.%
    }{Either use 'blacktext' in gnuplot or load the package
      color.sty in LaTeX.}%
    \renewcommand\color[2][]{}%
  }%
  \providecommand\includegraphics[2][]{%
    \GenericError{(gnuplot) \space\space\space\@spaces}{%
      Package graphicx or graphics not loaded%
    }{See the gnuplot documentation for explanation.%
    }{The gnuplot epslatex terminal needs graphicx.sty or graphics.sty.}%
    \renewcommand\includegraphics[2][]{}%
  }%
  \providecommand\rotatebox[2]{#2}%
  \@ifundefined{ifGPcolor}{%
    \newif\ifGPcolor
    \GPcolortrue
  }{}%
  \@ifundefined{ifGPblacktext}{%
    \newif\ifGPblacktext
    \GPblacktexttrue
  }{}%
  \let\gplgaddtomacro\g@addto@macro
  \gdef\gplbacktext{}%
  \gdef\gplfronttext{}%
  \makeatother
  \ifGPblacktext
    \def\colorrgb#1{}%
    \def\colorgray#1{}%
  \else
    \ifGPcolor
      \def\colorrgb#1{\color[rgb]{#1}}%
      \def\colorgray#1{\color[gray]{#1}}%
      \expandafter\def\csname LTw\endcsname{\color{white}}%
      \expandafter\def\csname LTb\endcsname{\color{black}}%
      \expandafter\def\csname LTa\endcsname{\color{black}}%
      \expandafter\def\csname LT0\endcsname{\color[rgb]{1,0,0}}%
      \expandafter\def\csname LT1\endcsname{\color[rgb]{0,1,0}}%
      \expandafter\def\csname LT2\endcsname{\color[rgb]{0,0,1}}%
      \expandafter\def\csname LT3\endcsname{\color[rgb]{1,0,1}}%
      \expandafter\def\csname LT4\endcsname{\color[rgb]{0,1,1}}%
      \expandafter\def\csname LT5\endcsname{\color[rgb]{1,1,0}}%
      \expandafter\def\csname LT6\endcsname{\color[rgb]{0,0,0}}%
      \expandafter\def\csname LT7\endcsname{\color[rgb]{1,0.3,0}}%
      \expandafter\def\csname LT8\endcsname{\color[rgb]{0.5,0.5,0.5}}%
    \else
      \def\colorrgb#1{\color{black}}%
      \def\colorgray#1{\color[gray]{#1}}%
      \expandafter\def\csname LTw\endcsname{\color{white}}%
      \expandafter\def\csname LTb\endcsname{\color{black}}%
      \expandafter\def\csname LTa\endcsname{\color{black}}%
      \expandafter\def\csname LT0\endcsname{\color{black}}%
      \expandafter\def\csname LT1\endcsname{\color{black}}%
      \expandafter\def\csname LT2\endcsname{\color{black}}%
      \expandafter\def\csname LT3\endcsname{\color{black}}%
      \expandafter\def\csname LT4\endcsname{\color{black}}%
      \expandafter\def\csname LT5\endcsname{\color{black}}%
      \expandafter\def\csname LT6\endcsname{\color{black}}%
      \expandafter\def\csname LT7\endcsname{\color{black}}%
      \expandafter\def\csname LT8\endcsname{\color{black}}%
    \fi
  \fi
  \setlength{\unitlength}{0.0500bp}%
  \begin{picture}(4336.00,3252.00)%
    \gplgaddtomacro\gplbacktext{%
      \colorrgb{0.00,0.00,0.00}%
      \put(431,358){\makebox(0,0)[r]{\strut{}\small -2}}%
      \colorrgb{0.00,0.00,0.00}%
      \put(431,1020){\makebox(0,0)[r]{\strut{}\small -1}}%
      \colorrgb{0.00,0.00,0.00}%
      \put(431,1683){\makebox(0,0)[r]{\strut{}\small 0}}%
      \colorrgb{0.00,0.00,0.00}%
      \put(431,2345){\makebox(0,0)[r]{\strut{}\small 1}}%
      \colorrgb{0.00,0.00,0.00}%
      \put(431,3007){\makebox(0,0)[r]{\strut{}\small 2}}%
      \colorrgb{0.00,0.00,0.00}%
      \put(563,138){\makebox(0,0){\strut{}\small -3}}%
      \colorrgb{0.00,0.00,0.00}%
      \put(1235,138){\makebox(0,0){\strut{}\small -2}}%
      \colorrgb{0.00,0.00,0.00}%
      \put(1907,138){\makebox(0,0){\strut{}\small -1}}%
      \colorrgb{0.00,0.00,0.00}%
      \put(2580,138){\makebox(0,0){\strut{}\small 0}}%
      \colorrgb{0.00,0.00,0.00}%
      \put(3252,138){\makebox(0,0){\strut{}\small 1}}%
      \colorrgb{0.00,0.00,0.00}%
      \put(3924,138){\makebox(0,0){\strut{}\small 2}}%
    }%
    \gplgaddtomacro\gplfronttext{%
    }%
    \gplbacktext
    \put(0,0){\includegraphics{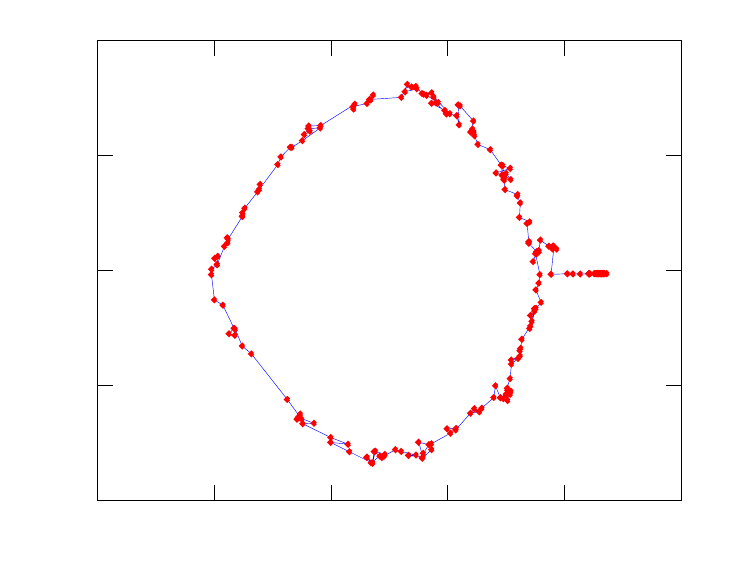}}%
    \gplfronttext
  \end{picture}%
\endgroup
  \hfill
\begingroup
  \makeatletter
  \providecommand\color[2][]{%
    \GenericError{(gnuplot) \space\space\space\@spaces}{%
      Package color not loaded in conjunction with
      terminal option `colourtext'%
    }{See the gnuplot documentation for explanation.%
    }{Either use 'blacktext' in gnuplot or load the package
      color.sty in LaTeX.}%
    \renewcommand\color[2][]{}%
  }%
  \providecommand\includegraphics[2][]{%
    \GenericError{(gnuplot) \space\space\space\@spaces}{%
      Package graphicx or graphics not loaded%
    }{See the gnuplot documentation for explanation.%
    }{The gnuplot epslatex terminal needs graphicx.sty or graphics.sty.}%
    \renewcommand\includegraphics[2][]{}%
  }%
  \providecommand\rotatebox[2]{#2}%
  \@ifundefined{ifGPcolor}{%
    \newif\ifGPcolor
    \GPcolortrue
  }{}%
  \@ifundefined{ifGPblacktext}{%
    \newif\ifGPblacktext
    \GPblacktexttrue
  }{}%
  \let\gplgaddtomacro\g@addto@macro
  \gdef\gplbacktext{}%
  \gdef\gplfronttext{}%
  \makeatother
  \ifGPblacktext
    \def\colorrgb#1{}%
    \def\colorgray#1{}%
  \else
    \ifGPcolor
      \def\colorrgb#1{\color[rgb]{#1}}%
      \def\colorgray#1{\color[gray]{#1}}%
      \expandafter\def\csname LTw\endcsname{\color{white}}%
      \expandafter\def\csname LTb\endcsname{\color{black}}%
      \expandafter\def\csname LTa\endcsname{\color{black}}%
      \expandafter\def\csname LT0\endcsname{\color[rgb]{1,0,0}}%
      \expandafter\def\csname LT1\endcsname{\color[rgb]{0,1,0}}%
      \expandafter\def\csname LT2\endcsname{\color[rgb]{0,0,1}}%
      \expandafter\def\csname LT3\endcsname{\color[rgb]{1,0,1}}%
      \expandafter\def\csname LT4\endcsname{\color[rgb]{0,1,1}}%
      \expandafter\def\csname LT5\endcsname{\color[rgb]{1,1,0}}%
      \expandafter\def\csname LT6\endcsname{\color[rgb]{0,0,0}}%
      \expandafter\def\csname LT7\endcsname{\color[rgb]{1,0.3,0}}%
      \expandafter\def\csname LT8\endcsname{\color[rgb]{0.5,0.5,0.5}}%
    \else
      \def\colorrgb#1{\color{black}}%
      \def\colorgray#1{\color[gray]{#1}}%
      \expandafter\def\csname LTw\endcsname{\color{white}}%
      \expandafter\def\csname LTb\endcsname{\color{black}}%
      \expandafter\def\csname LTa\endcsname{\color{black}}%
      \expandafter\def\csname LT0\endcsname{\color{black}}%
      \expandafter\def\csname LT1\endcsname{\color{black}}%
      \expandafter\def\csname LT2\endcsname{\color{black}}%
      \expandafter\def\csname LT3\endcsname{\color{black}}%
      \expandafter\def\csname LT4\endcsname{\color{black}}%
      \expandafter\def\csname LT5\endcsname{\color{black}}%
      \expandafter\def\csname LT6\endcsname{\color{black}}%
      \expandafter\def\csname LT7\endcsname{\color{black}}%
      \expandafter\def\csname LT8\endcsname{\color{black}}%
    \fi
  \fi
  \setlength{\unitlength}{0.0500bp}%
  \begin{picture}(4336.00,3252.00)%
    \gplgaddtomacro\gplbacktext{%
      \colorrgb{0.00,0.00,0.00}%
      \put(431,689){\makebox(0,0)[r]{\strut{}\small -3}}%
      \colorrgb{0.00,0.00,0.00}%
      \put(431,1020){\makebox(0,0)[r]{\strut{}\small -2}}%
      \colorrgb{0.00,0.00,0.00}%
      \put(431,1351){\makebox(0,0)[r]{\strut{}\small -1}}%
      \colorrgb{0.00,0.00,0.00}%
      \put(431,1683){\makebox(0,0)[r]{\strut{}\small 0}}%
      \colorrgb{0.00,0.00,0.00}%
      \put(431,2014){\makebox(0,0)[r]{\strut{}\small 1}}%
      \colorrgb{0.00,0.00,0.00}%
      \put(431,2345){\makebox(0,0)[r]{\strut{}\small 2}}%
      \colorrgb{0.00,0.00,0.00}%
      \put(431,2676){\makebox(0,0)[r]{\strut{}\small 3}}%
      \colorrgb{0.00,0.00,0.00}%
      \put(1739,138){\makebox(0,0){\strut{}\small -3}}%
      \colorrgb{0.00,0.00,0.00}%
      \put(1907,138){\makebox(0,0){\strut{}\small -2}}%
      \colorrgb{0.00,0.00,0.00}%
      \put(2075,138){\makebox(0,0){\strut{}\small -1}}%
      \colorrgb{0.00,0.00,0.00}%
      \put(2244,138){\makebox(0,0){\strut{}\small 0}}%
      \colorrgb{0.00,0.00,0.00}%
      \put(2412,138){\makebox(0,0){\strut{}\small 1}}%
      \colorrgb{0.00,0.00,0.00}%
      \put(2580,138){\makebox(0,0){\strut{}\small 2}}%
      \colorrgb{0.00,0.00,0.00}%
      \put(2748,138){\makebox(0,0){\strut{}\small 3}}%
    }%
    \gplgaddtomacro\gplfronttext{%
    }%
    \gplbacktext
    \put(0,0){\includegraphics{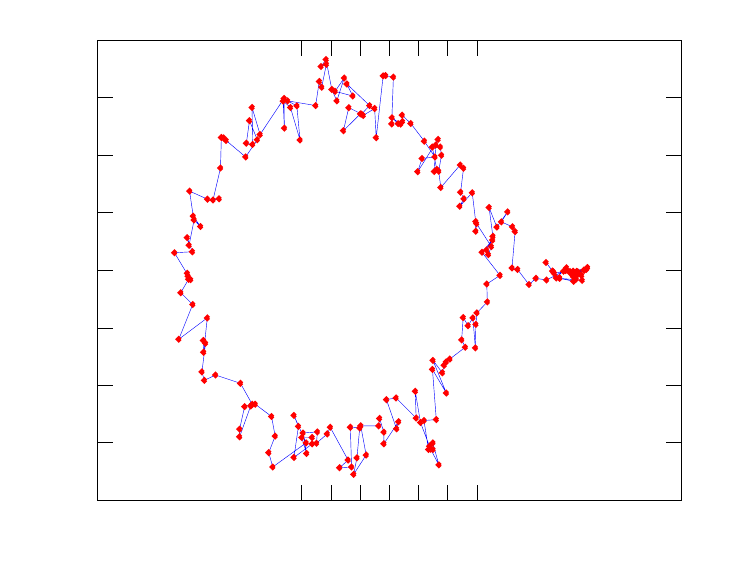}}%
    \gplfronttext
  \end{picture}%
\endgroup
\caption{Visualization of the robot WiFi navigation data for different spectral algorithms we reviewed with seven neighbors used to construct graphs. LE and LLE struggle to captured the loop structure (perhaps because of the higher level of noise). Several of the models also show the noise present in the WiFi signals.}\label{fig:embedRobot1}
\end{figure}

\begin{figure}
  \begin{center}
\begingroup
  \makeatletter
  \providecommand\color[2][]{%
    \GenericError{(gnuplot) \space\space\space\@spaces}{%
      Package color not loaded in conjunction with
      terminal option `colourtext'%
    }{See the gnuplot documentation for explanation.%
    }{Either use 'blacktext' in gnuplot or load the package
      color.sty in LaTeX.}%
    \renewcommand\color[2][]{}%
  }%
  \providecommand\includegraphics[2][]{%
    \GenericError{(gnuplot) \space\space\space\@spaces}{%
      Package graphicx or graphics not loaded%
    }{See the gnuplot documentation for explanation.%
    }{The gnuplot epslatex terminal needs graphicx.sty or graphics.sty.}%
    \renewcommand\includegraphics[2][]{}%
  }%
  \providecommand\rotatebox[2]{#2}%
  \@ifundefined{ifGPcolor}{%
    \newif\ifGPcolor
    \GPcolortrue
  }{}%
  \@ifundefined{ifGPblacktext}{%
    \newif\ifGPblacktext
    \GPblacktexttrue
  }{}%
  \let\gplgaddtomacro\g@addto@macro
  \gdef\gplbacktext{}%
  \gdef\gplfronttext{}%
  \makeatother
  \ifGPblacktext
    \def\colorrgb#1{}%
    \def\colorgray#1{}%
  \else
    \ifGPcolor
      \def\colorrgb#1{\color[rgb]{#1}}%
      \def\colorgray#1{\color[gray]{#1}}%
      \expandafter\def\csname LTw\endcsname{\color{white}}%
      \expandafter\def\csname LTb\endcsname{\color{black}}%
      \expandafter\def\csname LTa\endcsname{\color{black}}%
      \expandafter\def\csname LT0\endcsname{\color[rgb]{1,0,0}}%
      \expandafter\def\csname LT1\endcsname{\color[rgb]{0,1,0}}%
      \expandafter\def\csname LT2\endcsname{\color[rgb]{0,0,1}}%
      \expandafter\def\csname LT3\endcsname{\color[rgb]{1,0,1}}%
      \expandafter\def\csname LT4\endcsname{\color[rgb]{0,1,1}}%
      \expandafter\def\csname LT5\endcsname{\color[rgb]{1,1,0}}%
      \expandafter\def\csname LT6\endcsname{\color[rgb]{0,0,0}}%
      \expandafter\def\csname LT7\endcsname{\color[rgb]{1,0.3,0}}%
      \expandafter\def\csname LT8\endcsname{\color[rgb]{0.5,0.5,0.5}}%
    \else
      \def\colorrgb#1{\color{black}}%
      \def\colorgray#1{\color[gray]{#1}}%
      \expandafter\def\csname LTw\endcsname{\color{white}}%
      \expandafter\def\csname LTb\endcsname{\color{black}}%
      \expandafter\def\csname LTa\endcsname{\color{black}}%
      \expandafter\def\csname LT0\endcsname{\color{black}}%
      \expandafter\def\csname LT1\endcsname{\color{black}}%
      \expandafter\def\csname LT2\endcsname{\color{black}}%
      \expandafter\def\csname LT3\endcsname{\color{black}}%
      \expandafter\def\csname LT4\endcsname{\color{black}}%
      \expandafter\def\csname LT5\endcsname{\color{black}}%
      \expandafter\def\csname LT6\endcsname{\color{black}}%
      \expandafter\def\csname LT7\endcsname{\color{black}}%
      \expandafter\def\csname LT8\endcsname{\color{black}}%
    \fi
  \fi
  \setlength{\unitlength}{0.0500bp}%
  \begin{picture}(4336.00,3252.00)%
    \gplgaddtomacro\gplbacktext{%
      \colorrgb{0.00,0.00,0.00}%
      \put(431,736){\makebox(0,0)[r]{\strut{}\small -1}}%
      \colorrgb{0.00,0.00,0.00}%
      \put(431,1493){\makebox(0,0)[r]{\strut{}\small 0}}%
      \colorrgb{0.00,0.00,0.00}%
      \put(431,2250){\makebox(0,0)[r]{\strut{}\small 1}}%
      \colorrgb{0.00,0.00,0.00}%
      \put(431,3007){\makebox(0,0)[r]{\strut{}\small 2}}%
      \colorrgb{0.00,0.00,0.00}%
      \put(563,138){\makebox(0,0){\strut{}\small -2}}%
      \colorrgb{0.00,0.00,0.00}%
      \put(1403,138){\makebox(0,0){\strut{}\small -1}}%
      \colorrgb{0.00,0.00,0.00}%
      \put(2244,138){\makebox(0,0){\strut{}\small 0}}%
      \colorrgb{0.00,0.00,0.00}%
      \put(3084,138){\makebox(0,0){\strut{}\small 1}}%
      \colorrgb{0.00,0.00,0.00}%
      \put(3924,138){\makebox(0,0){\strut{}\small 2}}%
    }%
    \gplgaddtomacro\gplfronttext{%
    }%
    \gplbacktext
    \put(0,0){\includegraphics{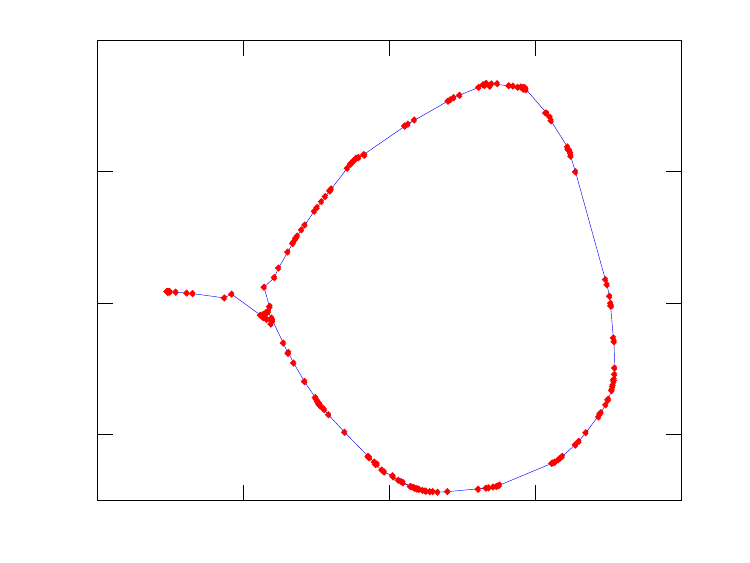}}%
    \gplfronttext
  \end{picture}%
\endgroup
  \hfill
\begingroup
  \makeatletter
  \providecommand\color[2][]{%
    \GenericError{(gnuplot) \space\space\space\@spaces}{%
      Package color not loaded in conjunction with
      terminal option `colourtext'%
    }{See the gnuplot documentation for explanation.%
    }{Either use 'blacktext' in gnuplot or load the package
      color.sty in LaTeX.}%
    \renewcommand\color[2][]{}%
  }%
  \providecommand\includegraphics[2][]{%
    \GenericError{(gnuplot) \space\space\space\@spaces}{%
      Package graphicx or graphics not loaded%
    }{See the gnuplot documentation for explanation.%
    }{The gnuplot epslatex terminal needs graphicx.sty or graphics.sty.}%
    \renewcommand\includegraphics[2][]{}%
  }%
  \providecommand\rotatebox[2]{#2}%
  \@ifundefined{ifGPcolor}{%
    \newif\ifGPcolor
    \GPcolortrue
  }{}%
  \@ifundefined{ifGPblacktext}{%
    \newif\ifGPblacktext
    \GPblacktexttrue
  }{}%
  \let\gplgaddtomacro\g@addto@macro
  \gdef\gplbacktext{}%
  \gdef\gplfronttext{}%
  \makeatother
  \ifGPblacktext
    \def\colorrgb#1{}%
    \def\colorgray#1{}%
  \else
    \ifGPcolor
      \def\colorrgb#1{\color[rgb]{#1}}%
      \def\colorgray#1{\color[gray]{#1}}%
      \expandafter\def\csname LTw\endcsname{\color{white}}%
      \expandafter\def\csname LTb\endcsname{\color{black}}%
      \expandafter\def\csname LTa\endcsname{\color{black}}%
      \expandafter\def\csname LT0\endcsname{\color[rgb]{1,0,0}}%
      \expandafter\def\csname LT1\endcsname{\color[rgb]{0,1,0}}%
      \expandafter\def\csname LT2\endcsname{\color[rgb]{0,0,1}}%
      \expandafter\def\csname LT3\endcsname{\color[rgb]{1,0,1}}%
      \expandafter\def\csname LT4\endcsname{\color[rgb]{0,1,1}}%
      \expandafter\def\csname LT5\endcsname{\color[rgb]{1,1,0}}%
      \expandafter\def\csname LT6\endcsname{\color[rgb]{0,0,0}}%
      \expandafter\def\csname LT7\endcsname{\color[rgb]{1,0.3,0}}%
      \expandafter\def\csname LT8\endcsname{\color[rgb]{0.5,0.5,0.5}}%
    \else
      \def\colorrgb#1{\color{black}}%
      \def\colorgray#1{\color[gray]{#1}}%
      \expandafter\def\csname LTw\endcsname{\color{white}}%
      \expandafter\def\csname LTb\endcsname{\color{black}}%
      \expandafter\def\csname LTa\endcsname{\color{black}}%
      \expandafter\def\csname LT0\endcsname{\color{black}}%
      \expandafter\def\csname LT1\endcsname{\color{black}}%
      \expandafter\def\csname LT2\endcsname{\color{black}}%
      \expandafter\def\csname LT3\endcsname{\color{black}}%
      \expandafter\def\csname LT4\endcsname{\color{black}}%
      \expandafter\def\csname LT5\endcsname{\color{black}}%
      \expandafter\def\csname LT6\endcsname{\color{black}}%
      \expandafter\def\csname LT7\endcsname{\color{black}}%
      \expandafter\def\csname LT8\endcsname{\color{black}}%
    \fi
  \fi
  \setlength{\unitlength}{0.0500bp}%
  \begin{picture}(4336.00,3252.00)%
    \gplgaddtomacro\gplbacktext{%
      \colorrgb{0.00,0.00,0.00}%
      \put(431,736){\makebox(0,0)[r]{\strut{}\small -1}}%
      \colorrgb{0.00,0.00,0.00}%
      \put(431,1493){\makebox(0,0)[r]{\strut{}\small 0}}%
      \colorrgb{0.00,0.00,0.00}%
      \put(431,2250){\makebox(0,0)[r]{\strut{}\small 1}}%
      \colorrgb{0.00,0.00,0.00}%
      \put(431,3007){\makebox(0,0)[r]{\strut{}\small 2}}%
      \colorrgb{0.00,0.00,0.00}%
      \put(563,138){\makebox(0,0){\strut{}\small -2}}%
      \colorrgb{0.00,0.00,0.00}%
      \put(1523,138){\makebox(0,0){\strut{}\small -1}}%
      \colorrgb{0.00,0.00,0.00}%
      \put(2484,138){\makebox(0,0){\strut{}\small 0}}%
      \colorrgb{0.00,0.00,0.00}%
      \put(3444,138){\makebox(0,0){\strut{}\small 1}}%
    }%
    \gplgaddtomacro\gplfronttext{%
    }%
    \gplbacktext
    \put(0,0){\includegraphics{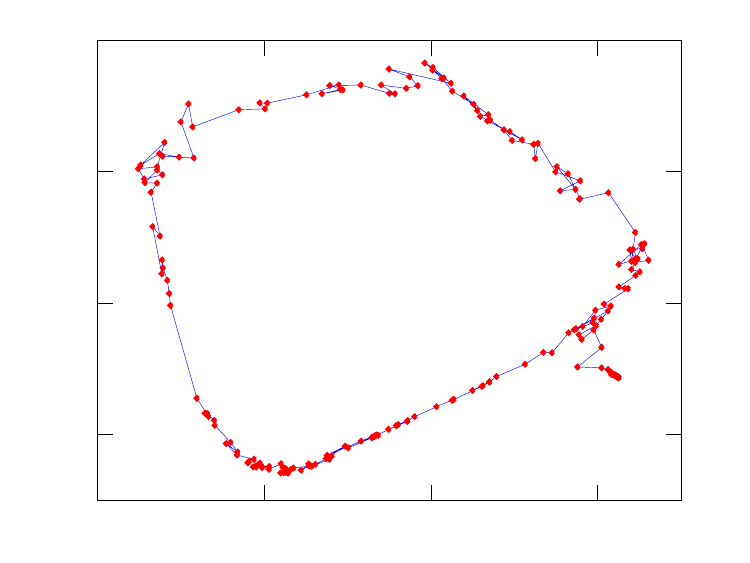}}%
    \gplfronttext
  \end{picture}%
\endgroup
  \hfill
\begingroup
  \makeatletter
  \providecommand\color[2][]{%
    \GenericError{(gnuplot) \space\space\space\@spaces}{%
      Package color not loaded in conjunction with
      terminal option `colourtext'%
    }{See the gnuplot documentation for explanation.%
    }{Either use 'blacktext' in gnuplot or load the package
      color.sty in LaTeX.}%
    \renewcommand\color[2][]{}%
  }%
  \providecommand\includegraphics[2][]{%
    \GenericError{(gnuplot) \space\space\space\@spaces}{%
      Package graphicx or graphics not loaded%
    }{See the gnuplot documentation for explanation.%
    }{The gnuplot epslatex terminal needs graphicx.sty or graphics.sty.}%
    \renewcommand\includegraphics[2][]{}%
  }%
  \providecommand\rotatebox[2]{#2}%
  \@ifundefined{ifGPcolor}{%
    \newif\ifGPcolor
    \GPcolortrue
  }{}%
  \@ifundefined{ifGPblacktext}{%
    \newif\ifGPblacktext
    \GPblacktexttrue
  }{}%
  \let\gplgaddtomacro\g@addto@macro
  \gdef\gplbacktext{}%
  \gdef\gplfronttext{}%
  \makeatother
  \ifGPblacktext
    \def\colorrgb#1{}%
    \def\colorgray#1{}%
  \else
    \ifGPcolor
      \def\colorrgb#1{\color[rgb]{#1}}%
      \def\colorgray#1{\color[gray]{#1}}%
      \expandafter\def\csname LTw\endcsname{\color{white}}%
      \expandafter\def\csname LTb\endcsname{\color{black}}%
      \expandafter\def\csname LTa\endcsname{\color{black}}%
      \expandafter\def\csname LT0\endcsname{\color[rgb]{1,0,0}}%
      \expandafter\def\csname LT1\endcsname{\color[rgb]{0,1,0}}%
      \expandafter\def\csname LT2\endcsname{\color[rgb]{0,0,1}}%
      \expandafter\def\csname LT3\endcsname{\color[rgb]{1,0,1}}%
      \expandafter\def\csname LT4\endcsname{\color[rgb]{0,1,1}}%
      \expandafter\def\csname LT5\endcsname{\color[rgb]{1,1,0}}%
      \expandafter\def\csname LT6\endcsname{\color[rgb]{0,0,0}}%
      \expandafter\def\csname LT7\endcsname{\color[rgb]{1,0.3,0}}%
      \expandafter\def\csname LT8\endcsname{\color[rgb]{0.5,0.5,0.5}}%
    \else
      \def\colorrgb#1{\color{black}}%
      \def\colorgray#1{\color[gray]{#1}}%
      \expandafter\def\csname LTw\endcsname{\color{white}}%
      \expandafter\def\csname LTb\endcsname{\color{black}}%
      \expandafter\def\csname LTa\endcsname{\color{black}}%
      \expandafter\def\csname LT0\endcsname{\color{black}}%
      \expandafter\def\csname LT1\endcsname{\color{black}}%
      \expandafter\def\csname LT2\endcsname{\color{black}}%
      \expandafter\def\csname LT3\endcsname{\color{black}}%
      \expandafter\def\csname LT4\endcsname{\color{black}}%
      \expandafter\def\csname LT5\endcsname{\color{black}}%
      \expandafter\def\csname LT6\endcsname{\color{black}}%
      \expandafter\def\csname LT7\endcsname{\color{black}}%
      \expandafter\def\csname LT8\endcsname{\color{black}}%
    \fi
  \fi
  \setlength{\unitlength}{0.0500bp}%
  \begin{picture}(4336.00,3252.00)%
    \gplgaddtomacro\gplbacktext{%
      \colorrgb{0.00,0.00,0.00}%
      \put(431,358){\makebox(0,0)[r]{\strut{}\small -3}}%
      \colorrgb{0.00,0.00,0.00}%
      \put(431,888){\makebox(0,0)[r]{\strut{}\small -2}}%
      \colorrgb{0.00,0.00,0.00}%
      \put(431,1418){\makebox(0,0)[r]{\strut{}\small -1}}%
      \colorrgb{0.00,0.00,0.00}%
      \put(431,1947){\makebox(0,0)[r]{\strut{}\small 0}}%
      \colorrgb{0.00,0.00,0.00}%
      \put(431,2477){\makebox(0,0)[r]{\strut{}\small 1}}%
      \colorrgb{0.00,0.00,0.00}%
      \put(431,3007){\makebox(0,0)[r]{\strut{}\small 2}}%
      \colorrgb{0.00,0.00,0.00}%
      \put(563,138){\makebox(0,0){\strut{}\small -3}}%
      \colorrgb{0.00,0.00,0.00}%
      \put(1235,138){\makebox(0,0){\strut{}\small -2}}%
      \colorrgb{0.00,0.00,0.00}%
      \put(1907,138){\makebox(0,0){\strut{}\small -1}}%
      \colorrgb{0.00,0.00,0.00}%
      \put(2580,138){\makebox(0,0){\strut{}\small 0}}%
      \colorrgb{0.00,0.00,0.00}%
      \put(3252,138){\makebox(0,0){\strut{}\small 1}}%
      \colorrgb{0.00,0.00,0.00}%
      \put(3924,138){\makebox(0,0){\strut{}\small 2}}%
    }%
    \gplgaddtomacro\gplfronttext{%
    }%
    \gplbacktext
    \put(0,0){\includegraphics{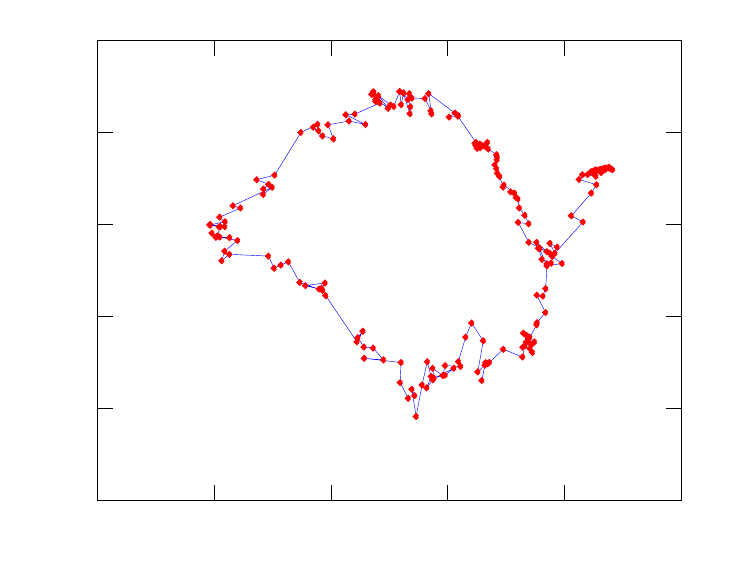}}%
    \gplfronttext
  \end{picture}%
\endgroup
  \end{center}
  \caption{Visualization of the robot WiFi navigation data using
    algorithms based on maximum entropy. Again seven neighbors are
    used.}\label{fig:embedRobot2}

\end{figure}

\begin{figure}
  \begin{center}
\begingroup
  \makeatletter
  \providecommand\color[2][]{%
    \GenericError{(gnuplot) \space\space\space\@spaces}{%
      Package color not loaded in conjunction with
      terminal option `colourtext'%
    }{See the gnuplot documentation for explanation.%
    }{Either use 'blacktext' in gnuplot or load the package
      color.sty in LaTeX.}%
    \renewcommand\color[2][]{}%
  }%
  \providecommand\includegraphics[2][]{%
    \GenericError{(gnuplot) \space\space\space\@spaces}{%
      Package graphicx or graphics not loaded%
    }{See the gnuplot documentation for explanation.%
    }{The gnuplot epslatex terminal needs graphicx.sty or graphics.sty.}%
    \renewcommand\includegraphics[2][]{}%
  }%
  \providecommand\rotatebox[2]{#2}%
  \@ifundefined{ifGPcolor}{%
    \newif\ifGPcolor
    \GPcolortrue
  }{}%
  \@ifundefined{ifGPblacktext}{%
    \newif\ifGPblacktext
    \GPblacktexttrue
  }{}%
  \let\gplgaddtomacro\g@addto@macro
  \gdef\gplbacktext{}%
  \gdef\gplfronttext{}%
  \makeatother
  \ifGPblacktext
    \def\colorrgb#1{}%
    \def\colorgray#1{}%
  \else
    \ifGPcolor
      \def\colorrgb#1{\color[rgb]{#1}}%
      \def\colorgray#1{\color[gray]{#1}}%
      \expandafter\def\csname LTw\endcsname{\color{white}}%
      \expandafter\def\csname LTb\endcsname{\color{black}}%
      \expandafter\def\csname LTa\endcsname{\color{black}}%
      \expandafter\def\csname LT0\endcsname{\color[rgb]{1,0,0}}%
      \expandafter\def\csname LT1\endcsname{\color[rgb]{0,1,0}}%
      \expandafter\def\csname LT2\endcsname{\color[rgb]{0,0,1}}%
      \expandafter\def\csname LT3\endcsname{\color[rgb]{1,0,1}}%
      \expandafter\def\csname LT4\endcsname{\color[rgb]{0,1,1}}%
      \expandafter\def\csname LT5\endcsname{\color[rgb]{1,1,0}}%
      \expandafter\def\csname LT6\endcsname{\color[rgb]{0,0,0}}%
      \expandafter\def\csname LT7\endcsname{\color[rgb]{1,0.3,0}}%
      \expandafter\def\csname LT8\endcsname{\color[rgb]{0.5,0.5,0.5}}%
    \else
      \def\colorrgb#1{\color{black}}%
      \def\colorgray#1{\color[gray]{#1}}%
      \expandafter\def\csname LTw\endcsname{\color{white}}%
      \expandafter\def\csname LTb\endcsname{\color{black}}%
      \expandafter\def\csname LTa\endcsname{\color{black}}%
      \expandafter\def\csname LT0\endcsname{\color{black}}%
      \expandafter\def\csname LT1\endcsname{\color{black}}%
      \expandafter\def\csname LT2\endcsname{\color{black}}%
      \expandafter\def\csname LT3\endcsname{\color{black}}%
      \expandafter\def\csname LT4\endcsname{\color{black}}%
      \expandafter\def\csname LT5\endcsname{\color{black}}%
      \expandafter\def\csname LT6\endcsname{\color{black}}%
      \expandafter\def\csname LT7\endcsname{\color{black}}%
      \expandafter\def\csname LT8\endcsname{\color{black}}%
    \fi
  \fi
  \setlength{\unitlength}{0.0500bp}%
  \begin{picture}(6264.00,4698.00)%
    \gplgaddtomacro\gplbacktext{%
      \colorrgb{0.00,0.00,0.00}%
      \put(682,996){\makebox(0,0)[r]{\strut{}\small Laplacian eigenmaps}}%
      \colorrgb{0.00,0.00,0.00}%
      \put(682,1474){\makebox(0,0)[r]{\strut{}\small LLE}}%
      \colorrgb{0.00,0.00,0.00}%
      \put(682,1953){\makebox(0,0)[r]{\strut{}\small isomap}}%
      \colorrgb{0.00,0.00,0.00}%
      \put(682,2431){\makebox(0,0)[r]{\strut{}\small MVU}}%
      \colorrgb{0.00,0.00,0.00}%
      \put(682,2910){\makebox(0,0)[r]{\strut{}\small MEU}}%
      \colorrgb{0.00,0.00,0.00}%
      \put(682,3388){\makebox(0,0)[r]{\strut{}\small Acyclic LLE}}%
      \colorrgb{0.00,0.00,0.00}%
      \put(682,3867){\makebox(0,0)[r]{\strut{}\small DRILL}}%
      \csname LTb\endcsname%
      \put(814,297){\makebox(0,0){\strut{}0}}%
      \put(3241,297){\makebox(0,0){\strut{}2000}}%
      \put(5668,297){\makebox(0,0){\strut{}4000}}%
    }%
    \gplgaddtomacro\gplfronttext{%
    }%
    \gplbacktext
    \put(0,0){\includegraphics{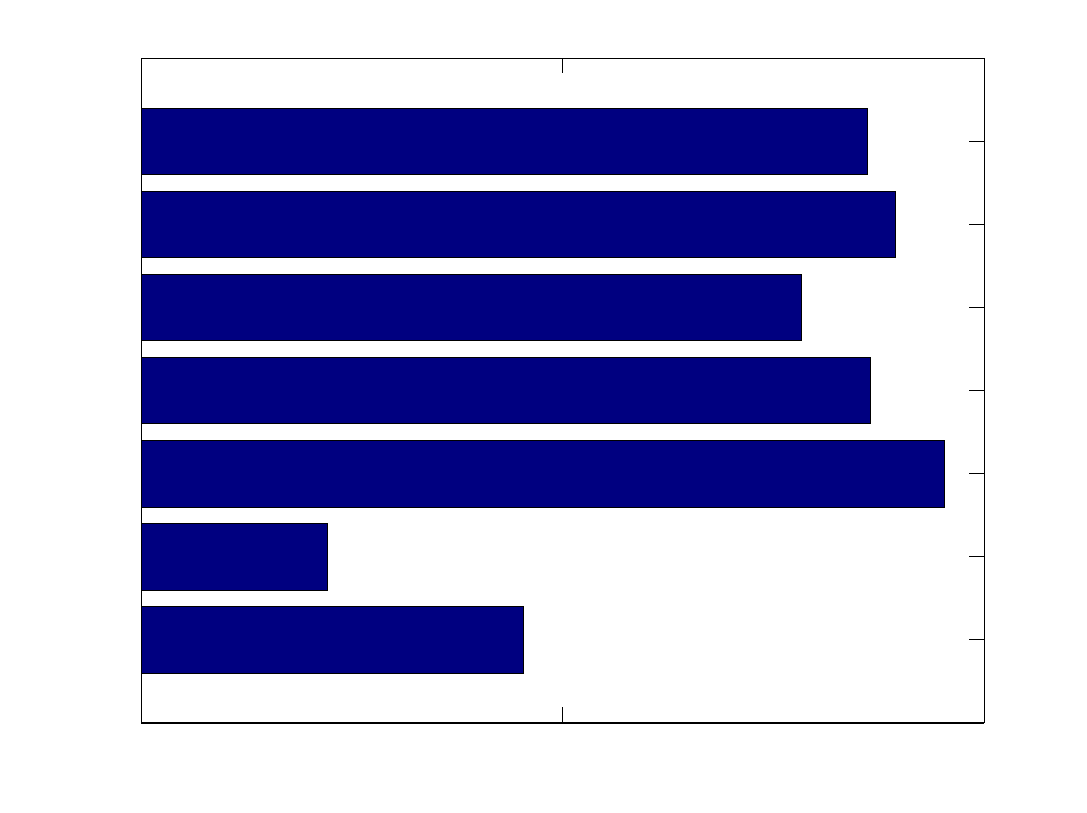}}%
    \gplfronttext
  \end{picture}%
\endgroup
  \hfill
\begingroup
  \makeatletter
  \providecommand\color[2][]{%
    \GenericError{(gnuplot) \space\space\space\@spaces}{%
      Package color not loaded in conjunction with
      terminal option `colourtext'%
    }{See the gnuplot documentation for explanation.%
    }{Either use 'blacktext' in gnuplot or load the package
      color.sty in LaTeX.}%
    \renewcommand\color[2][]{}%
  }%
  \providecommand\includegraphics[2][]{%
    \GenericError{(gnuplot) \space\space\space\@spaces}{%
      Package graphicx or graphics not loaded%
    }{See the gnuplot documentation for explanation.%
    }{The gnuplot epslatex terminal needs graphicx.sty or graphics.sty.}%
    \renewcommand\includegraphics[2][]{}%
  }%
  \providecommand\rotatebox[2]{#2}%
  \@ifundefined{ifGPcolor}{%
    \newif\ifGPcolor
    \GPcolortrue
  }{}%
  \@ifundefined{ifGPblacktext}{%
    \newif\ifGPblacktext
    \GPblacktexttrue
  }{}%
  \let\gplgaddtomacro\g@addto@macro
  \gdef\gplbacktext{}%
  \gdef\gplfronttext{}%
  \makeatother
  \ifGPblacktext
    \def\colorrgb#1{}%
    \def\colorgray#1{}%
  \else
    \ifGPcolor
      \def\colorrgb#1{\color[rgb]{#1}}%
      \def\colorgray#1{\color[gray]{#1}}%
      \expandafter\def\csname LTw\endcsname{\color{white}}%
      \expandafter\def\csname LTb\endcsname{\color{black}}%
      \expandafter\def\csname LTa\endcsname{\color{black}}%
      \expandafter\def\csname LT0\endcsname{\color[rgb]{1,0,0}}%
      \expandafter\def\csname LT1\endcsname{\color[rgb]{0,1,0}}%
      \expandafter\def\csname LT2\endcsname{\color[rgb]{0,0,1}}%
      \expandafter\def\csname LT3\endcsname{\color[rgb]{1,0,1}}%
      \expandafter\def\csname LT4\endcsname{\color[rgb]{0,1,1}}%
      \expandafter\def\csname LT5\endcsname{\color[rgb]{1,1,0}}%
      \expandafter\def\csname LT6\endcsname{\color[rgb]{0,0,0}}%
      \expandafter\def\csname LT7\endcsname{\color[rgb]{1,0.3,0}}%
      \expandafter\def\csname LT8\endcsname{\color[rgb]{0.5,0.5,0.5}}%
    \else
      \def\colorrgb#1{\color{black}}%
      \def\colorgray#1{\color[gray]{#1}}%
      \expandafter\def\csname LTw\endcsname{\color{white}}%
      \expandafter\def\csname LTb\endcsname{\color{black}}%
      \expandafter\def\csname LTa\endcsname{\color{black}}%
      \expandafter\def\csname LT0\endcsname{\color{black}}%
      \expandafter\def\csname LT1\endcsname{\color{black}}%
      \expandafter\def\csname LT2\endcsname{\color{black}}%
      \expandafter\def\csname LT3\endcsname{\color{black}}%
      \expandafter\def\csname LT4\endcsname{\color{black}}%
      \expandafter\def\csname LT5\endcsname{\color{black}}%
      \expandafter\def\csname LT6\endcsname{\color{black}}%
      \expandafter\def\csname LT7\endcsname{\color{black}}%
      \expandafter\def\csname LT8\endcsname{\color{black}}%
    \fi
  \fi
  \setlength{\unitlength}{0.0500bp}%
  \begin{picture}(6264.00,4698.00)%
    \gplgaddtomacro\gplbacktext{%
      \colorrgb{0.00,0.00,0.00}%
      \put(682,996){\makebox(0,0)[r]{\strut{}\small Laplacian eigenmaps}}%
      \colorrgb{0.00,0.00,0.00}%
      \put(682,1474){\makebox(0,0)[r]{\strut{}\small LLE}}%
      \colorrgb{0.00,0.00,0.00}%
      \put(682,1953){\makebox(0,0)[r]{\strut{}\small isomap}}%
      \colorrgb{0.00,0.00,0.00}%
      \put(682,2431){\makebox(0,0)[r]{\strut{}\small MVU}}%
      \colorrgb{0.00,0.00,0.00}%
      \put(682,2910){\makebox(0,0)[r]{\strut{}\small MEU}}%
      \colorrgb{0.00,0.00,0.00}%
      \put(682,3388){\makebox(0,0)[r]{\strut{}\small Acyclic LLE}}%
      \colorrgb{0.00,0.00,0.00}%
      \put(682,3867){\makebox(0,0)[r]{\strut{}\small DRILL}}%
      \csname LTb\endcsname%
      \put(814,297){\makebox(0,0){\strut{}-6000}}%
      \put(3241,297){\makebox(0,0){\strut{}-1000}}%
      \put(5668,297){\makebox(0,0){\strut{}4000}}%
    }%
    \gplgaddtomacro\gplfronttext{%
    }%
    \gplbacktext
    \put(0,0){\includegraphics{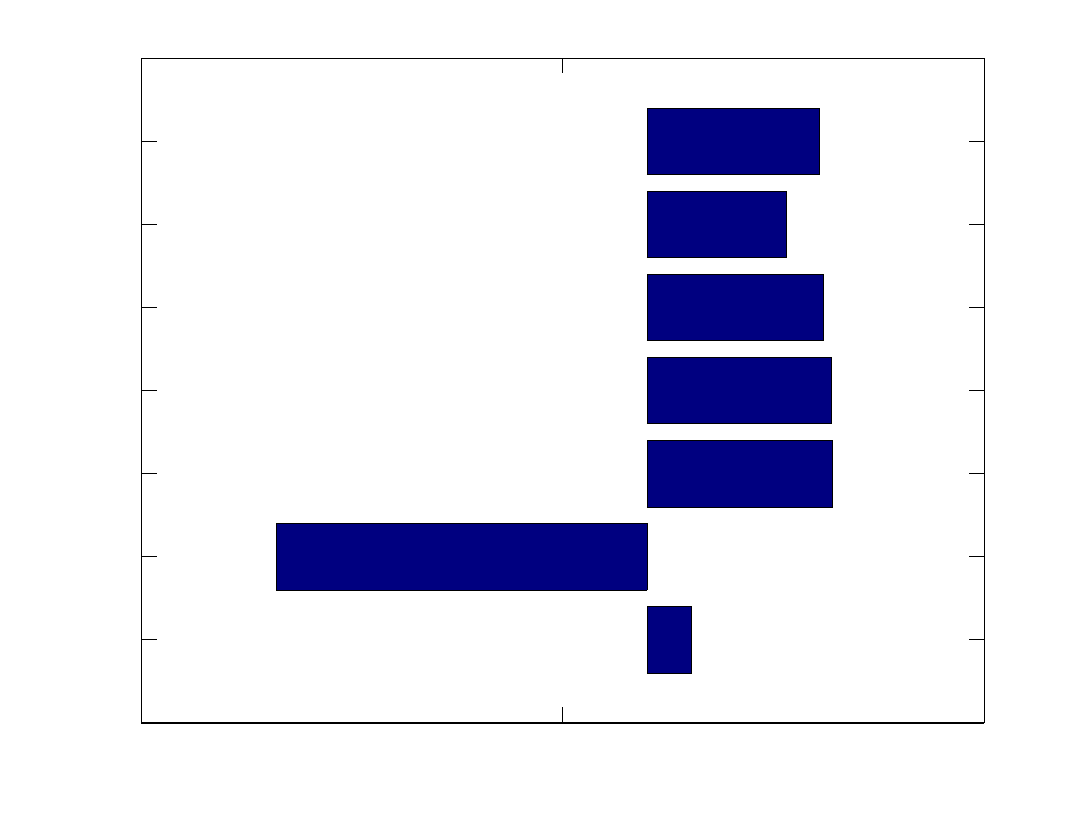}}%
    \gplfronttext
  \end{picture}%
\endgroup
  \end{center}
  \caption{Model scores for the different spectral approaches. (a) the
    motion capture data visualizations, (b) the robot navigation
    example visualizations.}\label{fig:histograms}
\end{figure}

The results for the range of spectral approaches are shown in
\reffigrange{fig:embedRobot1}{fig:embedRobot2} with the quality of the
methods scored in \reffig{fig:histograms}(b). Both in the
visualizations and in the GP-LVM scores we see a clear difference in
quality for the methods that preserve local distances (i.e. again
isomap, ALLE, MVU, MEU and DRILL are better than LLE and Laplacian
eigenmaps). Amongst the methods that do preserve local distance
relationships MEU seems to smooth the robot path more than the other
three approaches. Given that it has the lowest score of the four
distance preserving techniques this smoothing may be unwarranted. MVU
appears to have an overly noisy representation of the path.
\begin{octave}
  textSize = '\\small';
  clear score;
  models = {'le1',  'lle1', 'isomap1', 'mvu1', 'meu1',  'lle2', 'drill1'};
  names = {[textSize ' Laplacian eigenmaps'],  [textSize ' LLE'],  [textSize ' isomap'],  [textSize ' MVU'],  [textSize ' MEU'],  [textSize ' Acyclic LLE'], [textSize ' DRILL']};
  for i = 1:length(models);
    loadName = models{i};
    loadName(1) = upper(loadName(1));
    load(['demRobotWireless' loadName '.mat']);
    eval(['score(i) = ' models{i}(1:end-1) 'Info.score;']);
    eval(['X = ' models{i}(1:end-1) 'Info.X;']);
    figure(1)
    clf
    plot(X(:, 1), X(:, 2), 'b-');
    hold on
    b = plot(X(:, 1), X(:, 2), 'ro');
    set(b, 'markersize', 2, 'linewidth', 3);
    set(gca, 'xtick', [-3 -2 -1 0 1 2 3]);
    set(gca, 'xticklabel', {[textSize ' -3'], [textSize ' -2'], [textSize ' -1'], [textSize ' 0'], [textSize ' 1'], [textSize ' 2'], [textSize ' 3']});
    set(gca, 'ytick', [-3 -2 -1 0 1 2 3]);
    set(gca, 'yticklabel', {[textSize ' -3'], [textSize ' -2'], [textSize ' -1'], [textSize ' 0'], [textSize ' 1'], [textSize ' 2'], [textSize ' 3']});
    printLatexPlot(['demRobotWireless' loadName], '../../../meu/tex/diagrams/', 0.45*textWidth);
  end
  clf
  barh(score);
  set(gca, 'ytick', [1:length(names)])
  set(gca, 'yticklabel', names)
  set(gca, 'xtick', [-6000 -1000 4000]);
  printLatexPlot('demRobotWirelessBar1', '../../../meu/tex/diagrams/', 0.65*textWidth);
\end{octave}

\subsection{Learning the Neighborhood}

\begin{octave}
  for expNo = 2:3
    textSize = '\\small';
    clear score;
    if expNo == 3
      reg = [0.0005 0.001 0.0015 0.002 0.0025 0.003 0.0035 0.004 0.0045 0.005]; 
    else
      reg = [0 0.00001 0.00005 0.0001 0.0005 0.001 0.005 0.01 0.05 0.1];
    end

    for i = 1:length(reg);
      regStr = num2str(reg(i));
      regStr(find(regStr=='.')) = 'p';

      loadName = ['demStickDrill' num2str(expNo) '_' regStr];
      load(loadName);
      score(i) = model.score;
    end
    figure(1)
    clf
    barh(score);
    if expNo == 3
      set(gca, 'xlim', [2750 2900])
    end
    set(gca, 'ytick', [1:length(reg)])
    regS = cell(1, length(reg));
    for i = 1:length(reg)
      regS{i} = [textSize ' ' num2str(reg(i))];
    end
    set(gca, 'yticklabel', regS)
    if expNo == 3
      set(gca, 'xtick', [2750 2800 2850 2900]);
    else
      set(gca, 'xtick', [0 1000 2000 3000]);
    end
  
    printLatexPlot(['demStickDrillBar' num2str(expNo)], '../../../meu/tex/diagrams/', 0.45*textWidth);
  
    [void, ind] = max(score);
    regStr = num2str(reg(ind));
    regStr(find(regStr=='.')) = 'p';
    loadName = ['demStickDrill' num2str(expNo) '_' regStr '.mat'];
    load(loadName);
    X = model.X;
    figure(1)
    clf
    plot(X(:, 1), X(:, 2), 'b-');
    hold on
    b = plot(X(:, 1), X(:, 2), 'ro');
    set(b, 'markersize', 2, 'linewidth', 3);
    set(gca, 'xtick', [-3 -2 -1 0 1 2 3]);
    set(gca, 'xticklabel', {[textSize ' -3'], [textSize ' -2'], [textSize ' -1'], [textSize ' 0'], [textSize ' 1'], [textSize ' 2'], [textSize ' 3']});
    set(gca, 'ytick', [-3 -2 -1 0 1 2 3]);
    set(gca, 'yticklabel', {[textSize ' -3'], [textSize ' -2'], [textSize ' -1'], [textSize ' 0'], [textSize ' 1'], [textSize ' 2'], [textSize ' 3']});
    printLatexPlot(['demStickDrill' num2str(expNo)], '../../../meu/tex/diagrams/', 0.45*textWidth);
  end
  
\end{octave}

\begin{figure}
\begingroup
  \makeatletter
  \providecommand\color[2][]{%
    \GenericError{(gnuplot) \space\space\space\@spaces}{%
      Package color not loaded in conjunction with
      terminal option `colourtext'%
    }{See the gnuplot documentation for explanation.%
    }{Either use 'blacktext' in gnuplot or load the package
      color.sty in LaTeX.}%
    \renewcommand\color[2][]{}%
  }%
  \providecommand\includegraphics[2][]{%
    \GenericError{(gnuplot) \space\space\space\@spaces}{%
      Package graphicx or graphics not loaded%
    }{See the gnuplot documentation for explanation.%
    }{The gnuplot epslatex terminal needs graphicx.sty or graphics.sty.}%
    \renewcommand\includegraphics[2][]{}%
  }%
  \providecommand\rotatebox[2]{#2}%
  \@ifundefined{ifGPcolor}{%
    \newif\ifGPcolor
    \GPcolortrue
  }{}%
  \@ifundefined{ifGPblacktext}{%
    \newif\ifGPblacktext
    \GPblacktexttrue
  }{}%
  \let\gplgaddtomacro\g@addto@macro
  \gdef\gplbacktext{}%
  \gdef\gplfronttext{}%
  \makeatother
  \ifGPblacktext
    \def\colorrgb#1{}%
    \def\colorgray#1{}%
  \else
    \ifGPcolor
      \def\colorrgb#1{\color[rgb]{#1}}%
      \def\colorgray#1{\color[gray]{#1}}%
      \expandafter\def\csname LTw\endcsname{\color{white}}%
      \expandafter\def\csname LTb\endcsname{\color{black}}%
      \expandafter\def\csname LTa\endcsname{\color{black}}%
      \expandafter\def\csname LT0\endcsname{\color[rgb]{1,0,0}}%
      \expandafter\def\csname LT1\endcsname{\color[rgb]{0,1,0}}%
      \expandafter\def\csname LT2\endcsname{\color[rgb]{0,0,1}}%
      \expandafter\def\csname LT3\endcsname{\color[rgb]{1,0,1}}%
      \expandafter\def\csname LT4\endcsname{\color[rgb]{0,1,1}}%
      \expandafter\def\csname LT5\endcsname{\color[rgb]{1,1,0}}%
      \expandafter\def\csname LT6\endcsname{\color[rgb]{0,0,0}}%
      \expandafter\def\csname LT7\endcsname{\color[rgb]{1,0.3,0}}%
      \expandafter\def\csname LT8\endcsname{\color[rgb]{0.5,0.5,0.5}}%
    \else
      \def\colorrgb#1{\color{black}}%
      \def\colorgray#1{\color[gray]{#1}}%
      \expandafter\def\csname LTw\endcsname{\color{white}}%
      \expandafter\def\csname LTb\endcsname{\color{black}}%
      \expandafter\def\csname LTa\endcsname{\color{black}}%
      \expandafter\def\csname LT0\endcsname{\color{black}}%
      \expandafter\def\csname LT1\endcsname{\color{black}}%
      \expandafter\def\csname LT2\endcsname{\color{black}}%
      \expandafter\def\csname LT3\endcsname{\color{black}}%
      \expandafter\def\csname LT4\endcsname{\color{black}}%
      \expandafter\def\csname LT5\endcsname{\color{black}}%
      \expandafter\def\csname LT6\endcsname{\color{black}}%
      \expandafter\def\csname LT7\endcsname{\color{black}}%
      \expandafter\def\csname LT8\endcsname{\color{black}}%
    \fi
  \fi
  \setlength{\unitlength}{0.0500bp}%
  \begin{picture}(4336.00,3252.00)%
    \gplgaddtomacro\gplbacktext{%
      \colorrgb{0.00,0.00,0.00}%
      \put(431,579){\makebox(0,0)[r]{\strut{}\small 0}}%
      \colorrgb{0.00,0.00,0.00}%
      \put(431,800){\makebox(0,0)[r]{\strut{}\small 1e-05}}%
      \colorrgb{0.00,0.00,0.00}%
      \put(431,1020){\makebox(0,0)[r]{\strut{}\small 5e-05}}%
      \colorrgb{0.00,0.00,0.00}%
      \put(431,1241){\makebox(0,0)[r]{\strut{}\small 0.0001}}%
      \colorrgb{0.00,0.00,0.00}%
      \put(431,1462){\makebox(0,0)[r]{\strut{}\small 0.0005}}%
      \colorrgb{0.00,0.00,0.00}%
      \put(431,1683){\makebox(0,0)[r]{\strut{}\small 0.001}}%
      \colorrgb{0.00,0.00,0.00}%
      \put(431,1903){\makebox(0,0)[r]{\strut{}\small 0.005}}%
      \colorrgb{0.00,0.00,0.00}%
      \put(431,2124){\makebox(0,0)[r]{\strut{}\small 0.01}}%
      \colorrgb{0.00,0.00,0.00}%
      \put(431,2345){\makebox(0,0)[r]{\strut{}\small 0.05}}%
      \colorrgb{0.00,0.00,0.00}%
      \put(431,2566){\makebox(0,0)[r]{\strut{}\small 0.1}}%
      \csname LTb\endcsname%
      \put(1235,138){\makebox(0,0){\strut{}0}}%
      \put(1907,138){\makebox(0,0){\strut{}1000}}%
      \put(2580,138){\makebox(0,0){\strut{}2000}}%
      \put(3252,138){\makebox(0,0){\strut{}3000}}%
    }%
    \gplgaddtomacro\gplfronttext{%
    }%
    \gplbacktext
    \put(0,0){\includegraphics{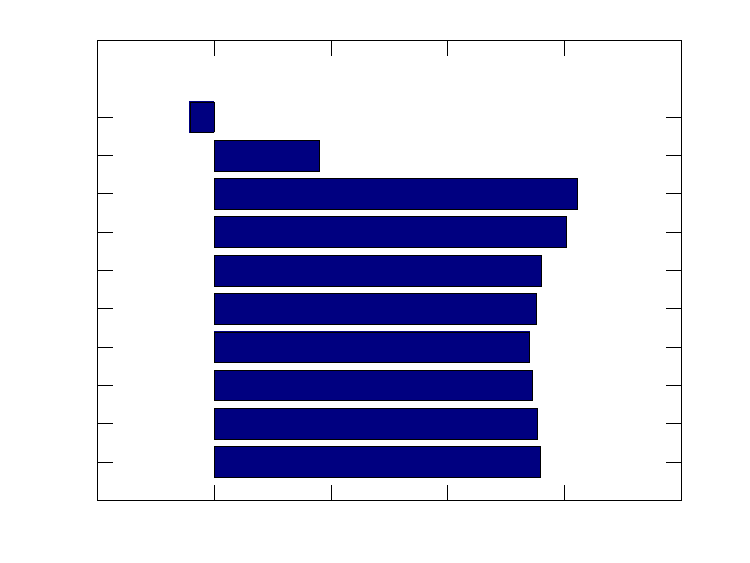}}%
    \gplfronttext
  \end{picture}%
\endgroup
  \hfill
\begingroup
  \makeatletter
  \providecommand\color[2][]{%
    \GenericError{(gnuplot) \space\space\space\@spaces}{%
      Package color not loaded in conjunction with
      terminal option `colourtext'%
    }{See the gnuplot documentation for explanation.%
    }{Either use 'blacktext' in gnuplot or load the package
      color.sty in LaTeX.}%
    \renewcommand\color[2][]{}%
  }%
  \providecommand\includegraphics[2][]{%
    \GenericError{(gnuplot) \space\space\space\@spaces}{%
      Package graphicx or graphics not loaded%
    }{See the gnuplot documentation for explanation.%
    }{The gnuplot epslatex terminal needs graphicx.sty or graphics.sty.}%
    \renewcommand\includegraphics[2][]{}%
  }%
  \providecommand\rotatebox[2]{#2}%
  \@ifundefined{ifGPcolor}{%
    \newif\ifGPcolor
    \GPcolortrue
  }{}%
  \@ifundefined{ifGPblacktext}{%
    \newif\ifGPblacktext
    \GPblacktexttrue
  }{}%
  \let\gplgaddtomacro\g@addto@macro
  \gdef\gplbacktext{}%
  \gdef\gplfronttext{}%
  \makeatother
  \ifGPblacktext
    \def\colorrgb#1{}%
    \def\colorgray#1{}%
  \else
    \ifGPcolor
      \def\colorrgb#1{\color[rgb]{#1}}%
      \def\colorgray#1{\color[gray]{#1}}%
      \expandafter\def\csname LTw\endcsname{\color{white}}%
      \expandafter\def\csname LTb\endcsname{\color{black}}%
      \expandafter\def\csname LTa\endcsname{\color{black}}%
      \expandafter\def\csname LT0\endcsname{\color[rgb]{1,0,0}}%
      \expandafter\def\csname LT1\endcsname{\color[rgb]{0,1,0}}%
      \expandafter\def\csname LT2\endcsname{\color[rgb]{0,0,1}}%
      \expandafter\def\csname LT3\endcsname{\color[rgb]{1,0,1}}%
      \expandafter\def\csname LT4\endcsname{\color[rgb]{0,1,1}}%
      \expandafter\def\csname LT5\endcsname{\color[rgb]{1,1,0}}%
      \expandafter\def\csname LT6\endcsname{\color[rgb]{0,0,0}}%
      \expandafter\def\csname LT7\endcsname{\color[rgb]{1,0.3,0}}%
      \expandafter\def\csname LT8\endcsname{\color[rgb]{0.5,0.5,0.5}}%
    \else
      \def\colorrgb#1{\color{black}}%
      \def\colorgray#1{\color[gray]{#1}}%
      \expandafter\def\csname LTw\endcsname{\color{white}}%
      \expandafter\def\csname LTb\endcsname{\color{black}}%
      \expandafter\def\csname LTa\endcsname{\color{black}}%
      \expandafter\def\csname LT0\endcsname{\color{black}}%
      \expandafter\def\csname LT1\endcsname{\color{black}}%
      \expandafter\def\csname LT2\endcsname{\color{black}}%
      \expandafter\def\csname LT3\endcsname{\color{black}}%
      \expandafter\def\csname LT4\endcsname{\color{black}}%
      \expandafter\def\csname LT5\endcsname{\color{black}}%
      \expandafter\def\csname LT6\endcsname{\color{black}}%
      \expandafter\def\csname LT7\endcsname{\color{black}}%
      \expandafter\def\csname LT8\endcsname{\color{black}}%
    \fi
  \fi
  \setlength{\unitlength}{0.0500bp}%
  \begin{picture}(4336.00,3252.00)%
    \gplgaddtomacro\gplbacktext{%
      \colorrgb{0.00,0.00,0.00}%
      \put(431,736){\makebox(0,0)[r]{\strut{}\small -1}}%
      \colorrgb{0.00,0.00,0.00}%
      \put(431,1493){\makebox(0,0)[r]{\strut{}\small 0}}%
      \colorrgb{0.00,0.00,0.00}%
      \put(431,2250){\makebox(0,0)[r]{\strut{}\small 1}}%
      \colorrgb{0.00,0.00,0.00}%
      \put(431,3007){\makebox(0,0)[r]{\strut{}\small 2}}%
      \colorrgb{0.00,0.00,0.00}%
      \put(563,138){\makebox(0,0){\strut{}\small -2}}%
      \colorrgb{0.00,0.00,0.00}%
      \put(1235,138){\makebox(0,0){\strut{}\small -1}}%
      \colorrgb{0.00,0.00,0.00}%
      \put(1907,138){\makebox(0,0){\strut{}\small 0}}%
      \colorrgb{0.00,0.00,0.00}%
      \put(2580,138){\makebox(0,0){\strut{}\small 1}}%
      \colorrgb{0.00,0.00,0.00}%
      \put(3252,138){\makebox(0,0){\strut{}\small 2}}%
      \colorrgb{0.00,0.00,0.00}%
      \put(3924,138){\makebox(0,0){\strut{}\small 3}}%
    }%
    \gplgaddtomacro\gplfronttext{%
    }%
    \gplbacktext
    \put(0,0){\includegraphics{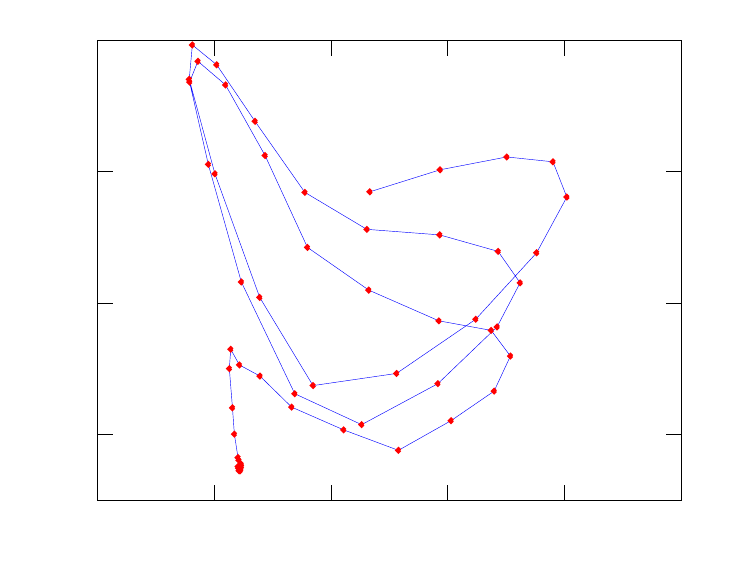}}%
    \gplfronttext
  \end{picture}%
\endgroup
  \caption{Structure learning for the DRILL algorithm on the motion
    capture data set. A model with 20 neighbors was fitted to the
    data. L1 regularization was used to reduce the number of neighbors
    associated with each data point. (a) shows the model score for the
    different L1 regularization parameters and (b) shows the
    visualization that corresponded to the best score (regularization
    parameter is 0.01).}\label{fig:twentyNeighbors}
\end{figure}

\begin{figure}
\begingroup
  \makeatletter
  \providecommand\color[2][]{%
    \GenericError{(gnuplot) \space\space\space\@spaces}{%
      Package color not loaded in conjunction with
      terminal option `colourtext'%
    }{See the gnuplot documentation for explanation.%
    }{Either use 'blacktext' in gnuplot or load the package
      color.sty in LaTeX.}%
    \renewcommand\color[2][]{}%
  }%
  \providecommand\includegraphics[2][]{%
    \GenericError{(gnuplot) \space\space\space\@spaces}{%
      Package graphicx or graphics not loaded%
    }{See the gnuplot documentation for explanation.%
    }{The gnuplot epslatex terminal needs graphicx.sty or graphics.sty.}%
    \renewcommand\includegraphics[2][]{}%
  }%
  \providecommand\rotatebox[2]{#2}%
  \@ifundefined{ifGPcolor}{%
    \newif\ifGPcolor
    \GPcolortrue
  }{}%
  \@ifundefined{ifGPblacktext}{%
    \newif\ifGPblacktext
    \GPblacktexttrue
  }{}%
  \let\gplgaddtomacro\g@addto@macro
  \gdef\gplbacktext{}%
  \gdef\gplfronttext{}%
  \makeatother
  \ifGPblacktext
    \def\colorrgb#1{}%
    \def\colorgray#1{}%
  \else
    \ifGPcolor
      \def\colorrgb#1{\color[rgb]{#1}}%
      \def\colorgray#1{\color[gray]{#1}}%
      \expandafter\def\csname LTw\endcsname{\color{white}}%
      \expandafter\def\csname LTb\endcsname{\color{black}}%
      \expandafter\def\csname LTa\endcsname{\color{black}}%
      \expandafter\def\csname LT0\endcsname{\color[rgb]{1,0,0}}%
      \expandafter\def\csname LT1\endcsname{\color[rgb]{0,1,0}}%
      \expandafter\def\csname LT2\endcsname{\color[rgb]{0,0,1}}%
      \expandafter\def\csname LT3\endcsname{\color[rgb]{1,0,1}}%
      \expandafter\def\csname LT4\endcsname{\color[rgb]{0,1,1}}%
      \expandafter\def\csname LT5\endcsname{\color[rgb]{1,1,0}}%
      \expandafter\def\csname LT6\endcsname{\color[rgb]{0,0,0}}%
      \expandafter\def\csname LT7\endcsname{\color[rgb]{1,0.3,0}}%
      \expandafter\def\csname LT8\endcsname{\color[rgb]{0.5,0.5,0.5}}%
    \else
      \def\colorrgb#1{\color{black}}%
      \def\colorgray#1{\color[gray]{#1}}%
      \expandafter\def\csname LTw\endcsname{\color{white}}%
      \expandafter\def\csname LTb\endcsname{\color{black}}%
      \expandafter\def\csname LTa\endcsname{\color{black}}%
      \expandafter\def\csname LT0\endcsname{\color{black}}%
      \expandafter\def\csname LT1\endcsname{\color{black}}%
      \expandafter\def\csname LT2\endcsname{\color{black}}%
      \expandafter\def\csname LT3\endcsname{\color{black}}%
      \expandafter\def\csname LT4\endcsname{\color{black}}%
      \expandafter\def\csname LT5\endcsname{\color{black}}%
      \expandafter\def\csname LT6\endcsname{\color{black}}%
      \expandafter\def\csname LT7\endcsname{\color{black}}%
      \expandafter\def\csname LT8\endcsname{\color{black}}%
    \fi
  \fi
  \setlength{\unitlength}{0.0500bp}%
  \begin{picture}(4336.00,3252.00)%
    \gplgaddtomacro\gplbacktext{%
      \colorrgb{0.00,0.00,0.00}%
      \put(431,579){\makebox(0,0)[r]{\strut{}\small 0.0005}}%
      \colorrgb{0.00,0.00,0.00}%
      \put(431,800){\makebox(0,0)[r]{\strut{}\small 0.001}}%
      \colorrgb{0.00,0.00,0.00}%
      \put(431,1020){\makebox(0,0)[r]{\strut{}\small 0.0015}}%
      \colorrgb{0.00,0.00,0.00}%
      \put(431,1241){\makebox(0,0)[r]{\strut{}\small 0.002}}%
      \colorrgb{0.00,0.00,0.00}%
      \put(431,1462){\makebox(0,0)[r]{\strut{}\small 0.0025}}%
      \colorrgb{0.00,0.00,0.00}%
      \put(431,1683){\makebox(0,0)[r]{\strut{}\small 0.003}}%
      \colorrgb{0.00,0.00,0.00}%
      \put(431,1903){\makebox(0,0)[r]{\strut{}\small 0.0035}}%
      \colorrgb{0.00,0.00,0.00}%
      \put(431,2124){\makebox(0,0)[r]{\strut{}\small 0.004}}%
      \colorrgb{0.00,0.00,0.00}%
      \put(431,2345){\makebox(0,0)[r]{\strut{}\small 0.0045}}%
      \colorrgb{0.00,0.00,0.00}%
      \put(431,2566){\makebox(0,0)[r]{\strut{}\small 0.005}}%
      \csname LTb\endcsname%
      \put(563,138){\makebox(0,0){\strut{}2750}}%
      \put(1683,138){\makebox(0,0){\strut{}2800}}%
      \put(2804,138){\makebox(0,0){\strut{}2850}}%
      \put(3924,138){\makebox(0,0){\strut{}2900}}%
    }%
    \gplgaddtomacro\gplfronttext{%
    }%
    \gplbacktext
    \put(0,0){\includegraphics{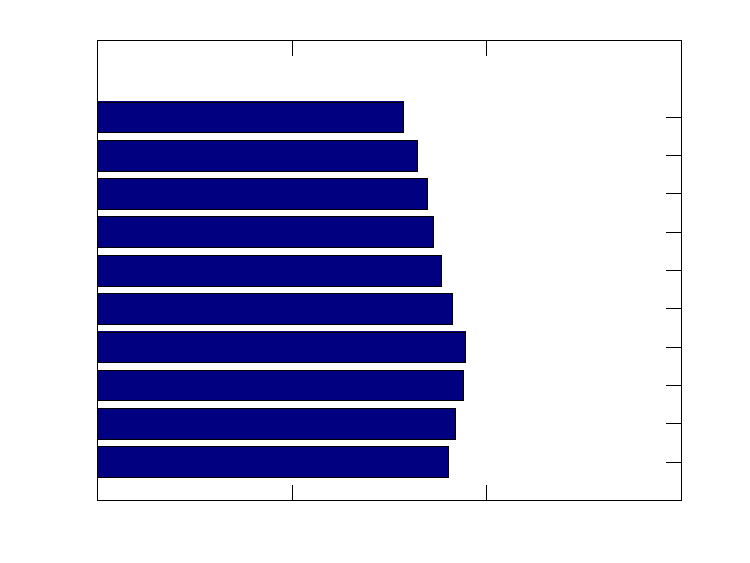}}%
    \gplfronttext
  \end{picture}%
\endgroup
  \hfill
\begingroup
  \makeatletter
  \providecommand\color[2][]{%
    \GenericError{(gnuplot) \space\space\space\@spaces}{%
      Package color not loaded in conjunction with
      terminal option `colourtext'%
    }{See the gnuplot documentation for explanation.%
    }{Either use 'blacktext' in gnuplot or load the package
      color.sty in LaTeX.}%
    \renewcommand\color[2][]{}%
  }%
  \providecommand\includegraphics[2][]{%
    \GenericError{(gnuplot) \space\space\space\@spaces}{%
      Package graphicx or graphics not loaded%
    }{See the gnuplot documentation for explanation.%
    }{The gnuplot epslatex terminal needs graphicx.sty or graphics.sty.}%
    \renewcommand\includegraphics[2][]{}%
  }%
  \providecommand\rotatebox[2]{#2}%
  \@ifundefined{ifGPcolor}{%
    \newif\ifGPcolor
    \GPcolortrue
  }{}%
  \@ifundefined{ifGPblacktext}{%
    \newif\ifGPblacktext
    \GPblacktexttrue
  }{}%
  \let\gplgaddtomacro\g@addto@macro
  \gdef\gplbacktext{}%
  \gdef\gplfronttext{}%
  \makeatother
  \ifGPblacktext
    \def\colorrgb#1{}%
    \def\colorgray#1{}%
  \else
    \ifGPcolor
      \def\colorrgb#1{\color[rgb]{#1}}%
      \def\colorgray#1{\color[gray]{#1}}%
      \expandafter\def\csname LTw\endcsname{\color{white}}%
      \expandafter\def\csname LTb\endcsname{\color{black}}%
      \expandafter\def\csname LTa\endcsname{\color{black}}%
      \expandafter\def\csname LT0\endcsname{\color[rgb]{1,0,0}}%
      \expandafter\def\csname LT1\endcsname{\color[rgb]{0,1,0}}%
      \expandafter\def\csname LT2\endcsname{\color[rgb]{0,0,1}}%
      \expandafter\def\csname LT3\endcsname{\color[rgb]{1,0,1}}%
      \expandafter\def\csname LT4\endcsname{\color[rgb]{0,1,1}}%
      \expandafter\def\csname LT5\endcsname{\color[rgb]{1,1,0}}%
      \expandafter\def\csname LT6\endcsname{\color[rgb]{0,0,0}}%
      \expandafter\def\csname LT7\endcsname{\color[rgb]{1,0.3,0}}%
      \expandafter\def\csname LT8\endcsname{\color[rgb]{0.5,0.5,0.5}}%
    \else
      \def\colorrgb#1{\color{black}}%
      \def\colorgray#1{\color[gray]{#1}}%
      \expandafter\def\csname LTw\endcsname{\color{white}}%
      \expandafter\def\csname LTb\endcsname{\color{black}}%
      \expandafter\def\csname LTa\endcsname{\color{black}}%
      \expandafter\def\csname LT0\endcsname{\color{black}}%
      \expandafter\def\csname LT1\endcsname{\color{black}}%
      \expandafter\def\csname LT2\endcsname{\color{black}}%
      \expandafter\def\csname LT3\endcsname{\color{black}}%
      \expandafter\def\csname LT4\endcsname{\color{black}}%
      \expandafter\def\csname LT5\endcsname{\color{black}}%
      \expandafter\def\csname LT6\endcsname{\color{black}}%
      \expandafter\def\csname LT7\endcsname{\color{black}}%
      \expandafter\def\csname LT8\endcsname{\color{black}}%
    \fi
  \fi
  \setlength{\unitlength}{0.0500bp}%
  \begin{picture}(4336.00,3252.00)%
    \gplgaddtomacro\gplbacktext{%
      \colorrgb{0.00,0.00,0.00}%
      \put(431,358){\makebox(0,0)[r]{\strut{}\small -2}}%
      \colorrgb{0.00,0.00,0.00}%
      \put(431,1020){\makebox(0,0)[r]{\strut{}\small -1}}%
      \colorrgb{0.00,0.00,0.00}%
      \put(431,1683){\makebox(0,0)[r]{\strut{}\small 0}}%
      \colorrgb{0.00,0.00,0.00}%
      \put(431,2345){\makebox(0,0)[r]{\strut{}\small 1}}%
      \colorrgb{0.00,0.00,0.00}%
      \put(431,3007){\makebox(0,0)[r]{\strut{}\small 2}}%
      \colorrgb{0.00,0.00,0.00}%
      \put(1043,138){\makebox(0,0){\strut{}\small -1}}%
      \colorrgb{0.00,0.00,0.00}%
      \put(2003,138){\makebox(0,0){\strut{}\small 0}}%
      \colorrgb{0.00,0.00,0.00}%
      \put(2964,138){\makebox(0,0){\strut{}\small 1}}%
      \colorrgb{0.00,0.00,0.00}%
      \put(3924,138){\makebox(0,0){\strut{}\small 2}}%
    }%
    \gplgaddtomacro\gplfronttext{%
    }%
    \gplbacktext
    \put(0,0){\includegraphics{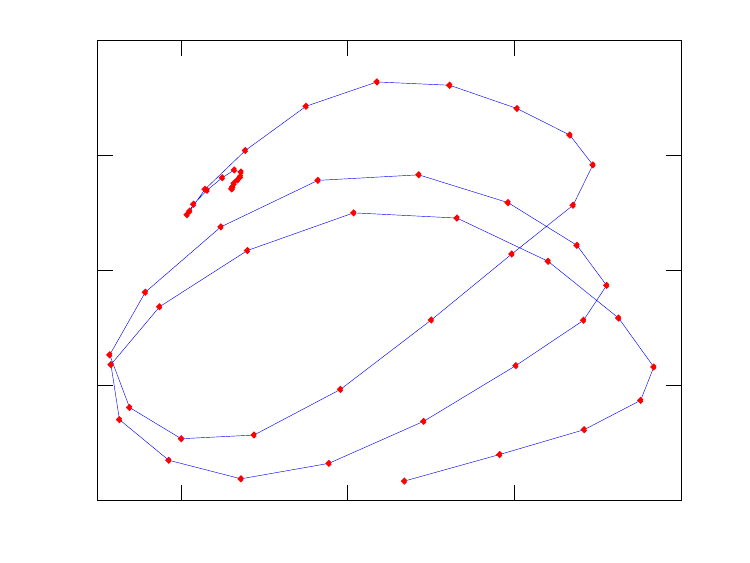}}%
    \gplfronttext
  \end{picture}%
\endgroup
  \caption{Full structure learning for the DRILL algorithm. Here all
    points are considered neighbors, the structure of the model is
    then recovered by L1 regularization. (a) shows the model score
    associated with the different L1 regularization parameters and (b)
    shows the visualization corresponding to the best score
    (regularization parameter 0.002).}\label{fig:fullNeighborhood}
\end{figure}

Our final experiments test the ability of L1 regularization of the
random field to learn the neighborhood. We firstly considered the
motion capture data and used the DRILL with a large neighborhood size
of 20 and L1 regularization on the parameters. As we varied the
regularization coefficient we found a maximum under the GP-LVM score
(\reffig{fig:twentyNeighbors}(a)). The visualization associated
with this maximum is shown in \reffig{fig:twentyNeighbors}(b) this
may be compared with \reffig{fig:embedStick2}(c) which used 6
neighbors. Finally we investigated whether L1 regularization alone
could recover a reasonable representation of the data. We again
considered the motion capture data but initialized all points as
neighbors. We then applied L1 regularization to learn a neighborhood
structure. Again a maximum under the GP-LVM score was found (\reffig{fig:fullNeighborhood}(a)) and the visualization associated with
this maximum is shown \reffig{fig:fullNeighborhood}(b).

The structural learning prior was able to improve the
model fitted with 20 neighbors considerably until its performance was
similar to that of the the six neighbor model shown in
\reffig{fig:embedStick2}(c). However, L1 regularization alone was not able
to obtain such a good performance, and was unable to tease out the
starting position from the rest of the run in the final
visualization. It appears that structural learning using L1-priors for
sparsity is not on its own enough to find an appropriate neighborhood
structure for this data set.

\section{Discussion and Conclusions}

We have introduced a new perspective on dimensionality reduction
algorithms based around maximum entropy. Our starting point was the
maximum variance unfolding and our end point was a novel approach to
dimensionality reduction based on Gaussian random fields and lasso
based structure learning. We hope that this new perspective on
dimensionality reduction will encourage new strands of research at the
interface between these areas.

One feature that stands out from our unifying perspective \citep[see
  also][]{Ham:kernelDimred04,Bengio:outofsample03,Bengio:eigenfunctions04}
is the three separate stages used in existing spectral dimensionality
algorithms.
\begin{enumerate}
\item A neighborhood between data points is selected. Normally
  $k$-nearest neighbors or similar algorithms are used.
\item Interpoint distances between neighbors are fed to the algorithms
  which provide a similarity matrix. The way the entries in the
  similarity matrix are computed is the main difference between the
  different algorithms.
\item The relationship between points in the similarity matrix is
  visualized using the eigenvectors of the similarity matrix.
\end{enumerate}

Our unifying perspective shows that actually each of these steps is
somewhat orthogonal. The neighborhood relations need not come from
nearest neighbors, we can use structural learning algorithms such as
that suggested in DRILL to learn the interpoint structure. The main
difference between the different approaches to spectral dimensionality
reduction is how the entries of the similarity matrix are
determined. Maximum variance unfolding looks to maximize the trace
under the distance constraints from the neighbours. Our new algorithms
maximize the entropy or, equivalently, the likelihood of the
data. Locally linear embedding maximizes an approximation to our
likelihood. Laplacian eigenmaps parameterize the inverse similarity
through appealing to physical analogies. Finally, isomap uses shortest
path algorithms to compute interpoint distances and centres the
resulting matrix to give the similarities.

The final step of the algorithm attempts to visualize the similarity
matrices using their eigenvectors. However, it simply makes use of one
possible objective function to perform this visualization. Considering
that underlying the similarity matrix, $\kernelMatrix$, is a sparse
Laplacian matrix, $\laplacianMatrix$, which represents a
Gaussian-Markov random field, we can see this final step as
visualizing that random field. There are many potential ways to
visualize that field and the eigenvectors of the precision is just one
of them. In fact, there is an entire field of graph visualization
proposing different approaches to visualizing such graphs. However, we
could even choose not to visualize the resulting graph. It may be that
the structure of the graph is of interest in itself. Work in human
cognition by \cite{Kemp:form08} has sought to fit Gaussian graphical
models to data in natural structures such as trees, chains and
rings. Visualization of such graphs through reduced dimensional spaces
is only likely to be appropriate in some cases, for example planar
structures. For this model only the first two steps are necessary.

One advantage to conflating the three steps we've identified is the
possibility to speed up the complete algorithm. For example,
conflating the second and third step allows us to speed up algorithms
through never explicitly computing the similarity matrix. Using the
fact that the principal eigenvectors of the similarity are the minor
eigenvalues of the Laplacian and exploiting fast eigensolvers that act
on sparse matrices very large data sets can be addressed. However, we
still can understand the algorithm from the unifying perspective while
exploiting the computational advantages offered by this neat shortcut.

\subsection{Gaussian Process Latent Variable Models}

Finally, there are similarities between maximum entropy unfolding and
the Gaussian process latent variable model (GP-LVM). Both specify a
Gaussian density over the training data and in practise the GP-LVM
normally makes an assumption of independence across the features. In
the GP-LVM a Gaussian process is defined that maps from the latent
space, $\latentMatrix$, to the data space, $\dataMatrix$. The
resulting likelihood is then optimized with respect to the latent
points, $\latentMatrix$. Maximum entropy unfolding leads to a Gauss
Markov Random field, where the conditional dependencies are between
neighbors. In one dimension, a Gauss Markov random field can easily be
specified by a Gaussian process through appropriate covariance
functions. The Ornstein-Uhlbeck covariance function is the unique
covariance function for a stationary Gauss Markov process. If such a
covariance was defined in a GP-LVM with a one dimensional latent space
\[
\kernelScalar(\latentScalar, \latentScalar^\prime) = \exp
(-\loneNorm{\latentScalar-\latentScalar\prime})
\]
then the inverse covariance will be sparse with only the nearest
neighbors in the one dimensional latent space connected. The elements
of the inverse covariance would be dependent on the distance between
the two latent points, which in the GP-LVM is optimized as part of the
training procedure. The resulting model is strikingly similar to the
MEU model, but in the GP-LVM the neighborhood is learnt by the model
(through optimization of $\latentMatrix$), rather than being specified
in advance. The visualization is given directly by the resulting
$\latentMatrix$. There is no secondary step of performing an
eigendecomposition to recover the point positions. For larger latent
dimensions and different neighborhood sizes, the exact correspondence
is harder to establish: Gaussian processes are defined on a continuous
space and the Markov property we exploit in MEU arises from discrete
relations. But the models are still similar in that they proscribe a
Gaussian covariance across the data points which is derived from the
spatial relationships between the points.

\subsection*{Notes}

The plots in this document were generated using MATweave. Code was
Octave version 3.2.3\verb+x86_64-pc-linux-gnu+
\subsection*{Acknowledgements}

Conversations with John Kent, Chris Williams, Brenden Lake, Joshua
Tenenbaum and John Lafferty have influenced the thinking in this
paper.  
{ 
  \bibliographystyle{plainnat}
  \bibliography{spectral} 
}
\appendix

\end{document}